\documentclass{article}

\usepackage{libertine}
\usepackage[T1]{fontenc}

\usepackage{booktabs}
\usepackage[letterpaper,margin=1in]{geometry}
\newcommand{\equal}[1]{{\hypersetup{linkcolor=black}\thanks{#1}}}
\usepackage{amsmath,amsthm,amssymb}
\usepackage[round]{natbib}

\usepackage[hyphens, spaces, obeyspaces]{url}     
\usepackage{hyperref}

\hypersetup{
    colorlinks=true,
    linkcolor=blue,
    filecolor=magenta,      
    urlcolor=blue,
}
\usepackage{graphicx}
\usepackage{adjustbox}
\usepackage{float}
\usepackage{cprotect}
\usepackage{siunitx}
\usepackage{subfigure}
\usepackage{todonotes}

\title{
Evaluating Model Performance in Medical Datasets Over Time
}
\date{}

\author{Helen Zhou\equal{These authors contributed equally}, Yuwen Chen\footnotemark[1], Zachary C. Lipton\\
Carnegie Mellon University\\
\text{\{
\href{mailto:hlzhou@andrew.cmu.edu}{hlzhou},
\href{mailto:yuwenc2@andrew.cmu.edu}{yuwenc2},
\href{mailto:zlipton@cmu.edu}{zlipton}\}@andrew.cmu.edu}}

\begin{document}

\maketitle

\begin{abstract}
Machine learning (ML) models deployed in healthcare systems
must face data drawn from continually evolving environments.
However, 
researchers proposing 
such models typically evaluate them
in a time-agnostic manner, splitting datasets 
according to patients sampled randomly 
throughout the entire study time period. 
This work proposes the Evaluation on Medical Datasets Over Time (EMDOT) framework, 
which evaluates the performance of a model class across time. 
Inspired by the concept of backtesting, 
EMDOT simulates possible training procedures that practitioners 
might have been able to execute at each point in time
and evaluates the resulting models on all future time points. 
Evaluating both linear and more complex models on six distinct medical data sources (tabular and imaging), we
show how depending on the dataset, 
using all historical data may be ideal in many cases, 
whereas using a window of the most recent data could be advantageous in others. 
In datasets 
where models suffer from sudden degradations in performance,
we investigate plausible explanations for these shocks.
We release the EMDOT package to help facilitate 
further works in deployment-oriented evaluation over time.
\end{abstract}

\paragraph*{Data and Code Availability}
We use the following data: 
(1) the Surveillance, Epidemiology, and End Results (SEER) cancer dataset \citep{seerdataset}, (2) the COVID-19 Case Surveillance Detailed Data provided by the CDC \citep{cdc_data}, (3) the Southwestern Pennsylvania (SWPA) COVID-19 dataset, (4) the MIMIC-IV intensive care database \citep{mimiciv_v1}, (5) the Organ Procurement and Transplantation Network (OPTN) database for liver transplant candidates \citep{optn_data}, and (6) the MIMIC-CXR-JPG database of chest radiographs \citep{johnson2019mimic_cxr_database,johnson_pollard_greenbaum_lungren_deng_peng_lu_mark_berkowitz_horng_etal_2019}. MIMIC-IV and MIMIC-CXR-JPG (referred to as MIMIC-CXR in this paper) are available on the PhysioNet repository
\citep{goldberger2000physiobank}. Except for the SWPA dataset, all are publicly accessible (after accepting a data usage agreement). Details for accessing each dataset are in Appendices \ref{app:seer_data}--\ref{app:optn_data}. The code is publicly available on GitHub.

\paragraph*{Institutional Review Board (IRB)}
This research does not require IRB approval. 

\section{Introduction}
\label{sec:intro}
As medical practices, healthcare systems, and community environments evolve over time, so does the distribution of collected data. Features are deprecated as new ones are introduced, data collection may fluctuate along with hospital policies, and the underlying 
patient and disease 
populations 
may shift. 

Amidst this ever-changing environment, models that perform well on one time period 
cannot 
be assumed to perform well in perpetuity. 
In the MIMIC-III critical care dataset, 
\citet{nestor2019feature} found that a 
change to the electronic health record (EHR) system 
in 2008 
coincided with %
sudden 
degradations 
in AUROC for mortality prediction. 
In 
COVID-19 data
from the Centers for Disease Control and Prevention (CDC), 
\citet{cheng2021unpacking} noted that the age distribution among 
cases shifted continually throughout the pandemic, 
and that these continual shifts 
confounded estimates of improvements in mortality rate. 
We propose an evaluation framework to characterize model performance over time by simulating 
training procedures that practitioners could have executed up to each time point, and subsequently deployed in future time points. %
We argue that
standard time-agnostic evaluation is insufficient for selecting deployment-ready models, showing
across several datasets
that 
it 
over-estimates deployment performance. 
Instead, we advocate for EMDOT as a worthwhile pre-deployment step 
to help practitioners
gain confidence in the robustness of their models to shifts in the data distribution that have occurred in the past and may to some extent repeat in the future. 

There is a large body of work that 
addresses 
adaptation 
under
various structured forms of distribution shift, including covariate shift
\citep{shimodaira2000improving,zadrozny2004learning,huang2006correcting,sugiyama2007direct,gretton2009covariate},
label shift
\citep{saerens2002adjusting, storkey2009training,zhang2013domain,lipton2018detecting,garg2020unified},
missingness shift \citep{zhou2022domain},
and concept drift \citep{tsymbal2004problem,gama2014survey}.
However, in the real-world medical datasets we analyze, none of these structural assumptions can be guaranteed, and distributional changes in covariates, labels, missingness, etc. could even occur simultaneously. This motivates our empirical work, as it is unclear across a variety of model classes and medical datasets, how existing models might degrade due to naturally occurring changes over time, and whether different training practices might impact on robustness over time.

However intuitive it might seem, 
evaluation of models over time 
remains uncommon in standard machine learning for healthcare (ML4H) research. 
In the proceedings of the Conference on Health, Inference, and Learning (CHIL) 2022, for example, none of the 23 papers performed evaluations which took time into account (see Appendix \ref{app:review_evals} for similar statistics from CHIL 2021 and the Radiology medical journal).
One possible reason for this is lack of access---as noted by \citet{nestor2019feature}, 
it is common practice to remove timestamps when
de-identifying medical datasets for public use.
In this work, we identify six sources of medical data containing varying granularities of 
temporal information per-record, 
five of which are \emph{publicly available}.
We profile the performance of various training strategies and model classes across time, and 
identify possible sources of distribution shifts 
within each dataset.
Finally, we release the Evaluation on Medical Datasets Over Time (EMDOT) Python package (details in Appendix \ref{app:emdot_appendix})
to allow researchers to apply EMDOT
to their own datasets and
test techniques for
handling
shifts over time.

\section{Related work}
The promise of ML 
for improving healthcare 
has been explored in several domains, 
including 
cancer survival prediction \citep{hegselmann2018reproducible}, 
diabetic retinopathy detection \citep{google_diabetic_retinopathy2016}, 
antimicrobial stewardship \citep{kanjilal2020decision,boominathan2020},
recognizing diagnoses from electronic health record data \citep{lipton2015learning},
and mortality prediction in liver transplant candidates \citep{
bertsimas2019development,
byrd2021predicting}.
Typically, these ML models are evaluated on randomly held out patients, and sometimes externally validated on other hospitals or newly collected data. Even with cross-site validations, we cannot be sure how models will perform in the future. 
For decades, the medical community has had a history of utilizing (mostly) fixed, simple risk scores to inform patient care \citep{hermansson2018systematic,kamath2001model,wilson1998prediction,wells1995accuracy}. Risk scores often prioritize ease-of-use, are computed from 
few variables, verified by domain experts for clear causal 
connections 
to outcomes of interest,
and validated through use over time and across hospitals. Together, these factors give clinicians confidence that the model will perform reliably for years to come.
With increasingly complex models, however, trust and adoption 
may be hindered by a lack of 
confidence in robustness to changing environments.

As noted by \citet{d2020underspecification}, ML models often exhibit unexpectedly poor behavior when deployed in real-world domains. A key reason for these failures, 
they argue, is \emph{under-specification}, where ML pipelines yield many predictors with equivalently strong held-out performance in the training domain, but such predictors can behave very differently in deployment. 
By testing performance across a variety of distribution shifts that have previously occurred over time, EMDOT could serve as a stress test to help combat under-specification.

Although evaluation over time is far from standard in ML4H literature, changes in performance over time have been noted in prior work.
To predict wound-healing, \cite{jung2015implications} 
found that when data were split by cutoff time instead of patients, benefits of model averaging and stacking disappeared.
\citet{pianykh2020continuous} found degradation in performance of a model for wait times dependent on how much historical data was trained on. To predict severe COVID-19, \citet{ahn_clinical_concepts} found that learned clinical concept features
performed more robustly over time than raw features. 
Closest to our work is \citet{nestor2019feature}, 
which
evaluated %
AUROC in MIMIC-III critical care data from 2003--2012, comparing training on 
just 2001--2002; the prior year; and the full history.
Using the full history and curated clinical concepts, they bridged a big 
drop in performance
due to changing EHR systems.
Whereas \citet{nestor2019feature} %
considers 
three models %
per test year, EMDOT simulates model deployment every year and evaluates %
across \emph{all future years}. 

While we do not consider time series models in this work 
(instead considering those which treat data as i.i.d.),
there are similarities between how training sets are defined in EMDOT and in techniques for evaluating time-series forecasts \citep{BERGMEIR2012192,cerqueira2020evaluating}. These techniques often roll forward in time, taking either a window of recent data or all historical data as training sets, and evaluate test performance on the next time point. Performance from each time point is then averaged to summarize performance. 
This type of back-testing technique is common in rapidly evolving, non-stationary applications like finance \citep{pmlr-v119-chauhan20a,alberg2017improving}, where time series models are constantly updated.
In the healthcare domain, however, models may not be so easily updated, with risk scores developed several years ago still being used to this day \citep{six2008chest,kamath2001model,wilson1998prediction,wells1995accuracy}.
Thus, we track performance not only the immediate year after the training set, but all subsequent years in the dataset. Additionally, instead of 
collapsing performance from models trained
at different time points into summary statistics, 
which
could conceal 
distribution 
shifts over time, 
our framework 
tracks these granular fluctuations over time, and creates tools to help 
provide insight into the nature and potential causes of such changes.

\begin{table*}[t]
\caption{Summary of datasets used for analysis. For more details, see Appendices \ref{app:seer_data}--\ref{app:optn_data}.}
\label{tab:dataset_info}
\centering
    \begin{tabular}{lccccc}
    \toprule
    Dataset name        &  Outcome          & Time Range (time point unit) & \# samples & \# positives \\
    \midrule
    SEER (Breast)  & 5-year Survival   & 1975--2013 (year)            & 462,023   & 378,758   \\
    SEER (Colon)   & 5-year Survival   & 1975--2013 (year)            & 254,112   & 135,065   \\
    SEER (Lung)    & 5-year Survival   & 1975--2013 (year)            & 457,695   & 49,997 \\
    CDC COVID-19        & Mortality         & Mar 2020--May 2022 (month)  & 941,140   & 190,786  \\
    SWPA COVID-19         & 90-day Mortality  & Mar 2020--Feb 2022 (month)  & 35,293    & 1,516  \\
    MIMIC-IV            & In-ICU Mortality  & 2009--2020 (year)            & 53,050    & 3,334  \\
    OPTN (Liver)          & 180-day Mortality &  2005--2017 (year)           & 143,709   & 4,635 \\
    MIMIC-CXR & 14 diagnostic labels & 2010--2018 (year)& 376,204 & 209,088 \\
    \bottomrule
    \end{tabular}%
\end{table*}

\section{Data}
We sought medical datasets that had: (1) a timestamp for each record, (2) interesting prediction task(s), and (3) enough distinct time points to evaluate over. 
Six data sources
satisfied these criteria:
SEER cancer data, national CDC COVID-19 data, COVID-19 data from a healthcare provider in Southwestern Pennsylvania (SWPA), MIMIC-IV critical care data, OPTN data from liver transplant candidates, and MIMIC-CXR chest radiographs. All datasets are tabular except for MIMIC-CXR (medical imaging data).
All but SWPA are publicly accessible.

Table \ref{tab:dataset_info} summarizes the dataset outcomes, time ranges, and number of samples. 
Figure \ref{fig:bar_plot_number_of_samples_pos_outcomes} visualizes 
data quantity over time. 
Appendices \ref{app:seer_data}--\ref{app:mimic_cxr_data} 
include cohort selection diagrams, cohort characteristics, features, heat maps of missingness, preprocessing steps, 
and additional details. 
Categorical variables are converged to dummies, and numerical variables are normalized and centered at 0. Missing values in categorical variables are treated as another category, and in numerical variables they are imputed with the mean.
In all datasets except MIMIC-CXR (where each sample is a distinct radiograph), each sample corresponds to a distinct patient. 

\subsection{SEER Cancer Data}

The Surveillance, Epidemiology, and End Results (SEER) Program collects cancer incidence data from 
registries
throughout the U.S. 
Each case includes 
demographics, primary tumor site, tumor morphology, stage, diagnosis, first course of treatment, and survival outcomes
(collected with follow-up) \citep{seerdataset}. 
We use the SEER$^{*}$Stat software \citep{surveillance2015national} to define three cohorts of interest: (1) breast cancer, (2) colon cancer, and (3) lung cancer. 
The outcome is 5-year survival, i.e. whether the patient was confirmed alive five years after the year of diagnosis. The amount of data has mostly increased each year (Figure \ref{fig:bar_plot_number_of_samples_pos_outcomes}). Performance over time is evaluated \emph{yearly}.
See Appendix \ref{app:seer_data} for more details.

\begin{figure}%
\begin{center}
\includegraphics[width=0.6\columnwidth]{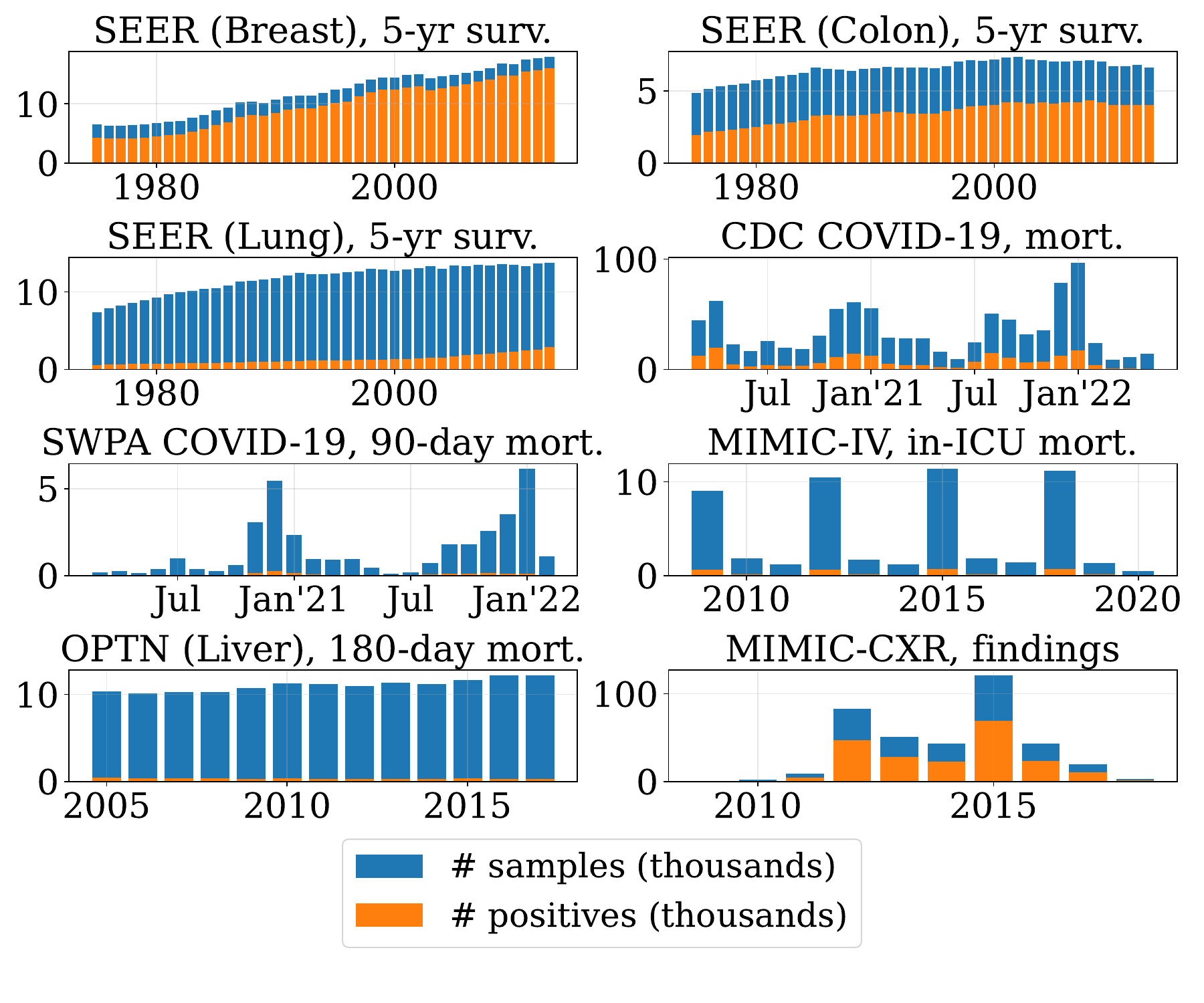}
\vspace{-0.8em}
\caption{Number of samples and positive\protect\footnotemark outcomes per time point.}
\vspace{-1.5em}
\label{fig:bar_plot_number_of_samples_pos_outcomes}
\end{center}
\end{figure}
\footnotetext{In MIMIC-CXR, all labels except ``No Finding'' are considered positive for the purposes of Figure \ref{fig:bar_plot_number_of_samples_pos_outcomes} and Table \ref{tab:dataset_info}.}

\subsection{National CDC COVID-19 Data}
The COVID-19 Case Surveillance Detailed Data \citep{cdc_data} is a national dataset provided by the CDC. It has the largest number of samples among the datasets considered, and contains
33 elements, with patient-level data including symptoms, demographics, and state of residence. The cohort consists of all lab-confirmed positive COVID-19 cases that were hospitalized, so the quantity of samples over time has a seasonality reflecting surges in COVID-19 (Figure \ref{fig:bar_plot_number_of_samples_pos_outcomes}).
The outcome of interest is mortality, defined by \verb|death_yn = Yes| in the dataset. Performance over time is evaluated on a \emph{monthly} basis. See Appendix \ref{app:cdc_data} for more details.

\subsection{SWPA COVID-19 Data}
The Southwestern Pennsylvania (SWPA) COVID-19 dataset consists of EHR data from patients tested for COVID-19. It is the smallest dataset considered in this paper, and was collected by a major healthcare provider in SWPA. Features include
patient demographics, labs, problem histories, medications, inpatient vs. outpatient status, and other information collected in the patient encounter. 
The cohort consists of COVID-19 patients testing positive for the first time, and not already in the ICU or mechanically ventilated. Similar to the CDC COVID-19 dataset, there is a seasonality to the monthly number of samples that reflects surges in COVID-19 (Figure \ref{fig:bar_plot_number_of_samples_pos_outcomes}).
The outcome of interest is 90-day mortality, derived by comparing the death date and test date.
The performance over time is evaluated on a \emph{monthly} basis. See Appendix \ref{app:swpa_data} for more details.

\subsection{MIMIC-IV Critical Care Data}

The Medical Information Mart for Intensive Care (MIMIC)-IV \citep{mimiciv_v1} database contains EHR data from patients admitted to critical care units 
from 2008--2019. MIMIC-IV is an update to MIMIC-III, adding time annotations placing each sample into a three-year time range, and removing elements from the old CareVue EHR system (before 2008). We approximate the year of each sample by taking the midpoint of its time range, but note that this causes certain years (2009, 2012, 2015, 2018) to have substantially more samples than others (Figure \ref{fig:bar_plot_number_of_samples_pos_outcomes}).
The cohort is selected by taking the first encounter of all patients in the \verb|icustays| table, and the outcome of interest is in-ICU mortality.
Performance over time is evaluated on a \emph{yearly} basis. See Appendix \ref{app:mimic_data} for more details.

\subsection{OPTN Liver Transplant Data}
The Organ Procurement and Transplantation Network (OPTN) database tracks organ donation and transplant events in the U.S. 
The selected cohort consists of liver transplant candidates on the waiting list. The same pipeline as \citet{byrd2021predicting} is used to extract the data, except that the first record is selected for each patient. 
The outcome of interest is 180-day mortality from when the patient was  
added to the list.
The performance over time is evaluated on a \emph{yearly} basis. More details are in Appendix \ref{app:optn_data}.

\subsection{MIMIC-CXR}
The MIMIC Chest X-ray (MIMIC-CXR) JPG dataset \citep{johnson_pollard_greenbaum_lungren_deng_peng_lu_mark_berkowitz_horng_etal_2019} contains chest radiographs in JPG format. Similar to MIMIC-IV, we approximate the year by taking the midpoint of its three-year time range. The selected cohort consists of all radiographs from 2010 to 2018. The outcomes of interest are 14 diagnostic labels: Atelectasis, Cardiomegaly, Consolidation, Edema, Enlarged Cardiomediastinum, Fracture, Lung Lesion, Lung Opacity, Pleural Effusion, Pneumonia, Pneumothorax, Pleural Other, Support Devices, and No Finding. Performance over time is evaluated on a \emph{yearly} basis. More details are in Appendix \ref{app:mimic_cxr_data}.

\section{Methods}

We tackle the following guiding questions:
\begin{enumerate}
    \item On each dataset, what would the reported performance of a model be if it were trained using standard time-agnostic splits (\textbf{all-period})?
    \item \textbf{Simulating} how a practitioner might have trained and deployed models in the past, how would performance have varied \textbf{over time}?
    \item When might it be better to train on a \textbf{recent window} of data 
    versus \textbf{all historical} data?
    \item What is the comparative performance of different \textbf{classes of models} over time?
    \item To what extent might we be able to diagnose possible \textbf{reasons} for changes in model performance?
\end{enumerate}

\subsection{All-period Training}\label{sec:standard_split}
We mimic common practice in evaluation by 
using time-agnostic data splits which randomly place patients from the entire study time range into train, validation, and test sets (details in Appendix \ref{app:datasplit_details}), and reporting the test set performance. We refer to training with this type of split as \emph{all-period} training.

\subsection{EMDOT Evaluation}\label{sec:eot}

For more realistic simulation of
how practitioners train models 
and subsequently deploy 
them on 
future data, we define the \emph{Evaluation on Medical Datasets Over Time} (EMDOT) framework. 
At each time point $t$ (termed \emph{simulated deployment date}), an \emph{in-period} subset of data from times $\leq t$ is available for model development. After training a model on this in-period data, one might be interested in both recent in-period performance (at time $t$) and future \emph{out-of-period} performance (at times $>t$). 

\begin{figure}[t]
\begin{center}
  \includegraphics[width=0.6\columnwidth]{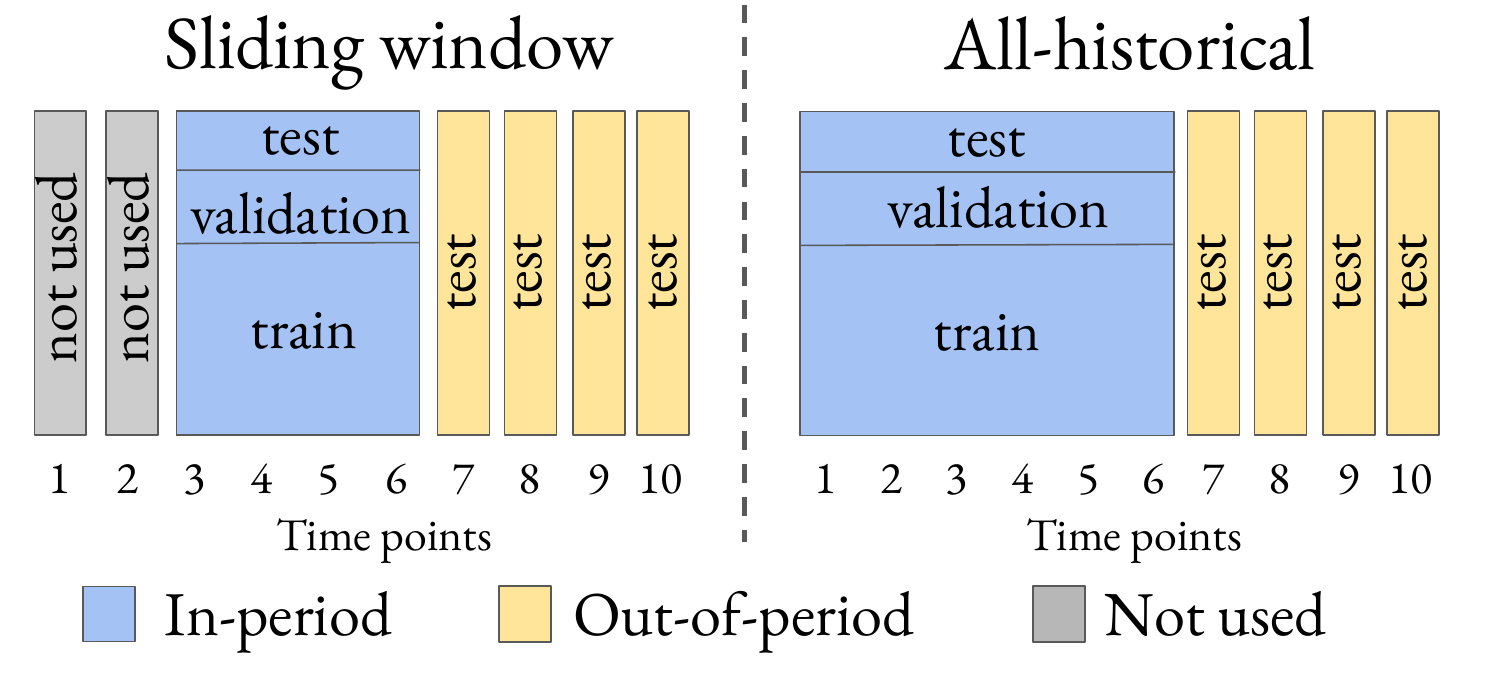}
  \caption{EMDOT training regimes, with a simulated deployment date of $t=6$.}
  \label{fig:training_regimes}
\end{center}
\end{figure}

In-period data is split into train, validation, and test sets 
(split ratios in Appendix \ref{app:datasplit_details}). For MIMIC-CXR, where one patient could have multiple radiographs, the data is split such that there are no overlapping patients between splits.
Recent in-period performance is evaluated on held-out test data from the most recent time point. Out-of-period performance is evaluated on all 
data from 
each future time point. For example, a model trained up to time 6 is tested on data from 6, 7, 8, etc. (Figure \ref{fig:training_regimes}). At time 8, the model is considered two time points \emph{stale}. 
Although this procedure can take $O(T)$ times more computation than all-period training for $T$ time points, we argue that this procedure yields a more realistic view of the type of performance that one might expect models to have over time.
Additionally, practitioners face a tradeoff between using %
recent data perhaps most reflective of the present %
and using 
all available 
historical data for a larger sample size. %
Intuitively, the former may be appealing in modern applications with massive datasets, whereas the latter may be necessary in data-scarce applications. We explore these two training regimes, with different definitions of in-period data (Figure \ref{fig:training_regimes}): 
\begin{enumerate}
    \item \textbf{Sliding window}: The last $W$ time points are considered in-period. In this paper, we use window size $W$ = 4 for sufficient positive examples.%
    \item \textbf{All-historical}: Any data prior to the current time point is considered in-period.
\end{enumerate}

To decouple the effect of sample size from that of shifts in the data distribution, comparisons are also performed with all-historical data that is \textbf{sub-sampled} to be the same size as the corresponding training set under the sliding window training regime.

To summarize more formally, let $D_t$ refer to the set of all data points occurring at time $t \in \{1, ..., T\}$, where $T$ is the number of time points that the dataset spans. 
Each $D_t$ can be partitioned by splitting patients at random into disjoint train, validation, and test sets: 
$D_t = D_t^{\text{train}} \cup D_t^{\text{val}} \cup D_t^{\text{test}}.$
For simulated deployment dates $t^* \in \{W, W+1, ..., T\}$, training, validation, and test sets are defined for the \emph{sliding window} training regime as follows:

\begin{itemize}
    \item training: $\bigcup_{k=t^* - W + 1}^{t^*} D_k^{\text{train}}$
    \item validation: $\bigcup_{k=t^* - W + 1}^{t^*} D_k^{\text{val}}$
    \item in-period test: $D_{t^*}^{\text{test}}$
    \item out-of-period test: $D_{k}$ for $k = t^* + 1, ..., T$
\end{itemize}

Training, validation, and test sets are defined for the \emph{all-historical} training regime as follows:

\begin{itemize}
    \item training: $\bigcup_{k=1}^{t^*} D_k^{\text{train}}$
    \item validation: $\bigcup_{k=1}^{t^*} D_k^{\text{val}}$
    \item in-period test: $D_{t^*}^{\text{test}}$
    \item out-of-period test: $D_{k}$ for $k = t^* + 1, ..., T$
\end{itemize}

At each simulated deployment date $t^*$, models are trained using the training set, validated using the validation set, and tested on the in-period test set as well as all out-of-period test sets. If a model with simulated deployment date $t^*$ is being evaluated on an out of period test set $D_{t^* + j}$, then the model is $j$ time points \emph{stale}.

\subsection{Evaluation Metrics} All binary classification tasks are evaluated by AUROC. For multi-label prediction in MIMIC-CXR, each of the 14 diagnostic labels is treated as a separate binary classification task, and a weighted sum of AUROCs is computed, where the weight for a particular label is given by the proportional prevalence of that label among all positive labels. That is, for some class $a$, its weight is $p_a / \sum_x p_x$, where $p_x$ is the number of positives with label $x$.
Samples are treated in an i.i.d. manner for training. 

\subsection{Models}
Logistic regression (LR), gradient boosted decision trees (GBDT) and feedforward neural networks (MLP) are trained on the tabular datasets. DenseNet-121 is trained on the MIMIC-CXR imaging dataset. Hyperparameters are selected based on in-period validation performance, and the hyperparameter grids are in Appendix \ref{app:hyperparameter_grids}.

\subsection{Detecting Sources of Change}
To better understand possible reasons for changing performance,
we create \emph{diagnostic plots} to 
track model performance alongside changes in the data distribution over time.

In tabular datasets, we 
plot feature importances and average 
values of the most important features over time. 
Generating these plots for logistic regression, we define feature importance by the %
magnitudes of the coefficients, but note that other feature importance techniques 
could be used for more complex model classes. To avoid overcrowding the plots, we take the union of the top $k$ most important features from each time point is taken, where $k$ is tuned depending on the dataset. We additionally highlight (using a thicker line) categorical features with consistently high prevalence or which experience a large change in prevalence across one time point, and numerical features with high average rank
(see Appendix \ref{app:diagnostic_plot} for thresholds for each dataset).

For the imaging dataset, where feature importance is less straightforward, we plot the distribution of pixel intensities over time, along with proportions of each of the 14 diagnostic labels. 

By highlighting sudden changes in model performance and the corresponding time periods in all other plots, diagnostic plots can help bring attention to shifts in the distribution of data that coincide with changing model performance.

\subsection{EMDOT Python Package}
We release the EMDOT python package\footnote{https://github.com/acmi-lab/EvaluationOverTime} to help practitioners move from standard model evaluation to EMDOT evaluation. See Appendix \ref{app:emdot_appendix} for a schematic of the EMDOT workflow, and see the GitHub repository for a step-by-step tutorial.

\begin{table*}[ht]
\caption{Test AUROC from all-period training and time-agnostic evaluation.}
\label{tab:AUROC_splitting_by_patient}
\centering
    \begin{tabular}{lcccccccc}
    \toprule
    Model
    & \multicolumn{1}{p{1.2cm}}{\centering SEER (Breast)}
    & \multicolumn{1}{p{1.2cm}}{\centering SEER (Colon)}
    & \multicolumn{1}{p{1.2cm}}{\centering SEER (Lung)}
    & \multicolumn{1}{p{1.8cm}}{\centering CDC COVID-19}
    & \multicolumn{1}{p{1.8cm}}{\centering SWPA COVID-19}
    & \multicolumn{1}{p{1.2cm}}{\centering MIMIC-IV}
    & \multicolumn{1}{p{1.2cm}}{\centering OPTN (Liver)}
    & \multicolumn{1}{p{1.2cm}}{\centering MIMIC-CXR}\\
    \midrule
    LR &         0.888 &        0.863 &       0.894 &        0.837 &         0.928 &    \textbf{0.935} &        0.846 & - \\
    GBDT &          \textbf{0.891} &        0.868 &       0.894 &         0.851 &         \textbf{0.930} &    0.931 &        \textbf{0.854} & - \\
    MLP &          \textbf{0.891} &        \textbf{0.869} &         \textbf{0.898} &        \textbf{0.852} &         0.928 &    0.898 &        0.847 & - \\
    DensetNet & - & - & - & - & - & - & - & \textbf{0.860} \\
    \bottomrule
    \end{tabular}%
\end{table*}
\begin{figure*}[ht]
\centering
\includegraphics[width=1\columnwidth]{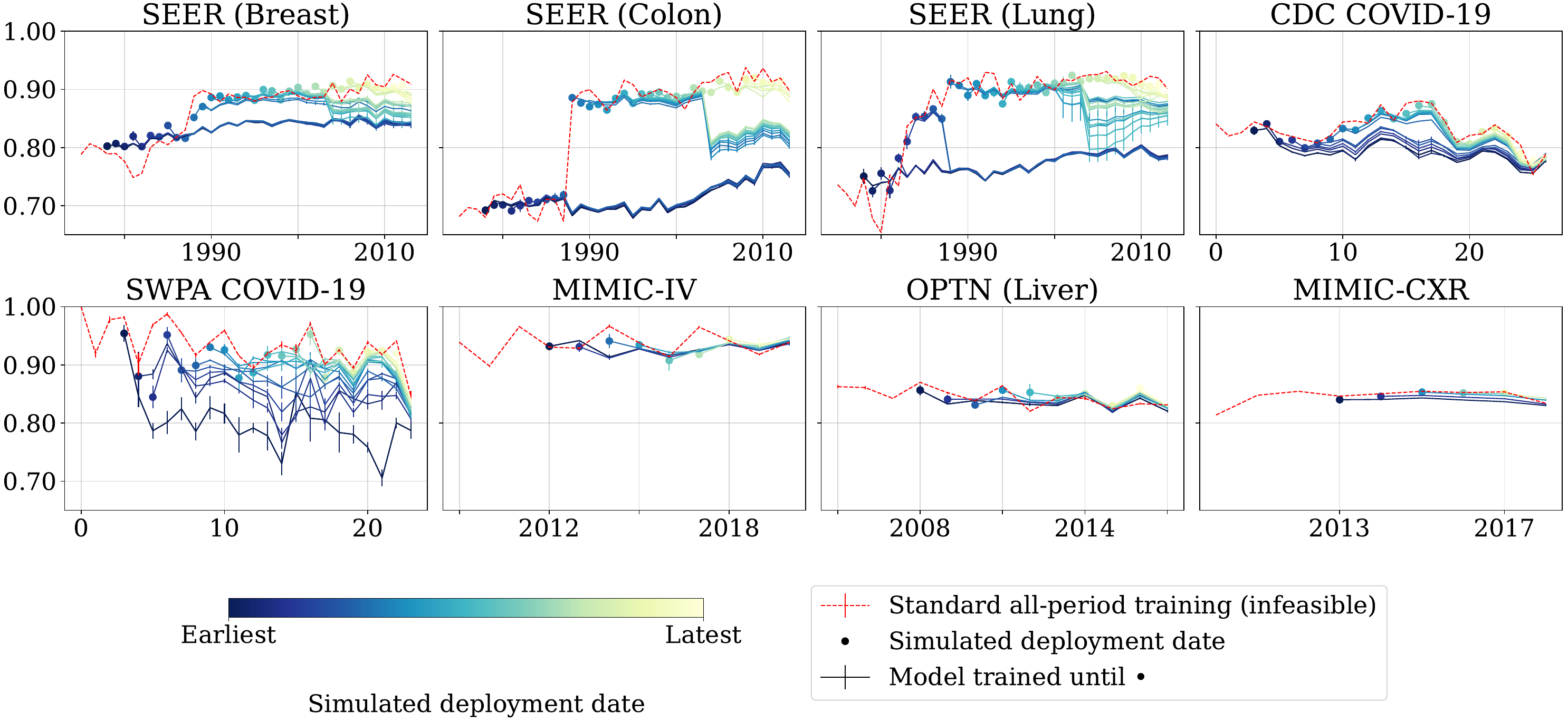}
  \caption{Average test AUROC of logistic regression vs. time. Each solid line 
  gives the performance of a model trained up to a simulated deployment time (marked by a dot), evaluated across future time points. 
  Error bars are 
  $\pm$ standard deviation
  computed over 5 random splits. 
  Red dotted line gives per-timepoint test performance of a model from all-period training (infeasible in reality, as it would involve training on data after the simulated deployment date).
  }
\label{fig:absolute_auc_over_time}
\end{figure*}
\section{Results}

\subsection{All-period Training} 
In standard time-agnostic evaluation, GBDT and MLP achieve the highest average test AUROC on all tabular datasets except MIMIC-IV (Table \ref{tab:AUROC_splitting_by_patient}). 
Note however that LR often has comparable or only slightly lower AUROC than the more complex models.  
The top 10 coefficients of each LR with all-period training are in Appendices \ref{app:seer_data}--\ref{app:optn_data}, and the per-label AUROC of MIMIC-CXR is in Appendix Table \ref{tab:mimic_cxr_label_level_auc}.
To form a baseline for comparison across time, we also evaluate the all-period models on subsets of the all-period test data that belong to each year (red dotted line in Figure \ref{fig:absolute_auc_over_time}), but note that this type of training (on future data) is not feasible in deployment.

\subsection{EMDOT Evaluation}
Figure \ref{fig:absolute_auc_over_time}
plots the AUROC of LR for all tabular datasets (and DenseNet-121 for MIMIC-CXR) over time when using the all-historical training regime. Plots for GBDT and MLP
are in Appendix \ref{app:alternative_metrics}, along with plots for AUPRC. We mainly discuss AUROC, but note that AUPRC observes similar trends as in AUROC. One difference however is that the baseline AUPRC performance is given by the label prevalence (rather than a constant 0.5, as in AUROC), and so observed trends in label prevalence over time appear to influence trends in AUPRC (Appendix Figure \ref{fig:auprc_over_time}).

For both AUROC and AUPRC, the reported test performance of a model from standard all-period training (red dotted line) mostly sits above the performance of any model that could have realistically been deployed by that date. Thus, all-period training tends to provide an over-optimistic estimate of performance upon deployment.

Across the datasets, a variety of trajectories of model performance are observed over time. In the SEER datasets, the AUROC of freshly trained models increases dramatically near 1988, but several of these models experience a large drop
in AUROC 
around 2003 (Figure \ref{fig:absolute_auc_over_time}). Additionally, in-period test AUROCs tend to increase over time.
By contrast, in CDC data, in-sample test AUROCs fluctuate up and down, and model performance over time varies more smoothly, appearing to loosely follow the in-sample performance.
Models trained after December 2020 
have a slight boost in AUROC, 
coinciding with
a surge in cases (and hence sample size, Figure \ref{fig:bar_plot_number_of_samples_pos_outcomes}), however by January 2022 the in-sample AUROC decreases.
In SWPA COVID-19, there is more variation 
and uncertainty 
in AUROC 
early in the pandemic,
where sample sizes are small. 
In December 2020, sample sizes increase, and models seem to become more robust to changes over time.  Finally, in the MIMIC-IV, MIMIC-CXR, and OPTN datasets, AUROC appears relatively stable across time.

\subsection{Training Regime Comparison}
\begin{figure}[!ht]
\centering
  \includegraphics[width=0.6\columnwidth]{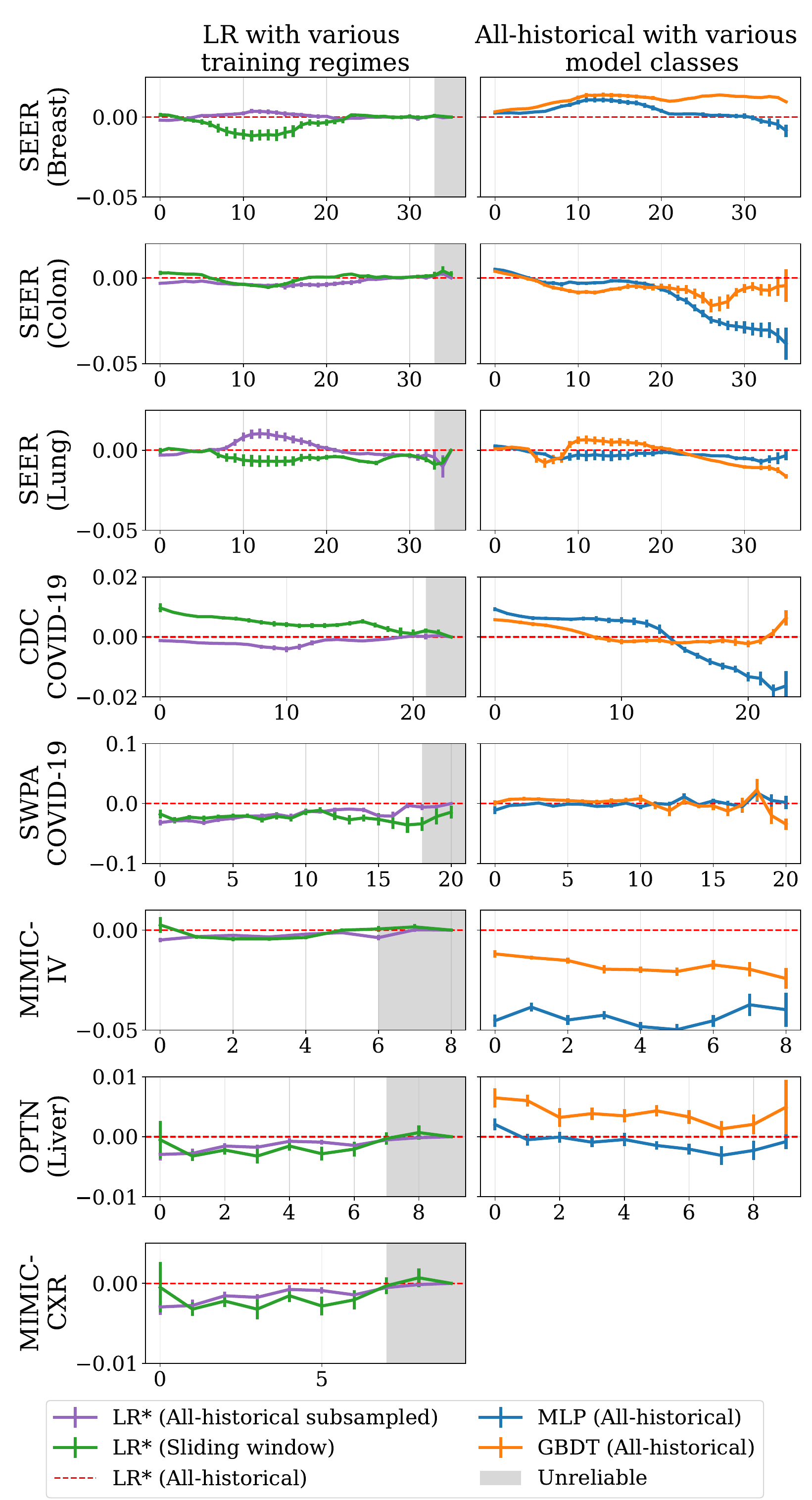}
  \caption{
  $\text{AUROC} - \text{AUROC}_{\text{LR* all-historical}}$ vs. staleness. i.e., AUROC difference relative to a
  LR* all-historical baseline
  across varying stalenesses of data,\protect\footnotemark for different training regimes (left) and model classes (right). Error bars are $\pm$ std. dev. (*in MIMIC-CXR, DenseNet-121 is used instead of LR)}
  \label{fig:window_model_comparison}
\end{figure}

As the staleness of training data increases (i.e. as the test date gets further from the simulated deployment date), different training regimes can fare differently depending on the dataset (Figure \ref{fig:window_model_comparison}, left).

In SEER (Breast) and SEER (Lung), sliding window is initially comparable to all-historical on fresh (low-staleness) data, but significantly underperforms both all-historical and all-historical (subsampled) when data are 8 to 22 years stale.
At larger stalenesses, all training regimes start to become comparable.
In CDC COVID-19, sliding window outperforms %
all-historical
regardless of how stale the data is.
By contrast, in SWPA COVID-19, which has the least amount of data (Table \ref{tab:dataset_info}), both sliding window and all-historical (subsampled) underperform all-historical.
In SEER (Colon),  performance is relatively stable regardless of training regime.
In MIMIC-IV, OPTN (Liver), and MIMIC-CXR, sliding window is on average comparable or slightly outperforms all-historical when staleness is 0, but at nonzero stalenesses all-historical outperforms both sliding window and all-historical subsampled.

\footnotetext{Note: at the largest stalenesses, there are fewer simulated deployment dates being averaged over, and they must be early in the dataset. Here, the sliding window and all-historical can be expected to perform similarly (especially when the sliding window is not much larger than or even matches the history). Since this is an artifact of finite time ranges,
we gray out stalenesses where at least half of the all-historical data 
is the first sliding window of data.}

\subsection{Model Comparison}
In SEER (Breast) and OPTN, %
GBDT outperforms both LR and MLP across the entire time range (Figure \ref{fig:window_model_comparison}, right). 
In SEER (Colon), SEER (Lung), and CDC COVID-19, both GBDT and 
MLP initially outperform 
LR when staleness of the training data is 
less than 4 years, 4 years, and 7 months, respectively,
however both eventually underperform LR as staleness increases further. 
While there is an uptick in GBDT performance on CDC COVID-19 towards 21-month staleness, we note this data point is derived from less data than other points on the line because the data time range is finite. 
In the SWPA COVID-19 dataset, LR, MLP, and GBDT appear to perform comparably over time. In the MIMIC-IV dataset, LR performed best to begin with and remained the best.

\subsection{Detecting Possible Sources of Change}
\begin{figure}[t]
\centering
  \includegraphics[width=0.7\columnwidth]{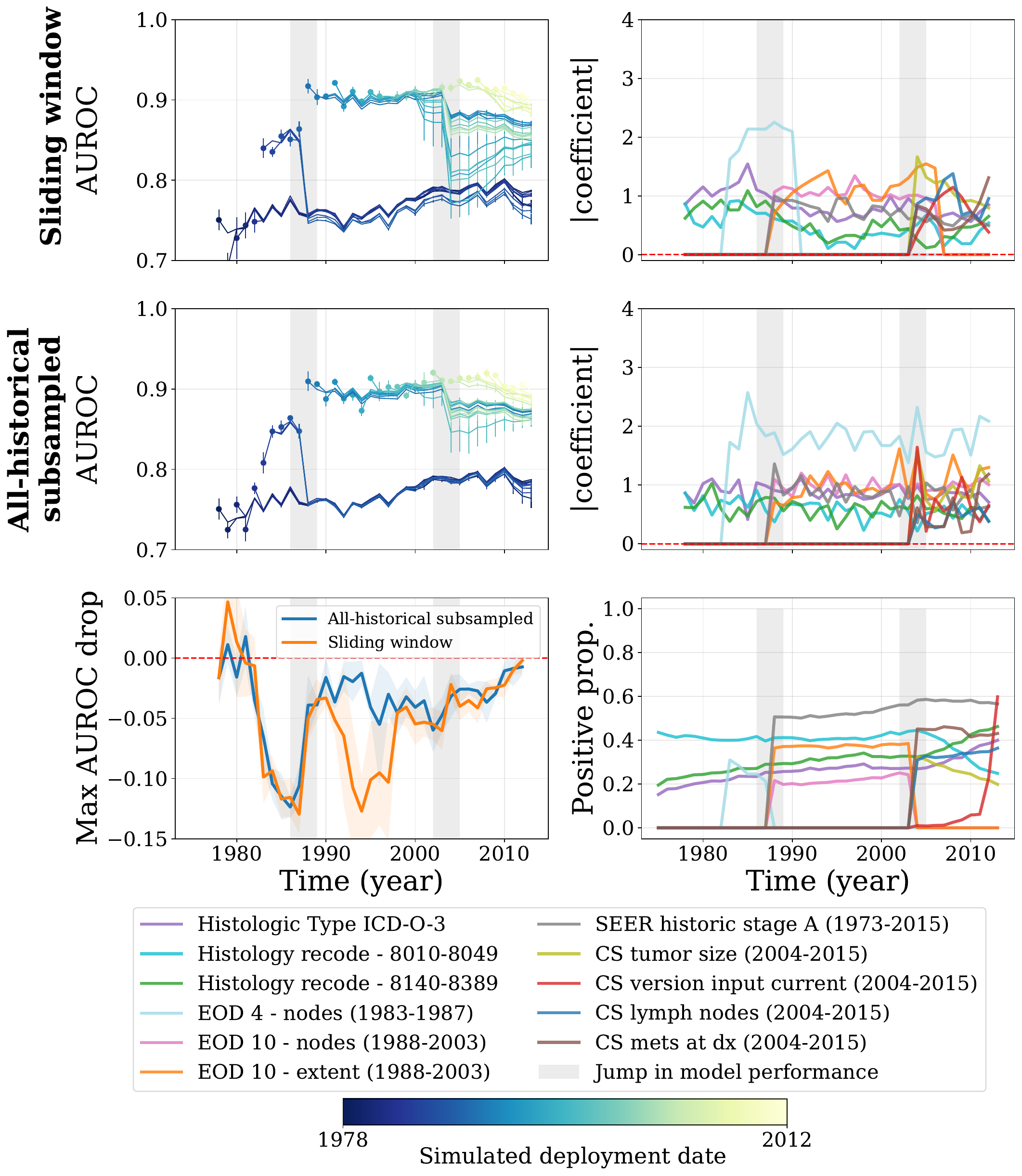}
  \caption{SEER (Lung) diagnostic plots. AUROC vs. time for sliding window (top-left) and all-historical subsampled (mid-left), max.~drop in AUROC for each simulated deployment time (low-left), absolute feature coefficients for LR models from sliding window (top-right) and all-historical subsampled (mid-right) and prevalences of important features over time (low-right).}
  \label{fig:diagnositic_plot_seer_lung_main_paper}
\end{figure}

Diagnostic plots for all datasets are 
in Appendix \ref{app:diagnostic_plot}. Here, we discuss SEER (Lung) (Figure \ref{fig:diagnositic_plot_seer_lung_main_paper})
in detail
as it has several interesting changes in model performance over time.
In 1983, as \verb|EOD 4| features from the extent of disease coding schema are introduced (Figure \ref{fig:diagnositic_plot_seer_lung_main_paper}, bottom right), 
a sudden jump in AUROC occurs (Figure \ref{fig:diagnositic_plot_seer_lung_main_paper}, top and middle left).
However, models trained at this time later experience a large %
AUROC drop (Figure \ref{fig:diagnositic_plot_seer_lung_main_paper}, bottom left).
By 1988, \verb|EOD 4| is phased out, and \verb|EOD 10| features are introduced.  
This coincides with another jump in AUROC, 
sustained until 2003 when the \verb|EOD 10| features are removed. %
In this dataset, the 
all-historical training regime seems more robust to changes over time,
as all-historical models trained after 1988 
avoid the %
drop 
that sliding window models undergo once their window excludes pre-1988 data (Figure \ref{fig:diagnositic_plot_seer_lung_main_paper}, bottom left).

\section{Discussion}
Reported model performance from standard all-period training tends to be over-optimistic (Figure \ref{fig:absolute_auc_over_time}) as models are evaluated on time points already seen in their training set (unrealistic in deployment settings).
Thus, AUROCs reported from all-period training do not capture degradation that would have occurred in deployment.

Comparing model classes, in all datasets except MIMIC-IV, GBDT and MLP slightly outperform 
LR %
under standard time-agnostic evaluation
(Appendix Table \ref{tab:AUROC_splitting_by_patient}). However, evaluated across time, 
LR is often comparable and even outperforms more complex models 
once enough time passes after the simulated deployment date.
For example, MLP achieves the best AUROCs in SEER Breast, Colon, and Lung 
in standard time-agnostic evaluation
(Table \ref{tab:AUROC_splitting_by_patient}). However, 
in evaluation over time,
LR had superior performance once some amount of time (30, 5, 4 years respectively) had passed (Figure \ref{fig:window_model_comparison}, right). In most datasets GBDT appears more robust over time than MLP, however as the training data becomes more stale it tends to become comparable to LR (in all datasets except OPTN Liver and SEER Breast, GBDT dipped below the performance of LR for several stalenesses). Thus, although complex model classes may appear to outperform simpler linear model classes in standard time-agnostic evaluation, one should consider performance over time when selecting a model class for deployment. As demonstrated by the different relative performances of model classes when evaluated over time versus in a time-agnostic manner, EMDOT can serve as a helpful stress-test to combat under-specification.

Regarding training regimes, we find that with increasing stalenesses, all-historical appears more reliable than sliding window across all datasets except for CDC COVID-19 (Figure \ref{fig:window_model_comparison}, left). In SWPA COVID-19, MIMIC-IV, OPTN (Liver), and MIMIC-CXR, the benefit of all-historical data likely comes from the increased sample size, as subsampling all-historical data to be the same size as the corresponding sliding window resulted in comparable performance to sliding window. 
In the SEER datasets, the effect of sample size is less pronounced, as sliding window and subsampled all-historical are frequently comparable to all-historical. There are certain stalenesses for which sliding window underperforms all-historical, which may be due to the addition and removal of features. If the sliding window model learns to rely on recently added features which are later removed, this could result in drops in performance whereas an all-historical model which had learned to predict without the presence of such features would be more robust to such changes. 
On the other hand, in CDC COVID-19 (the setting with the most data and fewest features), subsampled all-historical performs comparably to all-historical, and sliding window outperforms both across all stalenesses (Figure \ref{fig:window_model_comparison}, left). This suggests that the performance of LR may have been saturated even when a sub-sample of all-historical data was used, and the benefit of using more recent data outweighs the larger sample size afforded by all-historical. More broadly, in rapidly evolving environments with simple models, few features, and large quantities of data, the sliding window training regime could be advantageous.

The SEER datasets had dramatic changes in data distribution in both 1988 and 2003, when important features were added and/or removed (Figure \ref{fig:diagnositic_plot_seer_lung_main_paper}). 
One possible reason for the robustness of all-historical models in this dataset is that after 2003, when features like EOD 10 were removed, the model could still rely on features that were introduced prior to the use of EOD 10 in 1988. More broadly, we hypothesize that if a model was trained on a mixture of distributions that occurred throughout the past, it may be better equipped to handle shifts to settings 
similar to those distributions 
in the future.

While the SEER datasets and COVID-19 datasets displayed several changes in model performance over time, the OPTN and MIMIC datasets had relatively stable behavior. One possible reason for this is that the outcomes or diseases of interest were relatively stable in nature, we did not observe any substantial changes in the distribution of data. Another is that in the MIMIC datasets, a three-year range was given for each sample rather than a specific date. This uncertainty around the date, along with the limited number of date ranges, could result in a smoothing effect on the resulting estimates of performance.

In conclusion, EMDOT not only 
yields insights into 
the suitability of different model classes or training regimes for deployment, but also helps one detect distribution shifts that occurred in the past. Understanding such shifts may help practitioners be prepared for shifts of a similar nature in the future. Although the EMDOT framework does require additional computational time than the standard time-agnostic evaluation setup, we argue that the insights that could be gained from this procedure are worthwhile, especially before deployment in high-stakes settings.

\paragraph{Limitations and Future Work} One possible reservation that users might have about using EMDOT is that it could involve training up to $T$ times as many models as would normally be required (where $T$ is number of timepoints). To help alleviate this concern, in future work we plan to implement parallelization in EMDOT. For noisier estimates of model performance in less time, one could also subsample the dataset. Another interesting extension is exploring performance over time in other data modalities (e.g. time series, natural language, etc.). Depending on the complexity of models used in these modalities, this may require additional computational resources. More broadly, we hope that others may also build upon EMDOT to shine new light on how models and methodologies fare when evaluated with an eye towards deployment.

\paragraph{Acknowledgements}
We thank Jamin Chen for his initial work with the SEER datasets. We also thank Minyi Lee for providing helpful clinical context. We gratefully acknowledge the NSF (FAI 2040929 and IIS2211955), Amazon AI, UPMC, Highmark Health, Abridge, Ford, Mozilla, the PwC Center, the Block Center, the Center for Machine Learning and Health, and the CMU Software Engineering Institute (SEI) via Department of Defense contract FA8702-15-D-0002, for their generous support of ACMI Lab’s research.
This material is based upon work supported by the National Science Foundation Graduate
Research Fellowship Program under Grant No. DGE1745016 and DGE2140739. Any opinions, findings,
and conclusions or recommendations expressed in this material are those of the author(s)
and do not necessarily reflect the views of the National Science Foundation.

\newpage
\bibliographystyle{apalike}
\bibliography{refs}

\appendix

\newpage
\onecolumn

\section{Snapshot into the State of ML4H Model Evaluation}
\label{app:review_evals}
To get a snapshot of the current standards for model evaluation in machine learning for healthcare research, we manually reviewed all of the papers from the CHIL 2022 proceedings, the first 20 papers in the CHIL 2021 proceedings, and the first 20 papers that came up in the Radiology medical journal when searching for the keyword ``machine learning'' and filtering for papers from 2022 to 2023 (see README.md in https://github.com/acmi-lab/EvaluationOverTime). Out of 23 papers in the CHIL 2022 proceedings, 21 did not take time into account in their data split, and two were unclear about how they split data, but it is unlikely that they split by time. Out of the 20 papers reviewed at CHIL 2021, only one paper split by time. Out of the 20 papers reviewed from Radiology, 6 did not train or evaluate any machine learning models, but out of the remaining 14 papers, 13 did not take time into account in their data split, and one did not specify how data was split.

\section{EMDOT Python Package}
\label{app:emdot_appendix}

\begin{figure}[H]
  \includegraphics[width=1.05\columnwidth]{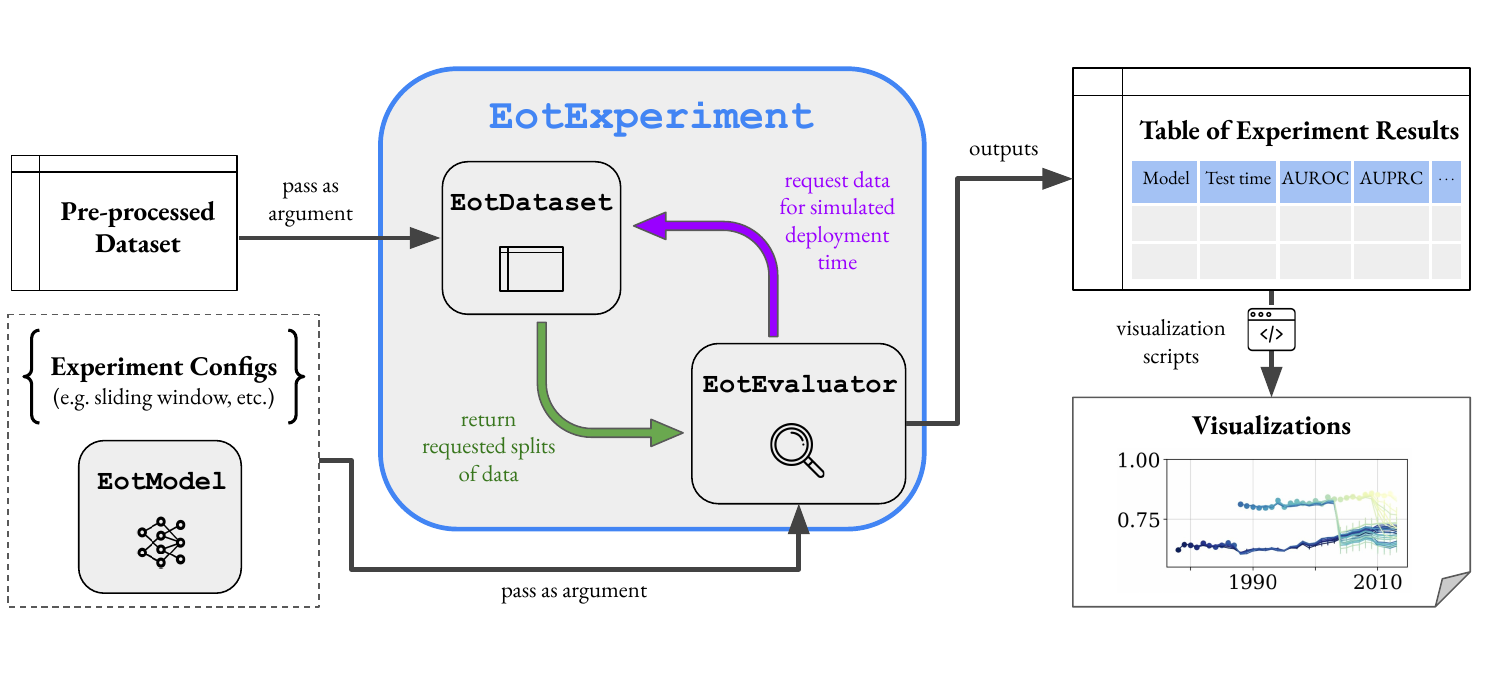}
  \caption{
EMDOT Python package workflow diagram. The primary touchpoint of the EMDOT package is the
EotExperiment
  object. Users provide a dataframe for their (mostly) preprocessed dataset (EMDOT takes care of normalization based on the relevant training set), their desired experiment configuration (e.g. sliding window), and model class (which should subclass the simple
  EotModel abstract class) in order to create an EotExperiment object. Running the run\_experiment() function of the EotExperiment returns a dataframe of experiment results that can then be visualized. The diagram also provides insight into some of the internals of the EotExperiment object -- there is an EotDataset object that handles data splits, and an EotEvaluator object that executes the main evaluation loop.
  }
  \label{fig:emdot_diagram}
\end{figure}

Figure \ref{fig:emdot_diagram} illustrates the workflow of the EMDOT Python package.

\newpage
\section{Additional SEER Data Details}\label{app:seer_data}

\begin{figure}[ht]
  \includegraphics[width=1.0\columnwidth]{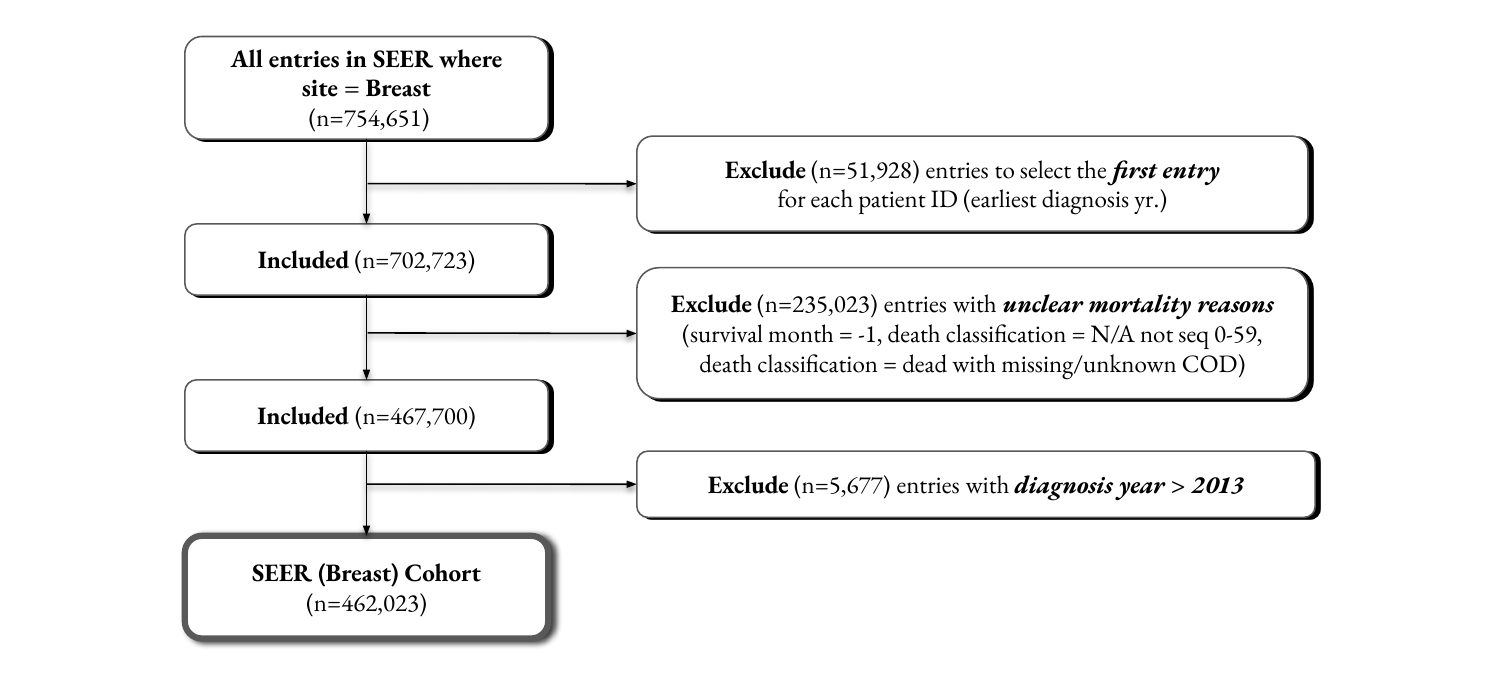}
  \caption{Cohort selection diagram - SEER (Breast)}
  \label{fig:seer_breast_cohort}
\end{figure}

\begin{figure}[ht]
  \includegraphics[width=1.0\columnwidth]{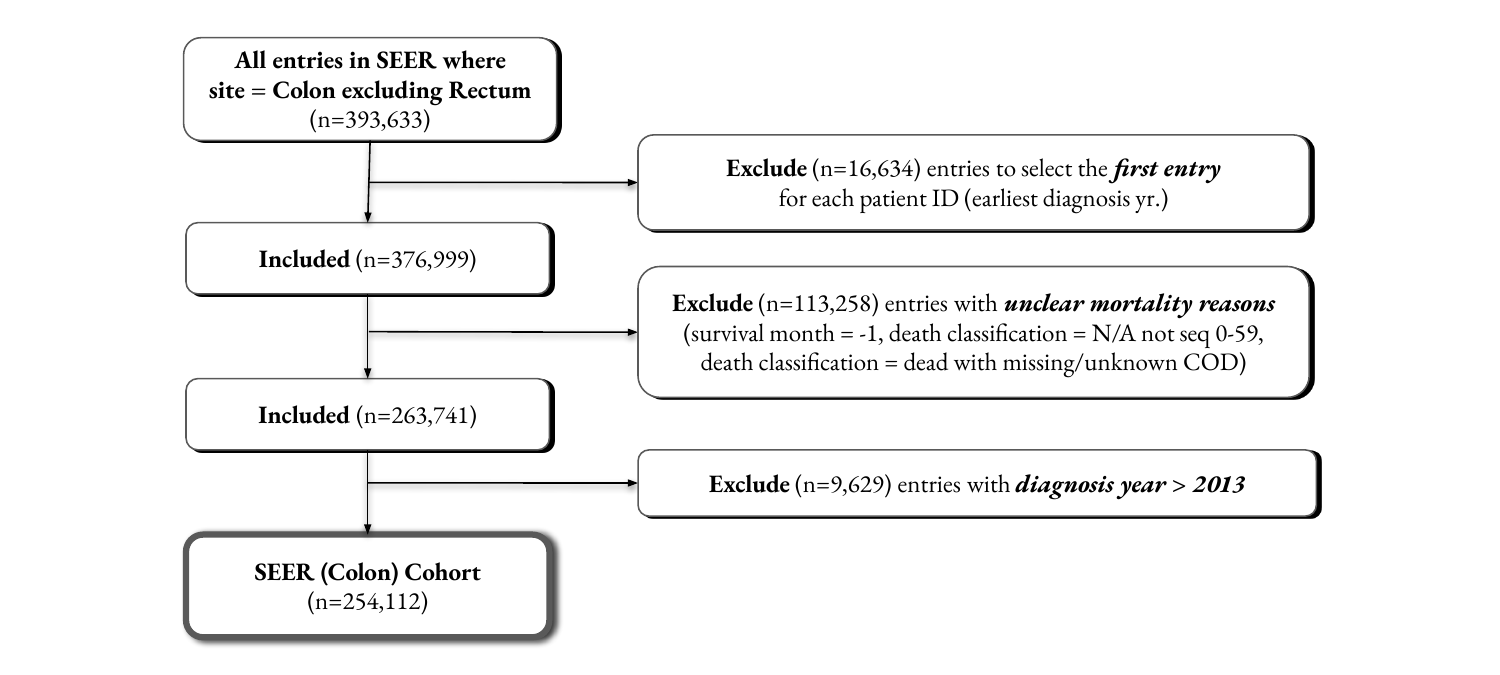}
  \caption{Cohort selection diagram - SEER (Colon)}
  \label{fig:seer_colon_cohort}
\end{figure}

\begin{figure}[ht]
  \includegraphics[width=1.0\columnwidth]{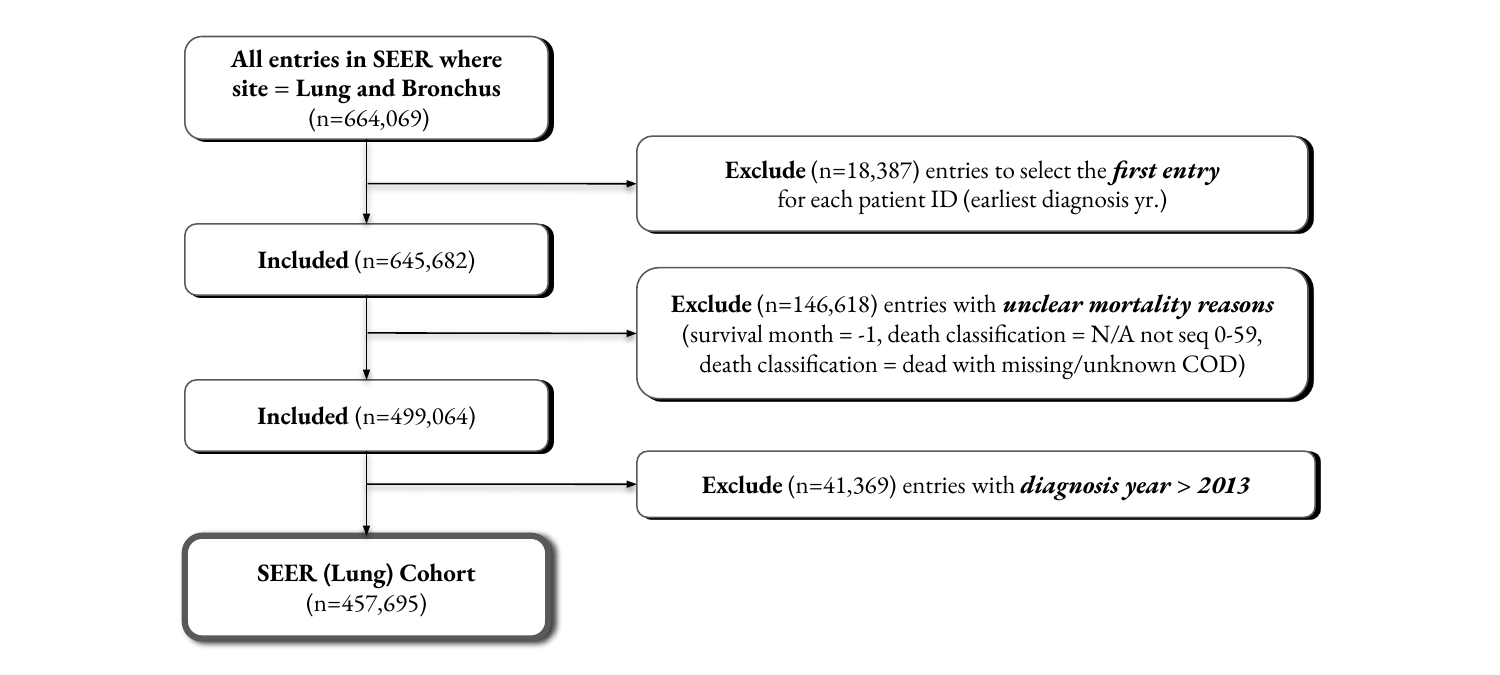}
  \caption{Cohort selection diagram - SEER (Lung)}
  \label{fig:seer_lung_cohort}
\end{figure}

\begin{figure}[ht]
  \includegraphics[width=0.9\columnwidth]{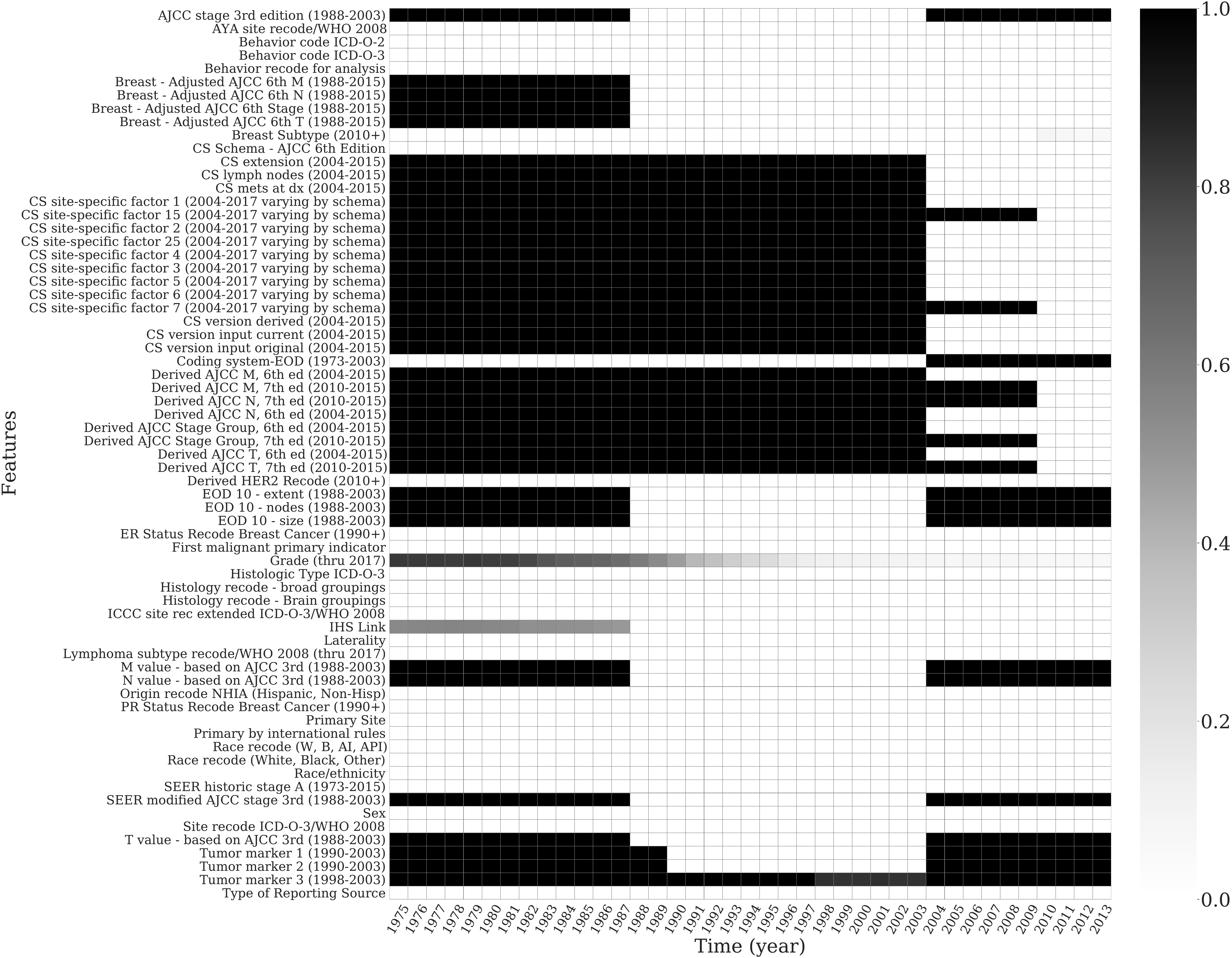}
  \caption{Missingness of categorical features in SEER (Breast).}
  \label{fig:heatmap_seer_breast_cate}
\end{figure}

\begin{figure}[ht]
  \includegraphics[width=0.9\columnwidth]{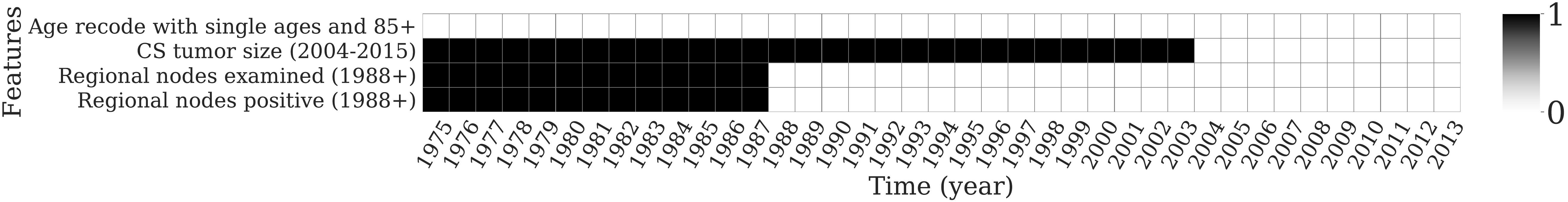}
  \caption{Missingness of numerical features in SEER (Breast).}
  \label{fig:heatmap_seer_breast_num}
  \vspace{2em}
\end{figure}

\begin{figure}[ht]
  \includegraphics[width=0.9\columnwidth]{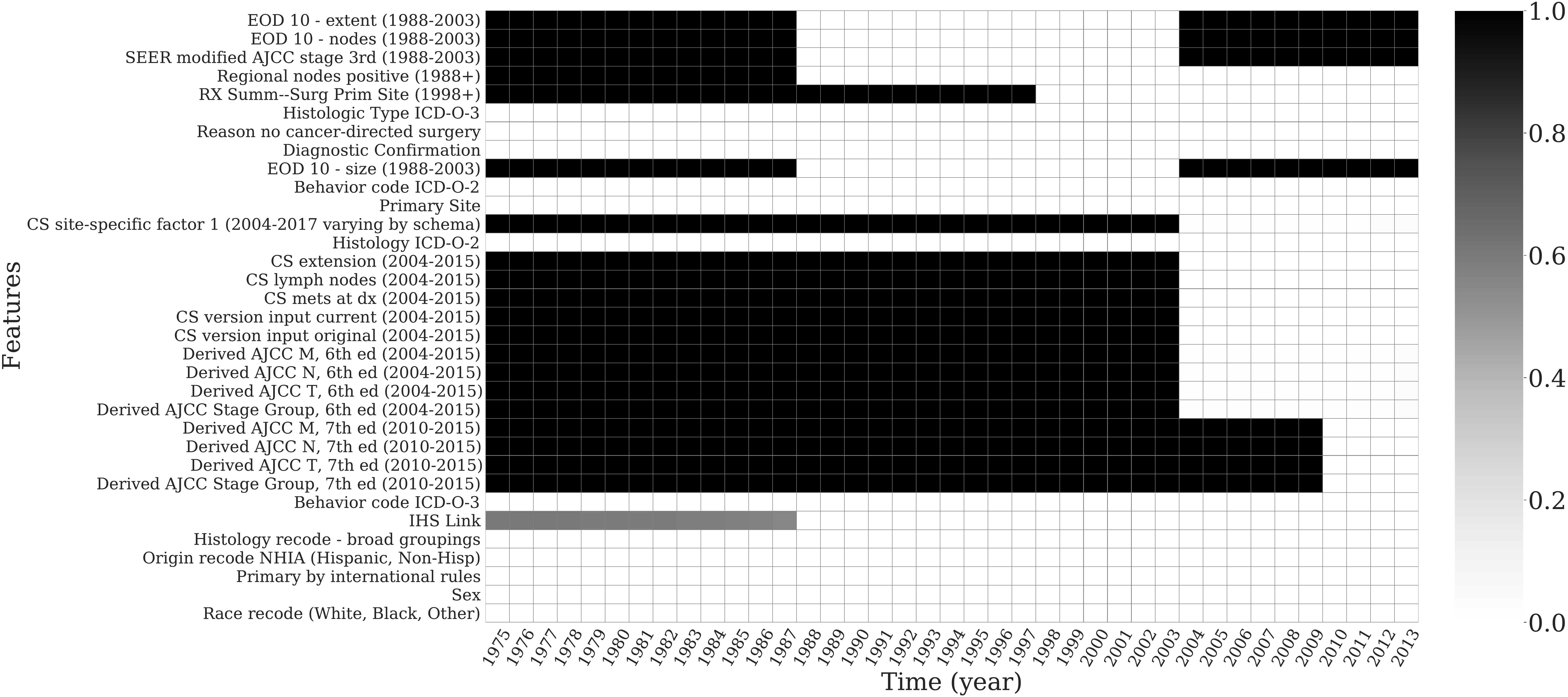}
  \vspace{-1em}
  \caption{Missingness of categorical features in SEER (Colon).}
  \label{fig:heatmap_seer_colon_cate}
\end{figure}

\begin{figure}[ht]
  \includegraphics[width=0.9\columnwidth]{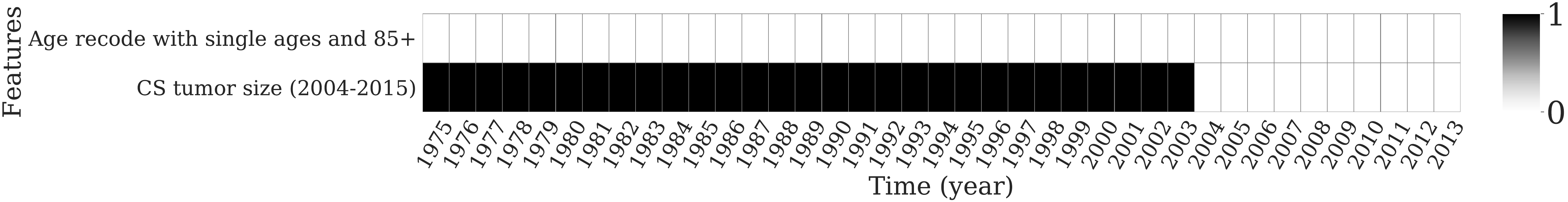}
  \vspace{-1em}
  \caption{Missingness of numerical features in SEER (Colon).}
  \label{fig:heatmap_seer_colon_num}
\end{figure}

\begin{figure}[ht]
  \includegraphics[width=0.9\columnwidth]{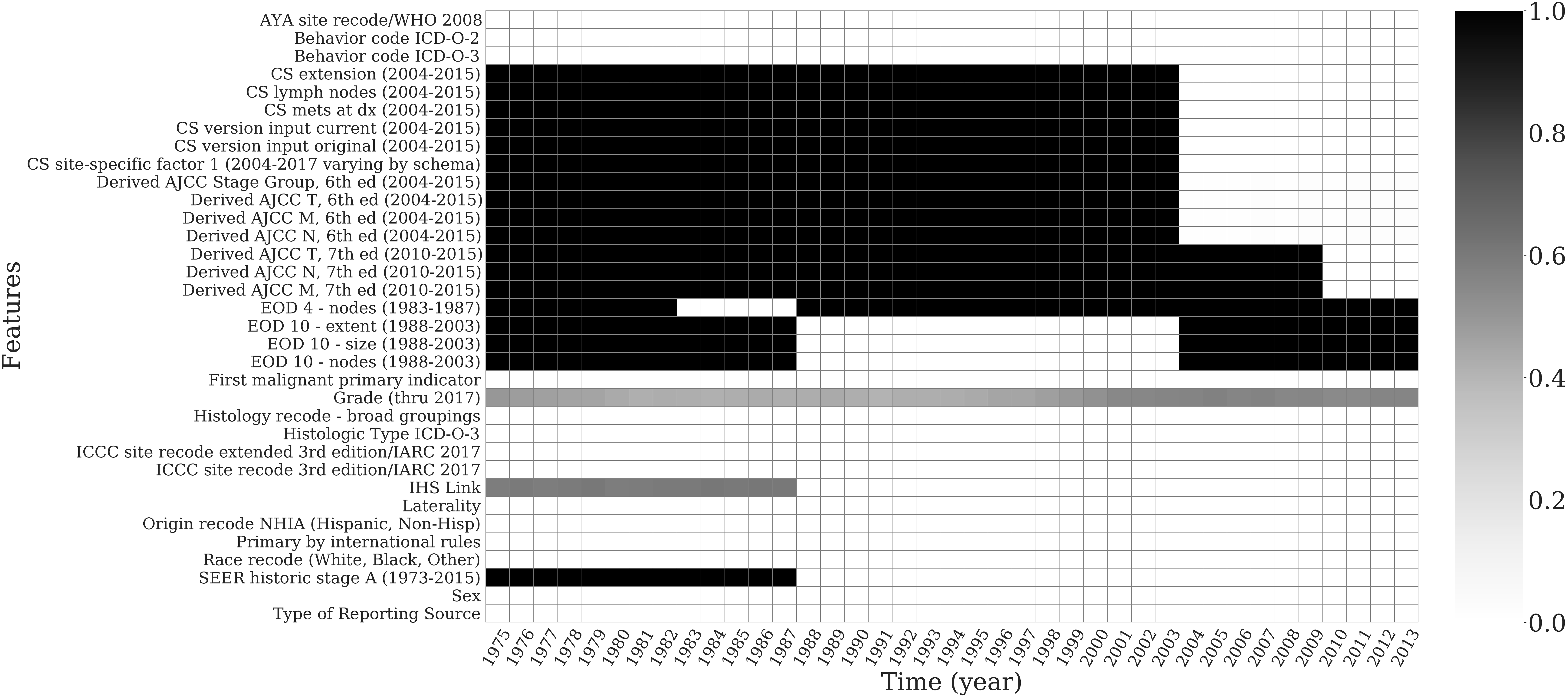}
  \vspace{-1em}
  \caption{Missingness of categorical features in SEER (Lung).}
  \label{fig:heatmap_seer_lung_cate}
\end{figure}

\begin{figure}[ht]
  \includegraphics[width=0.9\columnwidth]{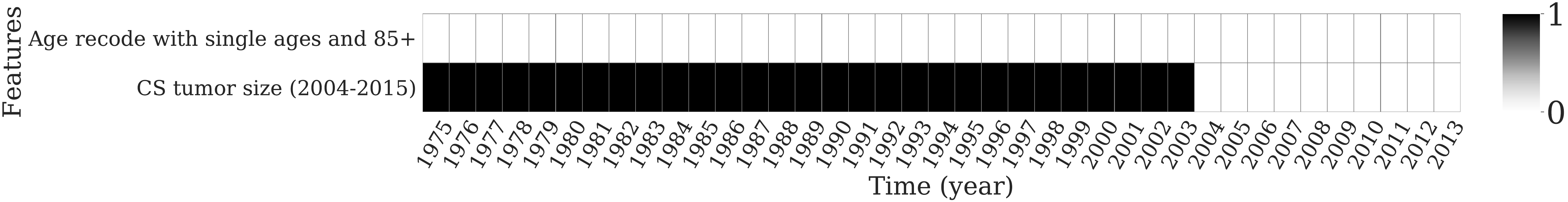}
  \vspace{-1em}
  \caption{Missingness of numerical features in SEER (Lung).}
  \label{fig:heatmap_seer_lung_num}
\end{figure}

The Surveillance, Epidemiology, and End Results (SEER) Program collects cancer incidence data from 
registries
throughout the U.S. 
This data has been used to study survival in several forms of cancer \citep{choi2008conditional,fuller2007conditional,taioli2015determinants,hegselmann2018reproducible}.
Each case includes 
demographics, primary tumor site, tumor morphology, stage and diagnosis, first course of treatment, and survival outcomes
(collected with follow-up) \citep{seerdataset}. 
The performance over time is evaluated on a \emph{yearly} basis. We use the November 2020 version of the SEER database with nine registries (SEER 9), which covers about 9.4\% of the U.S. population. While there are SEER databases that aggregate over more registries and hence cover a greater proportion of the U.S. population, we choose SEER 9 due to the large time range it covers (1975--2018). 
\begin{itemize}
    \item Data access: After filling out a Data Use Agreement and Best Practices Agreement, individuals can easily request access to the SEER dataset.
    \item Cohort selection: Using the SEER$^{*}$Stat software \citep{surveillance2015national}, we define three cohorts of interest: (1) breast cancer, (2) colon cancer, and (3) lung cancer. We primarily follow the cohort selection procedure from \cite{hegselmann2018reproducible}, but we use SEER 9 instead of SEER 18, 
and use data from all available years instead of limiting to 2004--2009. Cohort selection diagrams are given in Figures \ref{fig:seer_breast_cohort}, \ref{fig:seer_colon_cohort}, and \ref{fig:seer_lung_cohort}. If there are multiple samples per patient, we filter to the first entry per patient, which corresponds to when a patient first enters the dataset. This corresponds to a particular interpretation of the prediction: when a patient is first added to a cancer registry, given what we know about that patient, what is their estimated 5-year survival probability? 
    \item Cohort characteristics: Summaries of the SEER (Breast), SEER (Colon), and SEER (Lung) cohort characteristics are in Tables \ref{tab:seer_breast_characteristics}, \ref{tab:seer_colon_characteristics}, and \ref{tab:seer_lung_characteristics}.
    \item Outcome definition: 5-year survival is defined by a confirmation that the patient is alive five years after the year of diagnosis.
    \item Features: We list the features used in the SEER breast, colon, and lung cancer datasets in Section \ref{app:sec_seer_features}.  For all datasets, we convert all categorical variables into dummy features, and apply standard scaling to numerical variables (subtract mean and divide by standard deviation).
    \item Missingness heat maps: are given in Figures 
    \ref{fig:heatmap_seer_breast_cate}, \ref{fig:heatmap_seer_breast_num},
    \ref{fig:heatmap_seer_colon_cate}, \ref{fig:heatmap_seer_colon_num},
    \ref{fig:heatmap_seer_lung_cate}, and \ref{fig:heatmap_seer_lung_num}.
\end{itemize}

\clearpage
\subsection{Cohort Selection and Cohort Characteristics}

\clearpage

\begin{table*}[ht]
\caption{SEER (Breast) cohort characteristics, with count (\%) or median (Q1 -- Q3).}
\vspace{0.5em}
\label{tab:seer_breast_characteristics}
\centering
    \resizebox{0.98\columnwidth}{!}{%
    \begin{tabular}{lccc}
    \toprule
    Characteristic &        & Missingness &       Type \\
    \midrule
    \textbf{Sex} &                         &               &              \\
    \hspace{2em}Female &         459,184 (99.4\%) &          -- &  categorical \\
    \hspace{2em}Male &            2,839 (0.6\%) &          -- &  categorical \\
    \textbf{Age recode with single ages and 85+} &              60 (50-71) &          0.0\% &   continuous \\
    \textbf{Race/ethnicity} &                         &               &              \\
    \hspace{2em}White &         387,247 (83.8\%) &          -- &  categorical \\
    \hspace{2em}Black &           40,217 (8.7\%) &          -- &  categorical \\
    \hspace{2em}Other &           34,559 (7.5\%) &          -- &  categorical \\
    \textbf{Laterality} &                         &               &              \\
    \hspace{2em}Right - origin of primary &         224,777 (48.7\%) &          -- &  categorical \\
    \hspace{2em}Left - origin of primary &         233,549 (50.5\%) &          -- &  categorical \\
    \hspace{2em}Other &            3,697 (0.8\%) &          -- &  categorical \\
    \textbf{Regional nodes positive (1988+)} &                 0 (0-3) &         21.0\% &   continuous \\
    \textbf{T value - based on AJCC 3rd (1988-2003)} &              10 (10-20) &         56.2\% &  categorical \\
    \textbf{Derived AJCC T, 7th ed (2010-2015)} &              13 (13-20) &         85.3\% &  categorical \\
    \textbf{CS site-specific factor 3 (2004-2017 varying by schema)} &                 0 (0-2) &         64.8\% &  categorical \\
    \textbf{Regional nodes examined (1988+)} &                8 (2-15) &         21.0\% &   continuous \\
    \textbf{Coding system-EOD (1973-2003)} &                         &               &              \\
    \hspace{2em}Four-digit EOD (1983-1987) &           44,066 (9.5\%) &          -- &  categorical \\
    \hspace{2em}Ten-digit EOD (1988-2003) &         202,450 (43.8\%) &          -- &  categorical \\
    \hspace{2em}Thirteen-digit (expanded) site specific EOD (1973-1982) &          52,742 (11.4\%) &          -- &  categorical \\
    \hspace{2em}Blank(s) &         162,765 (35.2\%) &          -- &  categorical \\
    \textbf{CS version input original (2004-2015)} &  10,401 (10,300-20,302) &         64.8\% &  categorical \\
    \textbf{CS version input current (2004-2015)} &  20,520 (20,510-20,540) &         64.8\% &  categorical \\
    \textbf{EOD 10 - extent (1988-2003)} &              10 (10-13) &         56.2\% &  categorical \\
    \textbf{Grade (thru 2017)} &                         &               &              \\
    \hspace{2em}Unknown &         130,713 (28.3\%) &          -- &  categorical \\
    \hspace{2em}Moderately differentiated; Grade II &         135,970 (29.4\%) &          -- &  categorical \\
    \hspace{2em}Poorly differentiated; Grade III &         119,900 (26.0\%) &          -- &  categorical \\
    \hspace{2em}Undifferentiated; anaplastic; Grade IV &            8,081 (1.7\%) &          -- &  categorical \\
    \hspace{2em}Well differentiated; Grade I &          67,359 (14.6\%) &          -- &  categorical \\
    \textbf{SEER historic stage A} (1973-2015) &                         &               &              \\
    \hspace{2em}Regional &         136,207 (29.5\%) &          -- &  categorical \\
    \hspace{2em}Localized &         286,927 (62.1\%) &          -- &  categorical \\
    \hspace{2em}Unstaged &            9,242 (2.0\%) &          -- &  categorical \\
    \hspace{2em}Distant &           29,647 (6.4\%) &          -- &  categorical \\
    \textbf{IHS Link} &                         &               &              \\
    \hspace{2em}Record sent for linkage, no IHS match &         409,058 (88.5\%) &          -- &  categorical \\
    \hspace{2em}Record sent for linkage, IHS match &            1,505 (0.3\%) &          -- &  categorical \\
    \hspace{2em}Blank(s) &          51,460 (11.1\%) &          -- &  categorical \\
    \textbf{Histologic Type ICD-O-3} &     8,500 (8,500-8,500) &          0.0\% &  categorical \\
    \textbf{EOD 10 - size (1988-2003)} &              18 (10-30) &         56.2\% &  categorical \\
    \textbf{Type of Reporting Source} &                         &               &              \\
    \hspace{2em}Hospital inpatient/outpatient or clinic &         450,801 (97.6\%) &          -- &  categorical \\
    \hspace{2em}Other &           11,222 (2.4\%) &          -- &  categorical \\
    \textbf{SEER cause-specific death classification} &                         &               &              \\
    \hspace{2em}Alive or dead of other cause &         378,758 (82.0\%) &          -- &  categorical \\
    \hspace{2em}Dead (attributable to this cancer dx) &          83,265 (18.0\%) &          -- &  categorical \\
    \textbf{Survival months} &            135 (74-220) &          0.0\% &  categorical \\
    \textbf{5-year survival} &                         &               &              \\
    \hspace{2em}1 &         378,758 (82.0\%) &          -- &  categorical \\
    \hspace{2em}0 &          83,265 (18.0\%) &          -- &  categorical \\
    \bottomrule
    \end{tabular}%
    }
\end{table*}

\begin{table*}[ht]
\caption{SEER (Colon) cohort characteristics, with count (\%) or median (Q1--Q3).}
\vspace{0.5em}
\label{tab:seer_colon_characteristics}
    \centering
    \begin{tabular}{lccc}
    \toprule
    Characteristic &  & Missingness &       Type \\
    \midrule
    \textbf{Sex} &                         &               &              \\
    \hspace{2em}Female &         133,661 (52.6\%) &          -- &  categorical \\
    \hspace{2em}Male &         120,451 (47.4\%) &          -- &  categorical \\
    \textbf{Age recode with single ages and 85+} &              70 (61-79) &          0.0\% &   continuous \\
    \textbf{Race recode (White, Black, Other)} &                         &               &              \\
    \hspace{2em}White &         212,265 (83.5\%) &          -- &  categorical \\
    \hspace{2em}Black &           24,041 (9.5\%) &          -- &  categorical \\
    \hspace{2em}Other &           17,806 (7.0\%) &          -- &  categorical \\
    \textbf{CS version input current (2004-2015)} &  20,510 (20,510-20,540) &         72.8\% &  categorical \\
    \textbf{Derived AJCC T, 6th ed (2004-2015)} &              30 (20-40) &         73.3\% &  categorical \\
    \textbf{Histology ICD-O-2} &     8,140 (8,140-8,210) &          0.0\% &  categorical \\
    \textbf{IHS Link} &                         &               &              \\
    \hspace{2em}Record sent for linkage, no IHS match &         208,802 (82.2\%) &          -- &  categorical \\
    \hspace{2em}Record sent for linkage, IHS match &              744 (0.3\%) &          -- &  categorical \\
    \hspace{2em}Blank(s) &          44,566 (17.5\%) &          -- &  categorical \\
    \textbf{Histology recode - broad groupings} &                         &               &              \\
    \hspace{2em}8140-8389: adenomas and adenocarcinomas &         213,193 (83.9\%) &          -- &  categorical \\
    \hspace{2em}8440-8499: cystic, mucinous and serous neoplasms &          28,257 (11.1\%) &          -- &  categorical \\
    \hspace{2em}8010-8049: epithelial neoplasms, NOS &            8,797 (3.5\%) &          -- &  categorical \\
    \hspace{2em}Other &            3,865 (1.5\%) &          -- &  categorical \\
    \textbf{Regional nodes positive (1988+)} &                1 (0-10) &         29.8\% &   continuous \\
    \textbf{CS mets at dx (2004-2015)} &                0 (0-22) &         72.8\% &   continuous \\
    \textbf{Reason no cancer-directed surgery} &                         &               &              \\
    \hspace{2em}Surgery performed &         223,929 (88.1\%) &          -- &  categorical \\
    \hspace{2em}Not recommended &           13,003 (5.1\%) &          -- &  categorical \\
    \hspace{2em}Other &           17,180 (6.8\%) &          -- &  categorical \\
    \textbf{Derived AJCC T, 6th ed (2004-2015)} &              30 (20-40) &         73.3\% &  categorical \\
    \textbf{CS version input original (2004-2015)} &  10,401 (10,300-20,302) &         72.8\% &  categorical \\
    \textbf{Primary Site} &           184 (182-187) &          0.0\% &  categorical \\
    \textbf{Diagnostic Confirmation} &                         &               &              \\
    \hspace{2em}Positive histology &         244,616 (96.3\%) &          -- &  categorical \\
    \hspace{2em}Radiography without microscopic confirm &            4,822 (1.9\%) &          -- &  categorical \\
    \hspace{2em}Other &            4,674 (1.8\%) &          -- &  categorical \\
    \textbf{EOD 10 - extent (1988-2003)} &              45 (40-85) &         57.0\% &  categorical \\
    \textbf{Histologic Type ICD-O-3} &     8,140 (8,140-8,210) &          0.0\% &  categorical \\
    \textbf{EOD 10 - size (1988-2003)} &             55 (35-999) &         57.0\% &  categorical \\
    \textbf{CS lymph nodes (2004-2015)} &               0 (0-210) &         72.8\% &  categorical \\
    \textbf{SEER cause-specific death classification} &                         &               &              \\
    \hspace{2em}Dead (attributable to this cancer dx) &         119,047 (46.8\%) &          -- &  categorical \\
    \hspace{2em}Alive or dead of other cause &         135,065 (53.2\%) &          -- &  categorical \\
    \textbf{Survival months} &             68 (12-151) &          0.0\% &  categorical \\
    \textbf{5-year survival} &                         &               &              \\
    \hspace{2em}1 &         135,065 (53.2\%) &          -- &  categorical \\
    \hspace{2em}0 &         119,047 (46.8\%) &          -- &  categorical \\
    \bottomrule
    \end{tabular}%
    \vspace{2em}
\end{table*}

\begin{table*}[ht]
\centering
\caption{SEER (Lung) cohort characteristics, with count (\%) or median (Q1 -- Q3).}
\vspace{0.5em}
\label{tab:seer_lung_characteristics}
    \resizebox{1.0\columnwidth}{!}{%
    \begin{tabular}{lccc}
    \toprule
    Characteristic &  & Missingness &       Type \\
    \midrule
    \textbf{Sex} &                         &               &              \\
    \hspace{2em}Female &         187,967 (41.1\%) &          -- &  categorical \\
    \hspace{2em}Male &         269,728 (58.9\%) &          -- &  categorical \\
    \textbf{Age recode with single ages and 85+} &              68 (60-76) &          0.0\% &   continuous \\
    \textbf{Race recode (White, Black, Other)} &                         &               &              \\
    \hspace{2em}White &         384,184 (83.9\%) &          -- &  categorical \\
    \hspace{2em}Black &          47,237 (10.3\%) &          -- &  categorical \\
    \hspace{2em}Other &           26,274 (5.7\%) &          -- &  categorical \\
    \textbf{Histologic Type ICD-O-3} &     8,070 (8,041-8,140) &          0.0\% &  categorical \\
    \textbf{Laterality} &                         &               &              \\
    \hspace{2em}Left - origin of primary &         178,661 (39.0\%) &          -- &  categorical \\
    \hspace{2em}Right - origin of primary &         245,321 (53.6\%) &          -- &  categorical \\
    \hspace{2em}Paired site, but no information concerning laterality &           23,196 (5.1\%) &          -- &  categorical \\
    \hspace{2em}Other &           10,517 (2.3\%) &          -- &  categorical \\
    \textbf{EOD 10 - nodes (1988-2003)} &                 2 (1-9) &         56.3\% &  categorical \\
    \textbf{EOD 4 - nodes (1983-1987)} &                 3 (0-9) &         88.4\% &  categorical \\
    \textbf{Type of Reporting Source} &                         &               &              \\
    \hspace{2em}Hospital inpatient/outpatient or clinic &         445,606 (97.4\%) &          -- &  categorical \\
    \hspace{2em}Other &           12,089 (2.6\%) &          -- &  categorical \\
    \textbf{SEER historic stage A (1973-2015)} &                         &               &              \\
    \hspace{2em}Regional &          79,409 (17.3\%) &          -- &  categorical \\
    \hspace{2em}Distant &         182,467 (39.9\%) &          -- &  categorical \\
    \hspace{2em}Blank(s) &         123,161 (26.9\%) &          -- &  categorical \\
    \hspace{2em}Localized &          50,375 (11.0\%) &          -- &  categorical \\
    \hspace{2em}Unstaged &           22,283 (4.9\%) &          -- &  categorical \\
    \textbf{CS version input current (2004-2015)} &  20,520 (20,510-20,540) &         70.6\% &  categorical \\
    \textbf{CS mets at dx (2004-2015)} &               23 (0-40) &         70.6\% &   continuous \\
    \textbf{CS version input original (2004-2015)} &  10,401 (10,300-20,302) &         70.6\% &  categorical \\
    \textbf{CS tumor size (2004-2015)} &             50 (29-999) &         70.6\% &  categorical \\
    \textbf{EOD 10 - size (1988-2003)} &             80 (35-999) &         56.3\% &  categorical \\
    \textbf{CS lymph nodes (2004-2015)} &             200 (0-200) &         70.6\% &  categorical \\
    \textbf{Histology recode - broad groupings}&                         &               &              \\
    \hspace{2em}8140-8389: adenomas and adenocarcinomas &         147,127 (32.1\%) &          -- &  categorical \\
    \hspace{2em}8010-8049: epithelial neoplasms, NOS &         179,848 (39.3\%) &          -- &  categorical \\
    \hspace{2em}8440-8499: cystic, mucinous and serous neoplasms &            6,266 (1.4\%) &          -- &  categorical \\
    \hspace{2em}Other &         124,454 (27.2\%) &          -- &  categorical \\
    \textbf{EOD 10 - extent (1988-2003)} &              78 (40-85) &         56.3\% &  categorical \\
    \textbf{SEER cause-specific death classification} &                         &               &              \\
    \hspace{2em}Alive or dead of other cause &          49,997 (10.9\%) &          -- &  categorical \\
    \hspace{2em}Dead (attributable to this cancer dx) &         407,698 (89.1\%) &          -- &  categorical \\
    \textbf{Survival months} &                7 (2-19) &          0.0\% &  categorical \\
    \textbf{5-year survival} &                         &               &              \\
    \hspace{2em}1 &          49,997 (10.9\%) &          -- &  categorical \\
    \hspace{2em}0 &         407,698 (89.1\%) &          -- &  categorical \\
    \bottomrule
    \end{tabular}%
    }  
    \vspace{5em}
\end{table*}

\clearpage

\twocolumn
\subsection{Features}\label{app:sec_seer_features}
\textbf{SEER (Breast):}
{\tiny
\begin{verbatim}
AJCC stage 3rd edition (1988-2003)
AYA site recode/WHO 2008
Age recode with single ages and 85+
Behavior code ICD-O-2
Behavior code ICD-O-3
Behavior recode for analysis
Breast - Adjusted AJCC 6th M (1988-2015)
Breast - Adjusted AJCC 6th N (1988-2015)
Breast - Adjusted AJCC 6th Stage (1988-2015)
Breast - Adjusted AJCC 6th T (1988-2015)
Breast Subtype (2010+)
CS Schema - AJCC 6th Edition
CS extension (2004-2015)
CS lymph nodes (2004-2015)
CS mets at dx (2004-2015)
CS site-specific factor 1 (2004-2017 varying by schema)
CS site-specific factor 15 (2004-2017 varying by schema)
CS site-specific factor 2 (2004-2017 varying by schema)
CS site-specific factor 25 (2004-2017 varying by schema)
CS site-specific factor 3 (2004-2017 varying by schema)
CS site-specific factor 4 (2004-2017 varying by schema)
CS site-specific factor 5 (2004-2017 varying by schema)
CS site-specific factor 6 (2004-2017 varying by schema)
CS site-specific factor 7 (2004-2017 varying by schema)
CS tumor size (2004-2015)
CS version derived (2004-2015)
CS version input current (2004-2015)
CS version input original (2004-2015)
Coding system-EOD (1973-2003)
Derived AJCC M, 6th ed (2004-2015)
Derived AJCC M, 7th ed (2010-2015)
Derived AJCC N, 6th ed (2004-2015)
Derived AJCC N, 7th ed (2010-2015)
Derived AJCC Stage Group, 6th ed (2004-2015)
Derived AJCC Stage Group, 7th ed (2010-2015)
Derived AJCC T, 6th ed (2004-2015)
Derived AJCC T, 7th ed (2010-2015)
Derived HER2 Recode (2010+)
EOD 10 - extent (1988-2003)
EOD 10 - nodes (1988-2003)
EOD 10 - size (1988-2003)
ER Status Recode Breast Cancer (1990+)
First malignant primary indicator
Grade (thru 2017)
Histologic Type ICD-O-3
Histology recode - Brain groupings
Histology recode - broad groupings
ICCC site rec extended ICD-O-3/WHO 2008
IHS Link
Laterality
Lymphoma subtype recode/WHO 2008 (thru 2017)
M value - based on AJCC 3rd (1988-2003)
N value - based on AJCC 3rd (1988-2003)
Origin recode NHIA (Hispanic, Non-Hisp)
PR Status Recode Breast Cancer (1990+)
Primary Site
Primary by international rules
Race recode (W, B, AI, API)
Race recode (White, Black, Other)
Race/ethnicity
Regional nodes examined (1988+)
Regional nodes positive (1988+)
SEER historic stage A (1973-2015)
SEER modified AJCC stage 3rd (1988-2003)
Sex
Site recode ICD-O-3/WHO 2008
T value - based on AJCC 3rd (1988-2003)
Tumor marker 1 (1990-2003)
Tumor marker 2 (1990-2003)
Tumor marker 3 (1998-2003)
Type of Reporting Source
\end{verbatim}}
\textbf{SEER (Colon):}
{\tiny
\begin{verbatim}
Age recode with <1 year olds
Age recode with single ages and 85+
Behavior code ICD-O-2
Behavior code ICD-O-3
CS extension (2004-2015)
CS lymph nodes (2004-2015)
CS mets at dx (2004-2015)
CS site-specific factor 1 (2004-2017 varying by schema)
CS tumor size (2004-2015)
CS version input current (2004-2015)
CS version input original (2004-2015)
Derived AJCC M, 6th ed (2004-2015)
Derived AJCC M, 7th ed (2010-2015)
Derived AJCC N, 6th ed (2004-2015)
Derived AJCC N, 7th ed (2010-2015)
Derived AJCC Stage Group, 6th ed (2004-2015)
Derived AJCC Stage Group, 7th ed (2010-2015)
Derived AJCC T, 6th ed (2004-2015)
Derived AJCC T, 7th ed (2010-2015)
Diagnostic Confirmation
EOD 10 - extent (1988-2003)
EOD 10 - nodes (1988-2003)
EOD 10 - size (1988-2003)
Histologic Type ICD-O-3
Histology ICD-O-2
Histology recode - broad groupings
IHS Link
Origin recode NHIA (Hispanic, Non-Hisp)
Primary Site
Primary by international rules
RX Summ--Surg Prim Site (1998+)
Race recode (White, Black, Other)
Reason no cancer-directed surgery
Regional nodes positive (1988+)
SEER modified AJCC stage 3rd (1988-2003)
Sex
\end{verbatim}}
\textbf{SEER (Lung):}
{\tiny
\begin{verbatim}
AYA site recode/WHO 2008
Age recode with <1 year olds
Age recode with single ages and 85+
Behavior code ICD-O-2
Behavior code ICD-O-3
CS extension (2004-2015)
CS lymph nodes (2004-2015)
CS mets at dx (2004-2015)
CS site-specific factor 1 (2004-2017 varying by schema)
CS tumor size (2004-2015)
CS version input current (2004-2015)
CS version input original (2004-2015)
Derived AJCC M, 6th ed (2004-2015)
Derived AJCC M, 7th ed (2010-2015)
Derived AJCC N, 6th ed (2004-2015)
Derived AJCC N, 7th ed (2010-2015)
Derived AJCC Stage Group, 6th ed (2004-2015)
Derived AJCC T, 6th ed (2004-2015)
Derived AJCC T, 7th ed (2010-2015)
EOD 10 - extent (1988-2003)
EOD 10 - nodes (1988-2003)
EOD 10 - size (1988-2003)
EOD 4 - nodes (1983-1987)
First malignant primary indicator
Grade (thru 2017)
Histologic Type ICD-O-3
Histology recode - broad groupings
ICCC site recode 3rd edition/IARC 2017
ICCC site recode extended 3rd edition/IARC 2017
IHS Link
Laterality
Origin recode NHIA (Hispanic, Non-Hisp)
Primary by international rules
Race recode (White, Black, Other)
SEER historic stage A (1973-2015)
Sex
Type of Reporting Source
\end{verbatim}}
\onecolumn

\subsection{Missingness heatmaps}

This section plots missingness heatmaps of categorical and numerical features in each SEER dataset over time. Darker color means larger proportion of missing data.


\clearpage

\section{Additional CDC COVID-19 Data Details}\label{app:cdc_data}
The COVID-19 Case Surveillance Detailed Data \citep{cdc_data} is a national, publicly available dataset provided by the CDC. 
It contains 33 elements, with patient-level data including symptoms, demographics, and state of residence. The performance over time is evaluated on a \emph{monthly} basis. We use the version the released on June 6th, 2022. Disclaimer: ``The CDC does not take responsibility for the scientific validity or accuracy of methodology, results, statistical analyses, or conclusions presented.''

\begin{figure}[!h]
  \includegraphics[width=\columnwidth]{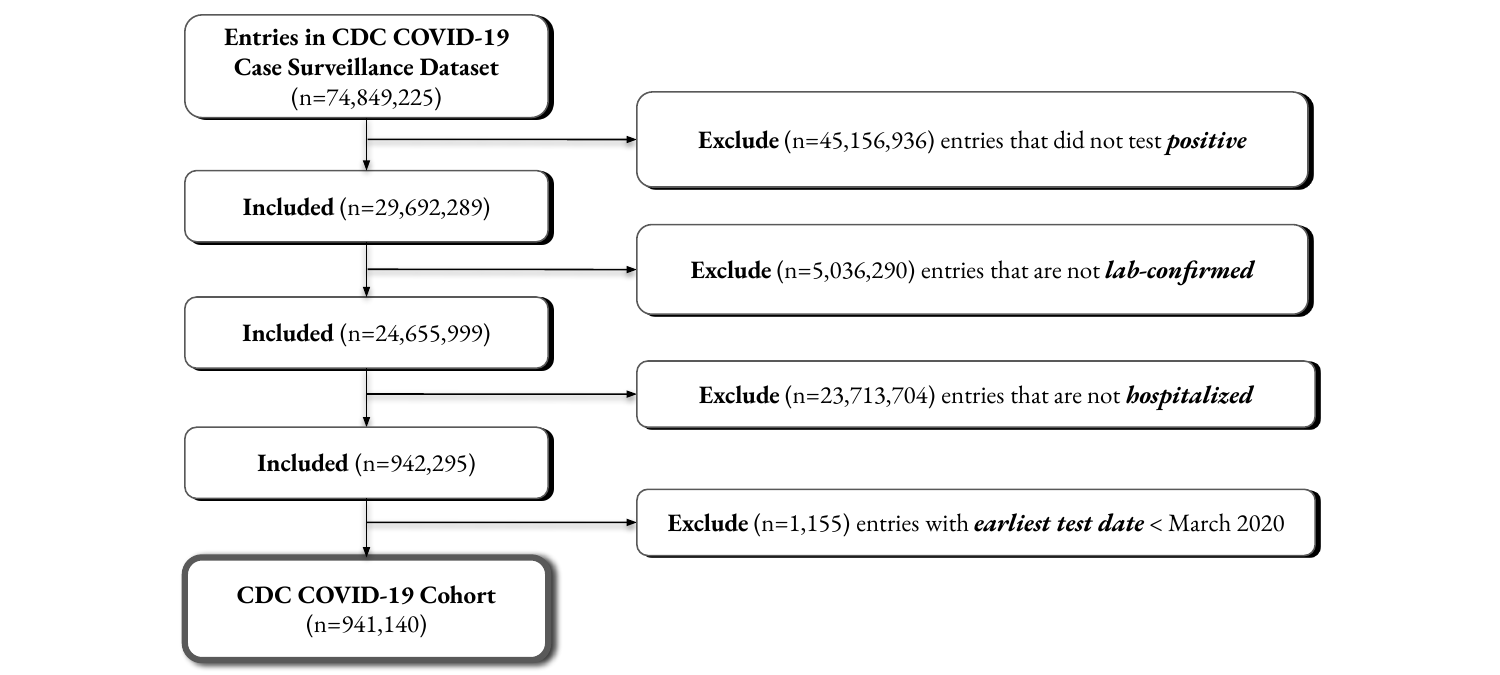}
  \vspace{-1em}
  \caption{Cohort selection diagram - CDC COVID-19}
  \label{fig:cdc_covid_cohort}
\end{figure}

\begin{figure}[!h]
  \includegraphics[width=1.0\columnwidth]{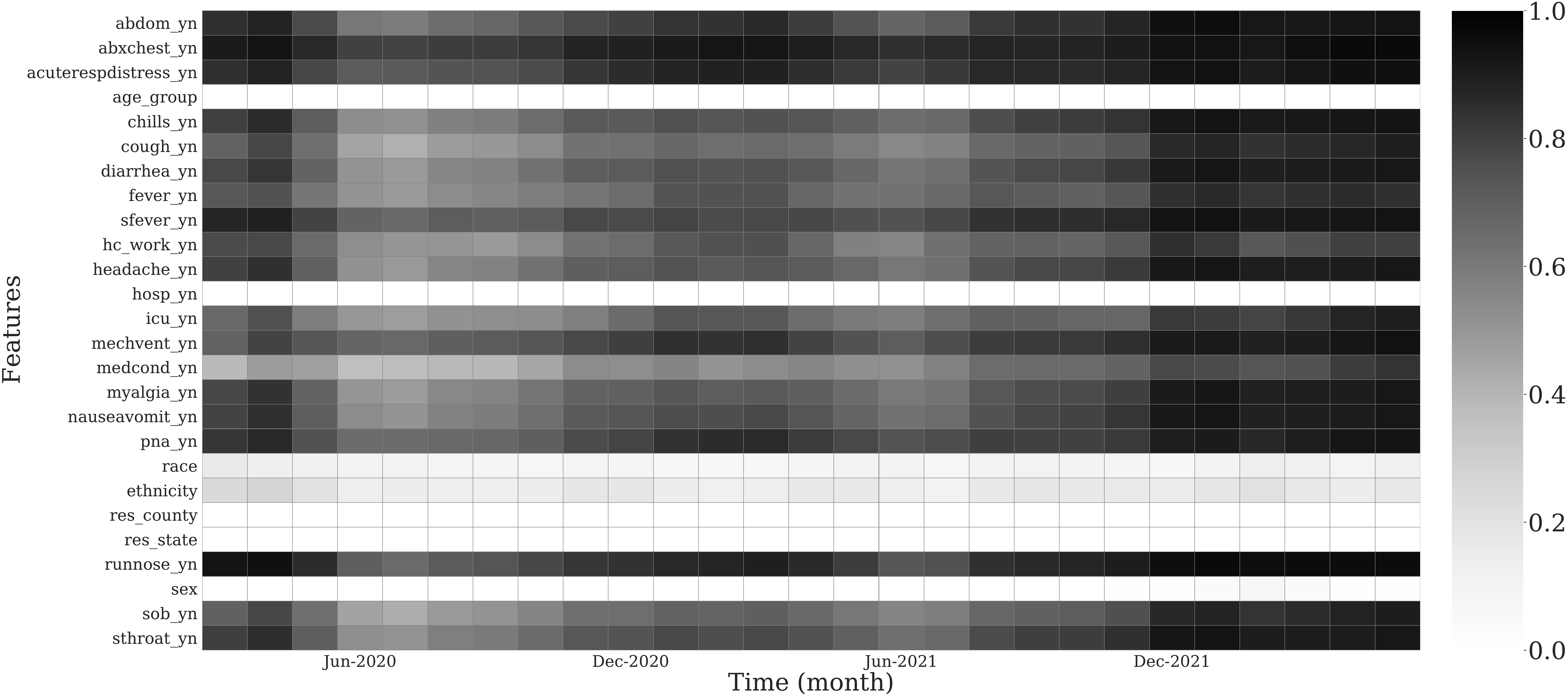}
  \caption{Missingness over time for features in CDC COVID-19 dataset after cohort selection. The darker the color, the larger the proportion of missing data.}
  \label{fig:heatmap_cdc_covid}
\end{figure}

\begin{figure}[H]
\centering
  {%
    \subfigure[By age group]{\label{fig:stack_age_group}%
      \includegraphics[width=0.5\columnwidth]{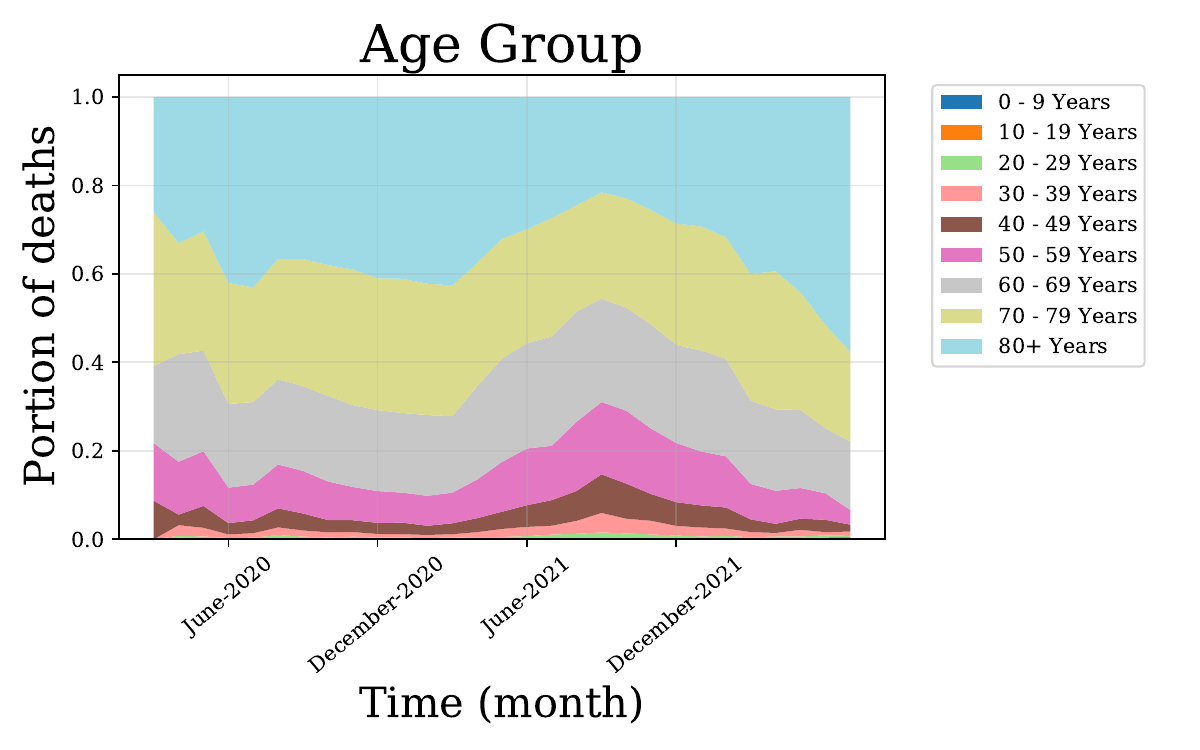}}%
    \subfigure[By state of residence]{\label{fig:stack_state_residence}%
      \includegraphics[width=0.5\columnwidth]{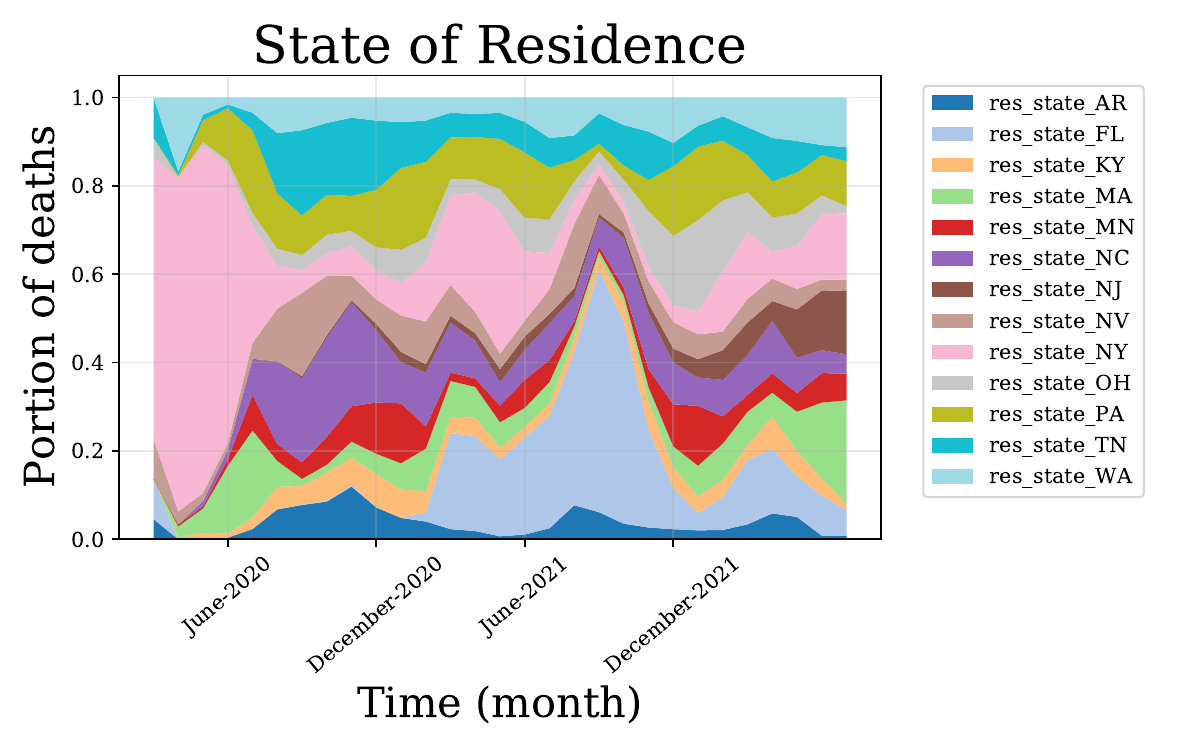}}%
  }
  \caption{Proportion of deaths over time for each age group and state of residence.}
  \label{fig:cdc_stack_area_plot}
\end{figure}

\begin{itemize}
    \item Data access: To access the data, users must complete a registration information and data use restrictions agreement (RIDURA).
    \item Cohort selection: The cohort consists of all patients who were lab-confirmed positive for COVID-19, had a non-null positive specimen date, and were hospitalized (\verb|hosp_yn = Yes|). Cohort selection diagrams are given in Figures \ref{fig:cdc_covid_cohort}
    \item Cohort characteristics: Cohort characteristics are given in Table \ref{tab:cdc_covid_characteristics}.
    \item Outcome definition: mortality, defined by \verb|death_yn = Yes|
    \item Features: We list the features used in the CDC COVID-19 datasets in Section \ref{app:sec_cdc_features}.
    We convert all categorical variables into dummy features, and apply standard scaling to numerical variables (subtract mean and divide by standard deviation).
    \item Missingness heat map: is given in Figure \ref{fig:heatmap_cdc_covid}. 
    \item Additionally, we provide stacked area plots showing how the distribution of ages (Figure \ref{fig:stack_age_group} and states \ref{fig:stack_state_residence} shifts over time.
\end{itemize}
\clearpage

\subsection{Cohort Selection and Cohort Characteristics}
\vspace{-1em}

\clearpage

\begin{table*}[htb]
\centering
\caption{CDC COVID-19 cohort characteristics, with count (\%) or median (Q1--Q3).}
\label{tab:cdc_covid_characteristics}
    \begin{tabular}{lccc}
    \toprule
    Characteristic &  & Missingness &       Type \\
    \midrule
    \textbf{Sex} &                   &               &              \\
    \hspace{2em}Female &   455,376 (48.4\%) &          -- &  categorical \\
    \hspace{2em}Male &   475,223 (50.5\%) &          -- &  categorical \\
    \hspace{2em}Unknown/Missing &      10,541 (1.1\%) &          -- &  categorical \\
    \textbf{Age Group} &                   &               &              \\
    \hspace{2em}0 - 9  &     16,373 (1.7\%) &          -- &  categorical \\
    \hspace{2em}10 - 19  &     17,252 (1.8\%) &          -- &  categorical \\
    \hspace{2em}20 - 29  &     48,505 (5.2\%) &          -- &  categorical \\
    \hspace{2em}30 - 39  &     71,776 (7.6\%) &          -- &  categorical \\
    \hspace{2em}40 - 49  &     88,531 (9.4\%) &          -- &  categorical \\
    \hspace{2em}50 - 59  &   141,805 (15.1\%) &          -- &  categorical \\
    \hspace{2em}60 - 69  &   189,354 (20.1\%) &          -- &  categorical \\
    \hspace{2em}70 - 79  &   189,018 (20.1\%) &          -- &  categorical \\
    \hspace{2em}80+  &   177,765 (18.9\%) &          -- &  categorical \\
    \hspace{2em}Missing &        761 (0.1\%) &          -- &  categorical \\
    \textbf{Race} &                   &               &              \\
    \hspace{2em}White &   544,199 (57.8\%) &          -- &  categorical \\
    \hspace{2em}Black &   173,847 (18.5\%) &          -- &  categorical \\
    \hspace{2em}Other &   205,547 (21.8\%) &          -- &  categorical \\
    \textbf{State of Residence} &                   &               &              \\
    \hspace{2em}NY &   189,684 (20.2\%) &          -- &  categorical \\
    \hspace{2em}OH &     70,097 (7.4\%) &          -- &  categorical \\
    \hspace{2em}FL &     35,679 (3.8\%) &          -- &  categorical \\
    \hspace{2em}WA &     58,854 (6.3\%) &          -- &  categorical \\
    \hspace{2em}MA &     31,441 (3.3\%) &          -- &  categorical \\
    \hspace{2em}Other &   555,353 (59.0\%) &          -- &  categorical \\
    \textbf{Mechanical Ventilation} &                   &               &              \\
    \hspace{2em}Yes &     38,009 (4.0\%) &          -- &  categorical \\
    \hspace{2em}No &   138,331 (14.7\%) &          -- &  categorical \\
    \hspace{2em}Unknown/Missing &   764,800 (81.2\%) &          -- &  categorical \\
    \textbf{Mortality} &                   &               &              \\
    \hspace{2em}1 &   190,786 (20.3\%) &          -- &  categorical \\
    \hspace{2em}0 &   750,354 (79.7\%) &          -- &  categorical \\
    \bottomrule
    \end{tabular}%
\end{table*}
\clearpage

\subsection{Features}\label{app:sec_cdc_features}
{\small \begin{verbatim}abdom_yn, abxchest_yn, acuterespdistress_yn, age_group, chills_yn, 
cough_yn, diarrhea_yn, ethnicity, fever_yn, hc_work_yn, headache_yn, 
hosp_yn, icu_yn, mechvent_yn, medcond_yn, month, myalgia_yn, 
nauseavomit_yn, pna_yn, race, relative_month, res_county, res_state, 
runnose_yn, sex, sfever_yn, sob_yn, sthroat_yn \end{verbatim}}

\subsection{Missingness heatmaps}

\subsection{Additional Figures}
\clearpage

\section{Additional SWPA COVID-19 Data Details}\label{app:swpa_data}
The Southwestern Pennsylvania (SWPA) COVID-19 dataset consists of EHR data from patients tested for COVID-19. It was collected by a major healthcare provider in SWPA, and includes
patient demographics, labs, problem histories, medications, inpatient vs. outpatient status, and other information collected in the patient encounter. The performance over time is evaluated on a \emph{monthly} basis.

\begin{figure}[!h]
  \includegraphics[width=1.0\columnwidth]{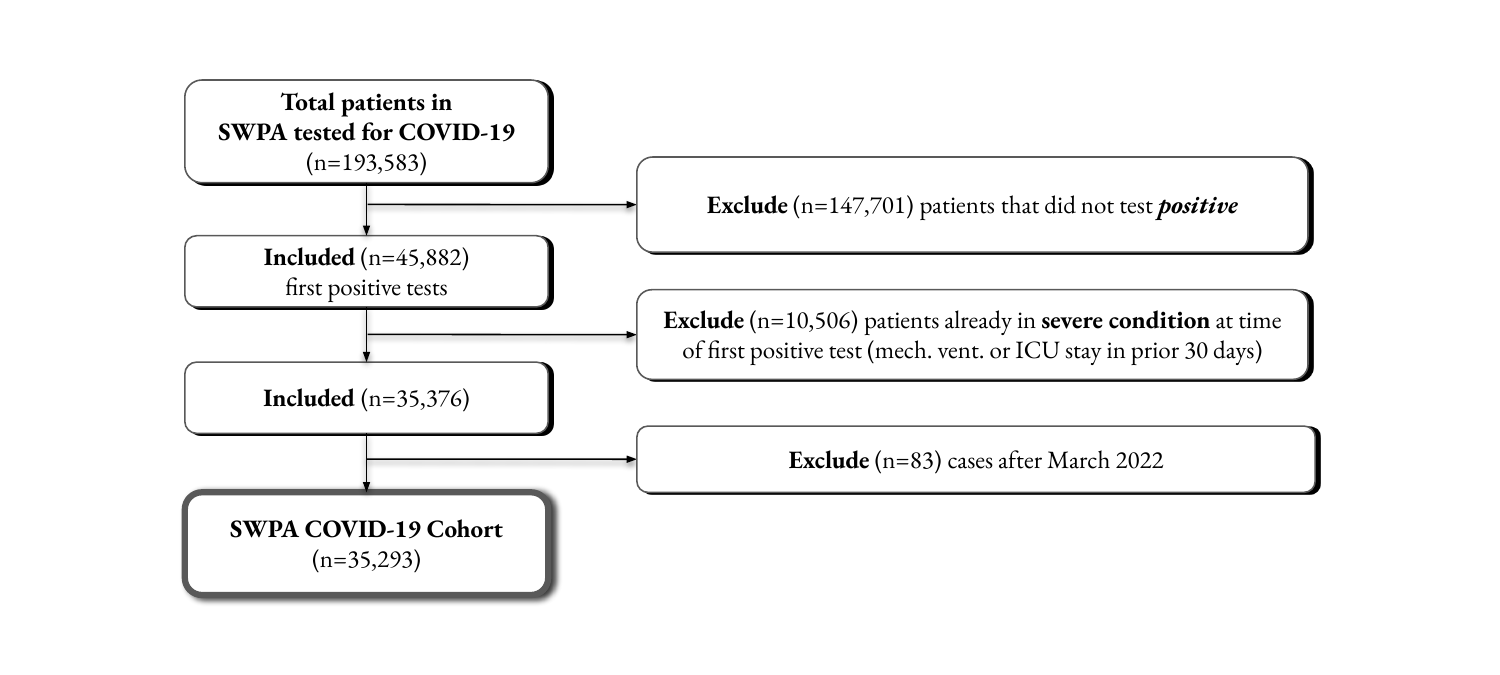}
  \caption{Cohort selection diagram - SWPA COVID-19}
  \label{fig:swpa_covid_cohort}
\end{figure}

\begin{figure}[H]
    \includegraphics[width=1.0\columnwidth]{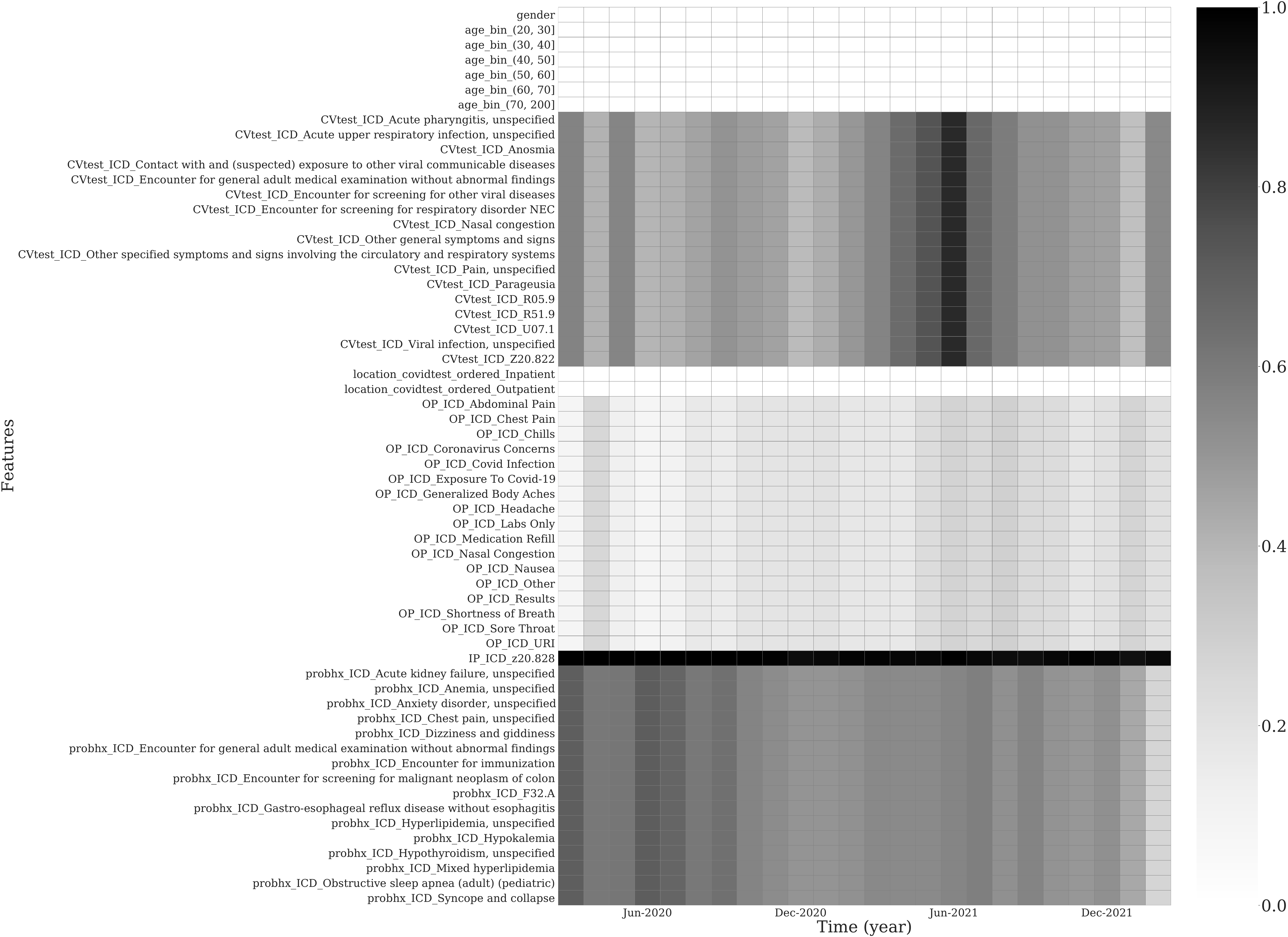}
  \caption{Missingness of categorical features in SWPA COVID-19 dataset (part 1).}
  \label{fig:heatmap_pa_covid_cate_1}
\end{figure}

\begin{figure}[H]
  \includegraphics[width=1.0\columnwidth]{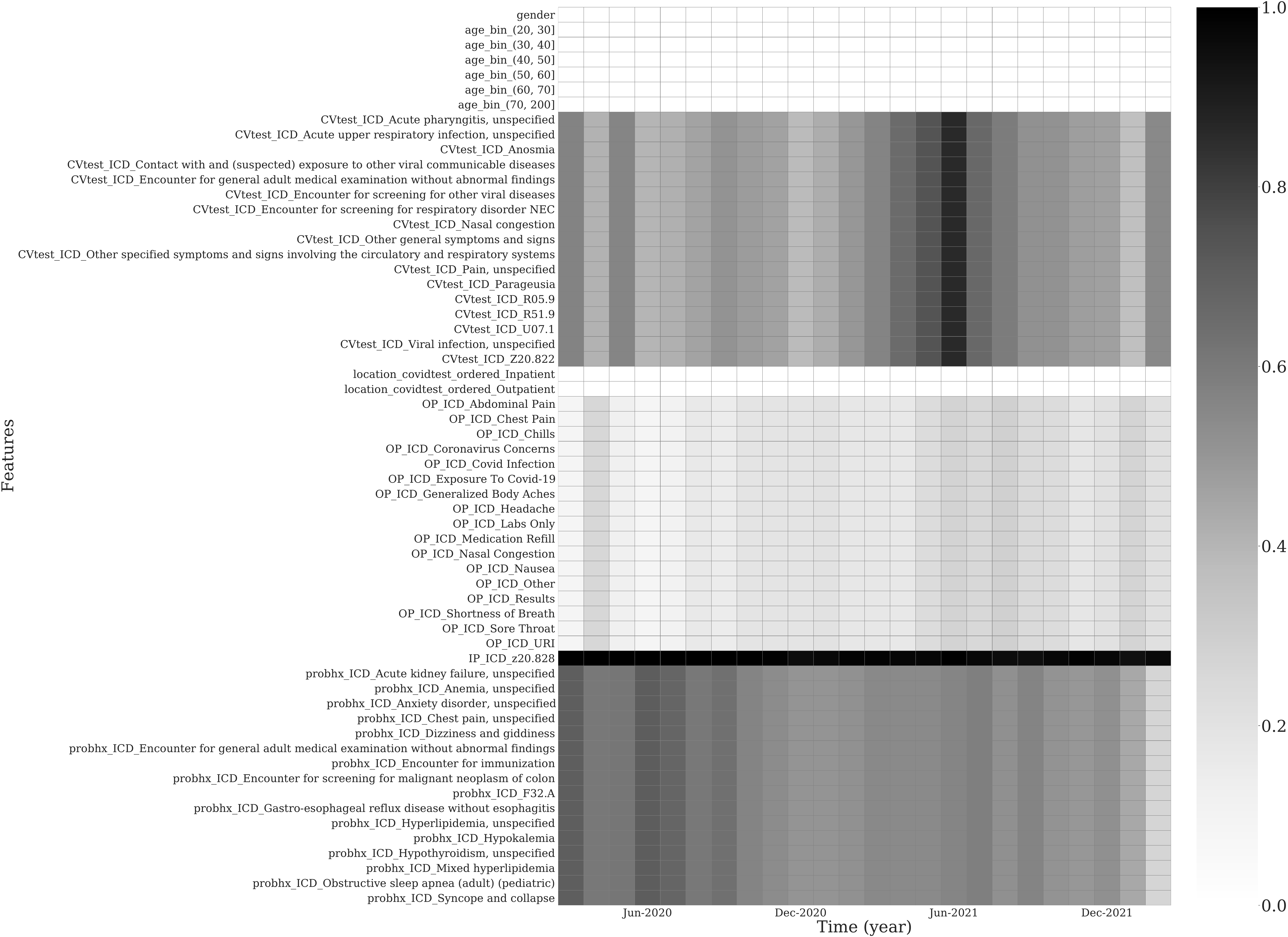}
  \caption{Missingness of categorical features in SWPA COVID-19 dataset (part 2).}
  \label{fig:heatmap_pa_covid_cate_2}
\end{figure}

\begin{figure}[H]
  \includegraphics[width=1.0\columnwidth]{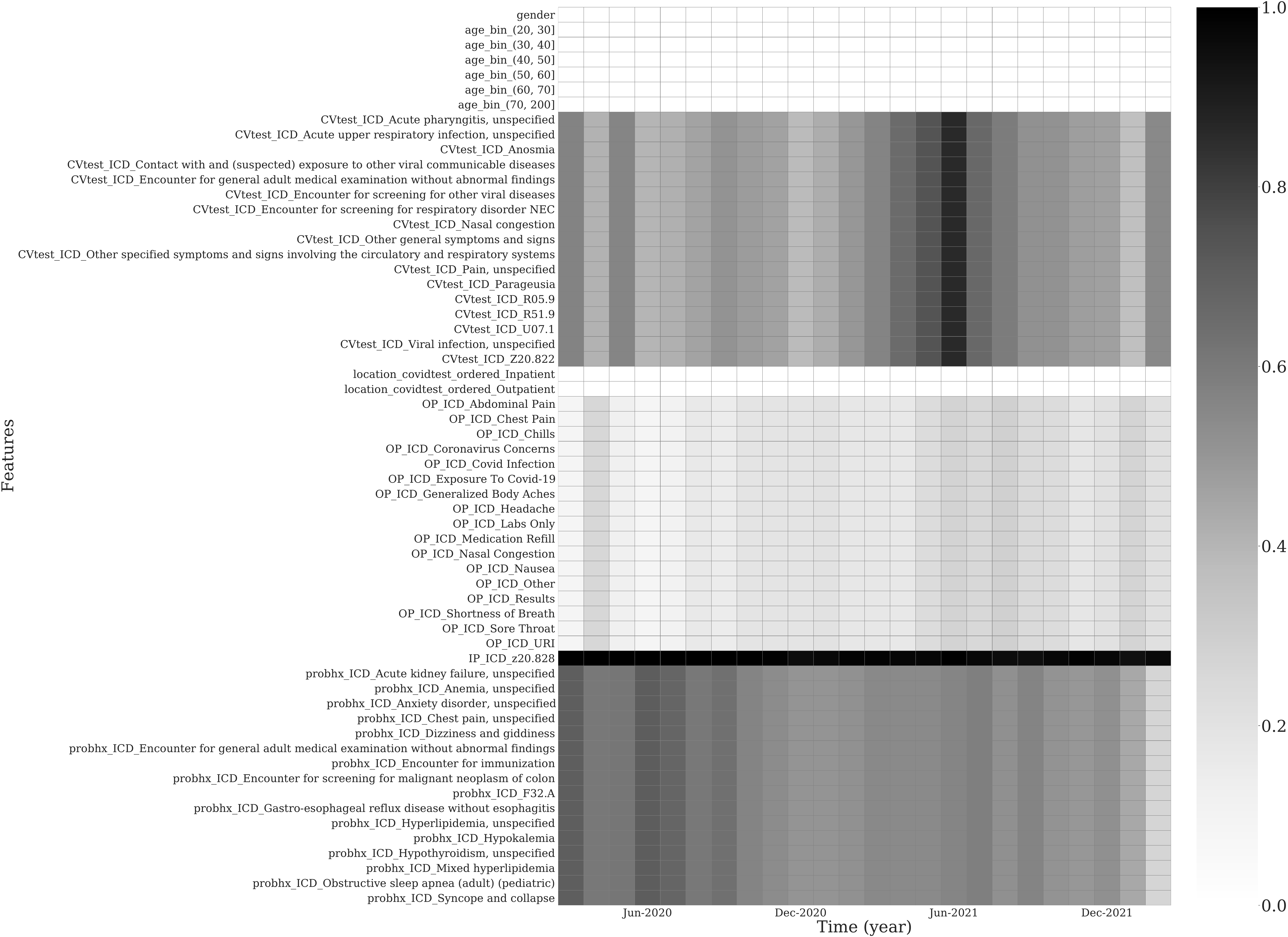}
  \caption{Missingness of categorical features in SWPA COVID-19 dataset (part 3).}
  \label{fig:heatmap_pa_covid_cate_3}
\end{figure}

\begin{figure}[H]
  \includegraphics[width=1.0\columnwidth]{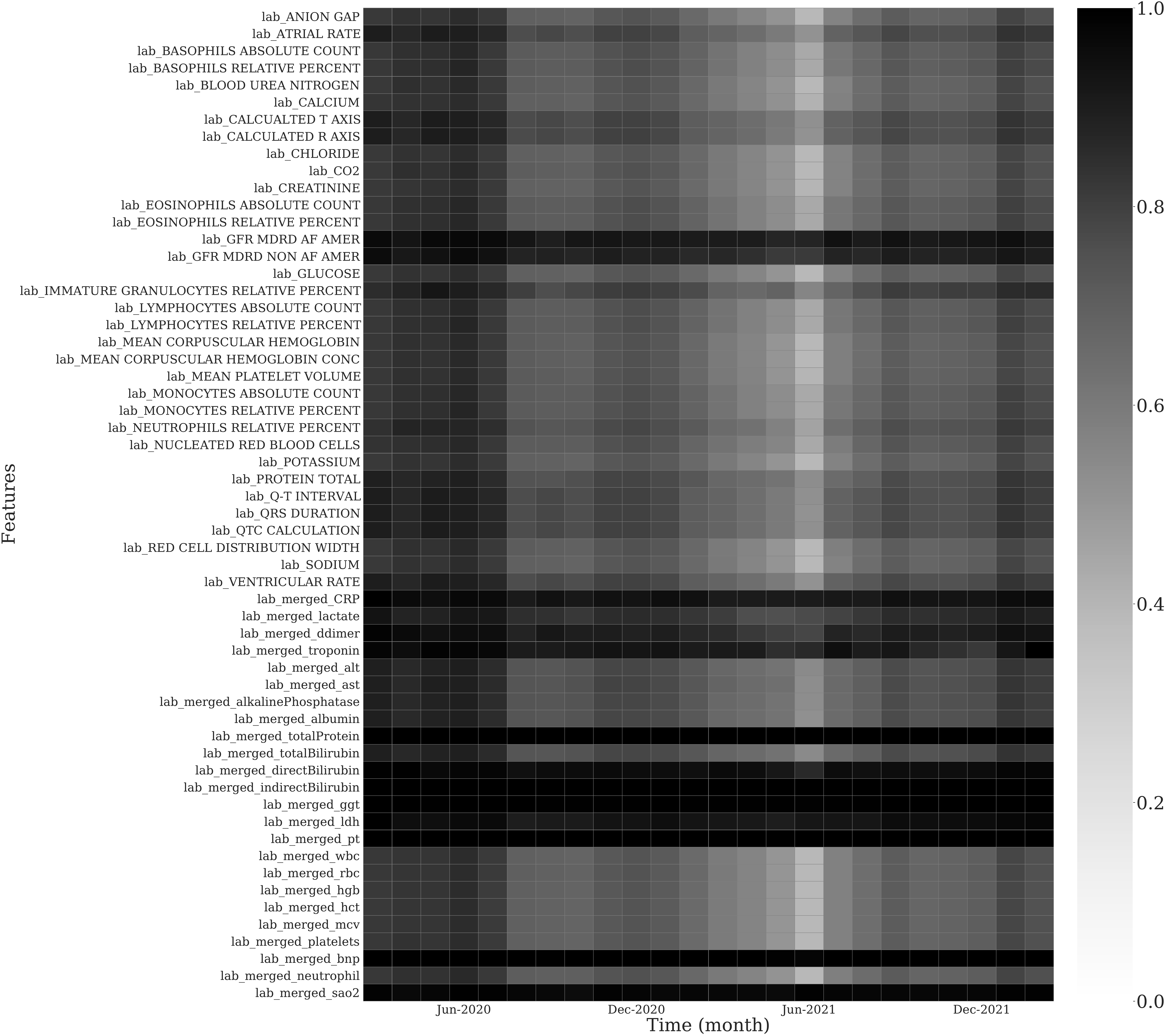}
  \caption{Missingness of numerical features in SWPA COVID-19.}
  \label{fig:heatmap_pa_covid_num}
\end{figure}

\begin{itemize}
    \item Data access: This is a private dataset.
    \item Cohort selection: The cohort consists of COVID-19 patients who tested positive for COVID-19 and were not already in the ICU or mechanically ventilated. We filter for the first positive test, and define features and outcomes relative to that time.  Cohort selection diagrams are given in Figures \ref{fig:swpa_covid_cohort}. If there are multiple samples per patient, we filter to the first entry per patient, which corresponds to when a patient first enters the dataset. This corresponds to a particular interpretation of the prediction: when a patient is first tests positive, given what we know about that patient, what is their estimated risk of 90-day mortality?
    \item Cohort characteristics: Cohort characteristics are given in Table \ref{tab:swpa_cohort_characteristics}.
    \item Outcome definition: 90-day mortality by comparing the death date and test date
    \item Features: We list the features used in the SWPA COVID-19 datasets in Section \ref{app:sec_swpa_features}. We convert all categorical variables into dummy features, and apply standard scaling to numerical variables (subtract mean and divide by standard deviation). To create a fixed length feature vector, where applicable we take the most recent value of each feature (e.g. most recent lab values).
    \item Missingness heat maps: are given in Figures \ref{fig:heatmap_pa_covid_cate_1}, \ref{fig:heatmap_pa_covid_cate_2}, \ref{fig:heatmap_pa_covid_cate_3}, and
    \ref{fig:heatmap_pa_covid_num},
\end{itemize}
\clearpage

\subsection{Cohort Selection and Cohort Characteristics}

\begin{table*}[ht]
\centering
\caption{SWPA COVID-19 cohort characteristics, with count (\%) or median (Q1--Q3).}
\vspace{0.5em}
\label{tab:swpa_cohort_characteristics}
    \begin{tabular}{lccc}
    \toprule
    Characteristic &  & Missingness &       Type \\
    \midrule
    \textbf{Gender} &                   &               &              \\
    \hspace{2em}Female &    20,283 (57.5\%) &          -- &  categorical \\
    \hspace{2em}Male &    15,003 (42.5\%) &          -- &  categorical \\
    \hspace{2em}Unknown &          7 (0.0\%) &          --&  categorical \\
    \textbf{Age} &                   &               &              \\
    \hspace{2em}Under 20 &                   3,210 (9.1\%)&               --&              categorical\\
    \hspace{2em}20 -- 30 &                   4,349 (12.3\%)&               --&              categorical\\
    \hspace{2em}30 -- 40 &                   4,667 (13.2\%)&               --&              categorical\\
    \hspace{2em}40 -- 50 &                   4,653 (13.2\%)&               --&              categorical\\
    \hspace{2em}50 -- 60 &                   6,111 (17.3\%)&               --&              categorical\\
    \hspace{2em}60 -- 70 &                   5,700 (16.2\%)&               --&              categorical\\
    \hspace{2em}70+ &                   6,603 (18.7\%)&               --&              categorical\\
    \textbf{Location of test} & & & \\
    \hspace{2em}Inpatient &                   14,911 (42.2\%)&               --&              categorical\\
    \hspace{2em}Outpatient &                   17,661 (50.0\%)&               --&              categorical\\
    \hspace{2em}Unknown &                   2,721 (7.7\%)&               --&              categorical\\
    \textbf{90-day mortality} &                   &               &              \\
    \hspace{2em}True &      1,516 (4.3\%) &          -- &  categorical \\
    \hspace{2em}False &    33,777 (95.7\%) &          -- &  categorical \\
    \bottomrule
    \end{tabular}%
\end{table*}

\clearpage
\twocolumn
\subsection{Features}\label{app:sec_swpa_features}
{\tiny
\begin{verbatim}
Asthma
CAD
CHF
CKD
COPD
CRP
CVtest_ICD_Acute pharyngitis, unspecified
CVtest_ICD_Acute upper respiratory infection, unspecified
CVtest_ICD_Anosmia
CVtest_ICD_Contact with and (suspected) exposure to other viral 
    communicable diseases
CVtest_ICD_Encounter for general adult medical 
    examination without 
    abnormal findings
CVtest_ICD_Encounter for screening for other viral diseases
CVtest_ICD_Encounter for screening for respiratory disorder NEC
CVtest_ICD_Nasal congestion
CVtest_ICD_Other general symptoms and signs
CVtest_ICD_Other specified symptoms and signs involving the 
    circulatory and respiratory systems
CVtest_ICD_Pain, unspecified
CVtest_ICD_Parageusia
CVtest_ICD_R05.9
CVtest_ICD_R51.9
CVtest_ICD_U07.1
CVtest_ICD_Viral infection, unspecified
CVtest_ICD_Z20.822
ESLD
Hypertension
IP_ICD_z20.828
Immunocompromised
Interstitial Lung disease
OP_ICD_Abdominal Pain
OP_ICD_Chest Pain
OP_ICD_Chills
OP_ICD_Coronavirus Concerns
OP_ICD_Covid Infection
OP_ICD_Exposure To Covid-19
OP_ICD_Generalized Body Aches
OP_ICD_Headache
OP_ICD_Labs Only
OP_ICD_Medication Refill
OP_ICD_Nasal Congestion
OP_ICD_Nausea
OP_ICD_Other
OP_ICD_Results
OP_ICD_Shortness of Breath
OP_ICD_Sore Throat
OP_ICD_URI
age_bin_(20, 30]
age_bin_(30, 40]
age_bin_(40, 50]
age_bin_(50, 60]
age_bin_(60, 70]
age_bin_(70, 200]
bmi
cancer
cough
covid_vaccination_given
diabetes
fatigue
fever
gender
hyperglycemia
lab_ANION GAP
lab_ATRIAL RATE
lab_BASOPHILS ABSOLUTE COUNT
lab_BASOPHILS RELATIVE PERCENT
lab_BLOOD UREA NITROGEN
lab_CALCIUM
lab_CALCUALTED T AXIS
lab_CALCULATED R AXIS
lab_CHLORIDE
lab_CO2
lab_CREATININE
lab_EOSINOPHILS ABSOLUTE COUNT
lab_EOSINOPHILS RELATIVE PERCENT
lab_GFR MDRD AF AMER
lab_GFR MDRD NON AF AMER
lab_GLUCOSE
lab_IMMATURE GRANULOCYTES RELATIVE PERCENT
lab_LYMPHOCYTES ABSOLUTE COUNT
lab_LYMPHOCYTES RELATIVE PERCENT
lab_MEAN CORPUSCULAR HEMOGLOBIN
lab_MEAN CORPUSCULAR HEMOGLOBIN CONC
lab_MEAN PLATELET VOLUME
lab_MONOCYTES ABSOLUTE COUNT
lab_MONOCYTES RELATIVE PERCENT
lab_NEUTROPHILS RELATIVE PERCENT
lab_NUCLEATED RED BLOOD CELLS
lab_POTASSIUM
lab_PROTEIN TOTAL
lab_Q-T INTERVAL
lab_QRS DURATION
lab_QTC CALCULATION
lab_RED CELL DISTRIBUTION WIDTH
lab_SODIUM
lab_VENTRICULAR RATE
lab_merged_CRP
lab_merged_albumin
lab_merged_alkalinePhosphatase
lab_merged_alt
lab_merged_ast
lab_merged_bnp
lab_merged_ddimer
lab_merged_directBilirubin
lab_merged_ggt
lab_merged_hct
lab_merged_hgb
lab_merged_indirectBilirubin
lab_merged_lactate
lab_merged_ldh
lab_merged_mcv
lab_merged_neutrophil
lab_merged_platelets
lab_merged_pt
lab_merged_rbc
lab_merged_sao2
lab_merged_totalBilirubin
lab_merged_totalProtein
lab_merged_troponin
lab_merged_wbc
labs_ICD_Acute pharyngitis, unspecified
labs_ICD_Acute upper respiratory infection, unspecified
labs_ICD_Chest pain, unspecified
labs_ICD_Contact with and (suspected) exposure to other 
    viral communicable diseases
labs_ICD_Dyspnea, unspecified
labs_ICD_Encounter for other preprocedural examination
labs_ICD_Essential (primary) hypertension
labs_ICD_Fever, unspecified
labs_ICD_Heart failure, unspecified
labs_ICD_Other general symptoms and signs
labs_ICD_Other pulmonary embolism without acute cor pulmonale
labs_ICD_Other specified abnormalities of plasma proteins
labs_ICD_R05.9
labs_ICD_Shortness of breath
labs_ICD_Syncope and collapse
labs_ICD_U07.1
labs_ICD_Unspecified atrial fibrillation
labs_ICD_Viral infection, unspecified
labs_ICD_Z20.822
liver disease
location_covidtest_ordered_Inpatient
location_covidtest_ordered_Outpatient
lung disease
med_dx_Acquired hypothyroidism
med_dx_Anxiety
med_dx_COVID-19
med_dx_Encounter for antineoplastic chemotherapy
med_dx_Encounter for antineoplastic chemotherapy and immunotherapy
med_dx_Encounter for antineoplastic immunotherapy
med_dx_Encounter for immunization
med_dx_Gastroesophageal reflux disease without esophagitis
med_dx_Gastroesophageal reflux disease, esophagitis presence 
    not specified
med_dx_Generalized anxiety disorder
med_dx_Hyperlipidemia, unspecified hyperlipidemia type
med_dx_Hypomagnesemia
med_dx_Hypothyroidism, unspecified type
med_dx_Iron deficiency anemia, unspecified iron deficiency anemia type
med_dx_Mixed hyperlipidemia
med_dx_Primary osteoarthritis of right knee
medication_ACETAMINOPHEN 325 MG TABLET
medication_ALBUTEROL SULFATE 2.5 MG/3 ML (0.083 %
    FOR NEBULIZATION
medication_ALBUTEROL SULFATE HFA 90 MCG/ACTUATION AEROSOL INHALER
medication_ASPIRIN 81 MG TABLET,DELAYED RELEASE
medication_DEXAMETHASONE SODIUM PHOSPHATE 4 MG/ML INJECTION SOLUTION
medication_DIPHENHYDRAMINE 50 MG/ML INJECTION (WRAPPER)
medication_EPINEPHRINE 0.3 MG/0.3 ML INJECTION, AUTO-INJECTOR
medication_FENTANYL (PF) 50 MCG/ML INJECTION SOLUTION
medication_HYDROCODONE 5 MG-ACETAMINOPHEN 325 MG TABLET
medication_HYDROCORTISONE SOD SUCCINATE (PF) 100 MG/2 ML SOLUTION 
    FOR INJECTION
medication_IOPAMIDOL 76 %
medication_LACTATED RINGERS INTRAVENOUS SOLUTION
medication_MIDAZOLAM 1 MG/ML INJECTION SOLUTION
medication_NALOXONE 0.4 MG/ML INJECTION SOLUTION
medication_ONDANSETRON HCL (PF) 4 MG/2 ML INJECTION SOLUTION
medication_OXYCODONE 5 MG TABLET
medication_PANTOPRAZOLE 40 MG TABLET,DELAYED RELEASE
medication_PROPOFOL 10 MG/ML INTRAVENOUS BOLUS (20 ML)
medication_SODIUM CHLORIDE 0.9 %
medication_SODIUM CHLORIDE 0.9 %
myalgia
obesity
past7Dprobhx_ICD_Acute kidney failure, unspecified
past7Dprobhx_ICD_Anemia, unspecified
past7Dprobhx_ICD_Anxiety disorder, unspecified
past7Dprobhx_ICD_Chest pain, unspecified
past7Dprobhx_ICD_Dizziness and giddiness
past7Dprobhx_ICD_Encounter for general adult medical examination 
    without abnormal findings
past7Dprobhx_ICD_Encounter for immunization
past7Dprobhx_ICD_Encounter for screening for malignant 
    neoplasm of colon
past7Dprobhx_ICD_F32.A
past7Dprobhx_ICD_Gastro-esophageal reflux disease 
    without esophagitis
past7Dprobhx_ICD_Hyperlipidemia, unspecified
past7Dprobhx_ICD_Hypokalemia
past7Dprobhx_ICD_Hypothyroidism, unspecified
past7Dprobhx_ICD_Mixed hyperlipidemia
past7Dprobhx_ICD_Obstructive sleep apnea (adult) (pediatric)
past7Dprobhx_ICD_Syncope and collapse
past7Dprobhx_ICD_Type 2 diabetes mellitus without complications
past7Dprobhx_ICD_Unspecified atrial fibrillation
probhx_ICD_Acute kidney failure, unspecified
probhx_ICD_Anemia, unspecified
probhx_ICD_Anxiety disorder, unspecified
probhx_ICD_Chest pain, unspecified
probhx_ICD_Dizziness and giddiness
probhx_ICD_Encounter for general adult medical examination without 
    abnormal findings
probhx_ICD_Encounter for immunization
probhx_ICD_Encounter for screening for malignant neoplasm of colon
probhx_ICD_F32.A
probhx_ICD_Gastro-esophageal reflux disease without esophagitis
probhx_ICD_Hyperlipidemia, unspecified
probhx_ICD_Hypokalemia
probhx_ICD_Hypothyroidism, unspecified
probhx_ICD_Mixed hyperlipidemia
probhx_ICD_Obstructive sleep apnea (adult) (pediatric)
probhx_ICD_Syncope and collapse
probhx_ICD_Type 2 diabetes mellitus without complications
probhx_ICD_Unspecified atrial fibrillation
transplant
troponin
vaccine_COVID-19 RS-AD26 (PF) Vaccine (Janssen)
vaccine_COVID-19 Vaccine, Unspecified
vaccine_COVID-19 mRNA (PF) Vaccine (Moderna)
vaccine_COVID-19 mRNA (PF) Vaccine (Pfizer)
vaccine_Flu Whole
vaccine_INFLUENZA, CCIV4
vaccine_Influenza
vaccine_Influenza High PF
vaccine_Influenza ID PF
vaccine_Influenza PF
vaccine_Influenza Vaccine, Quadrivalent, Adjuvanted
vaccine_Influenza, High-dose, Quadrivalent
vaccine_Influenza, Quadrivalent
vaccine_Influenza, Recombinant (RIV4)
vaccine_Influenza, Recombinant (Riv3)
vaccine_Influenza, Trivalent, Adjuvanted
vaccine_LAIV3
vaccine_Pneumococcal
vaccine_Pneumococcal Conjugate 13-valent
vaccine_Pneumococcal Polysaccharide
vaccine_TIVA
\end{verbatim}}
\onecolumn

\subsection{Missingness heatmaps}

This section plots missingness heatmaps of categorical and numerical features over time. Darker color means larger proportion of missing data.

\clearpage

\clearpage

\clearpage

\clearpage

\section{Additional MIMIC-IV Data Details}\label{app:mimic_data}
The Medical Information Mart for Intensive Care (MIMIC)-IV \citep{mimiciv_v1} database contains EHR data from patients admitted to critical care units 
from 2008--2019. MIMIC-IV is an update to MIMIC-III, adding time annotations placing each sample into a three-year time range, and removing elements from the old CareVue EHR system (before 2008). 
Each patient has an \verb|anchor_year_group|, \verb|anchor_year| and \verb|intime|. For each patient, we first calculated an offset as the difference between \verb|intime| and \verb|anchor_year|. Then, we approximated the admit time as the midpoint of \verb|anchor_year_group| after applying the computed offset.

The performance over time is evaluated on a \emph{yearly} basis. Our study uses MIMIC-IV-1.0. 

\begin{figure}[!h]
\centering
  \includegraphics[width=0.9\columnwidth]{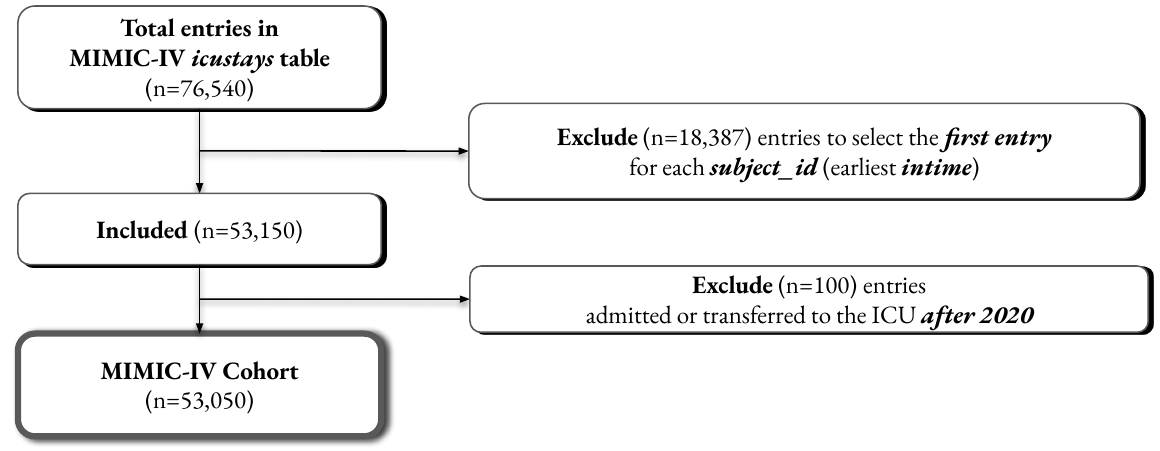}
  \vspace{-1em}
  \caption{Cohort selection diagram - MIMIC-IV}
  \label{fig:mimic_iv_cohort}
\end{figure}

\begin{figure}[H]
  \includegraphics[width=1.0\columnwidth]{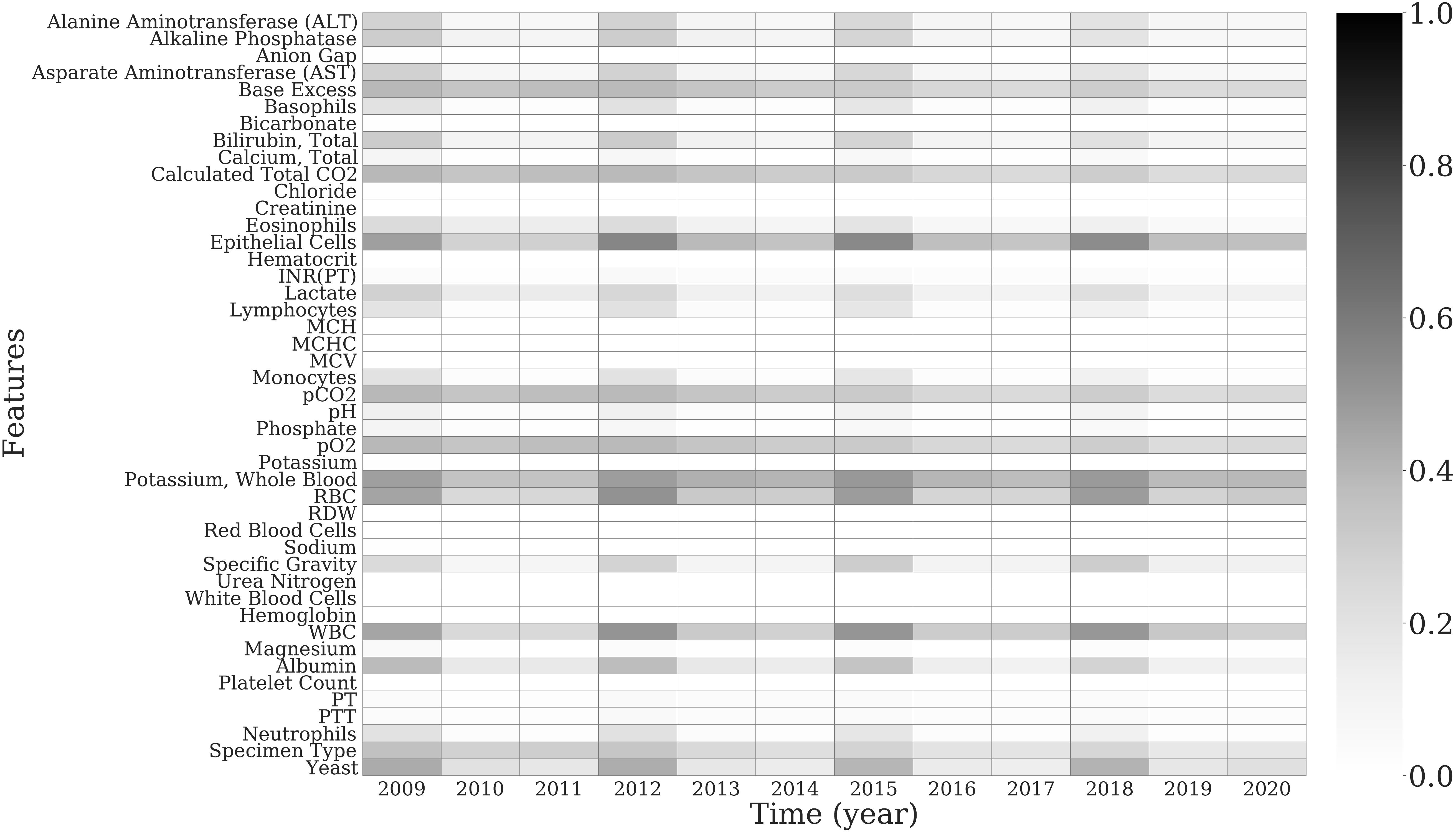}
  \caption{Missingness over time for labevents features in MIMIC-IV dataset after cohort selection. The darker the color, the larger the proportion of missing data.}
  \label{fig:heatmap_mimic_lab}
\end{figure}

\begin{figure}[H]
  \includegraphics[width=1.0\columnwidth]{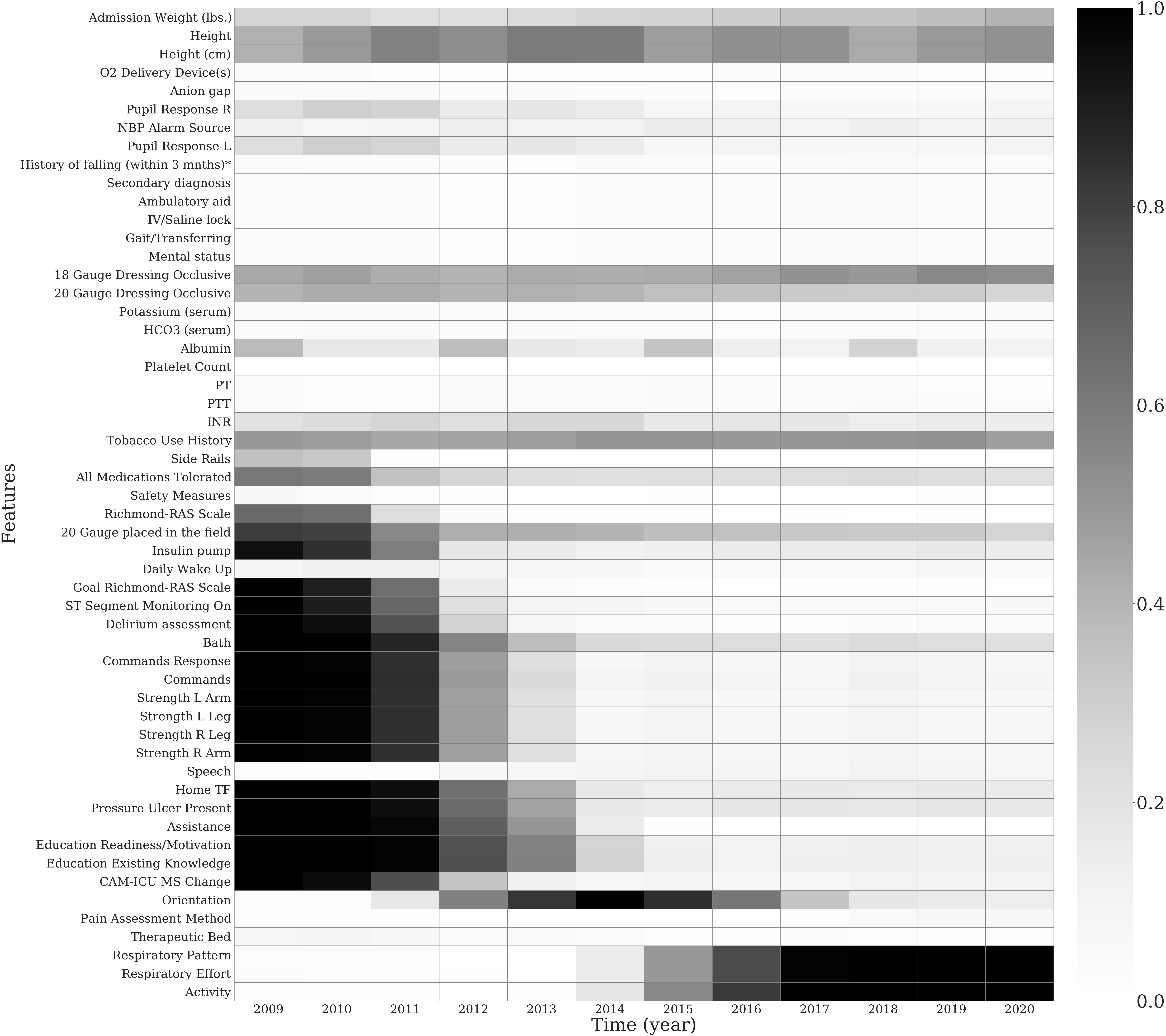}
  \caption{Missingness over time for chartevents features in MIMIC-IV dataset after cohort selection. The darker the color, the larger the proportion of missing data. (part 1)}
  \label{fig:heatmap_mimic_chart_1}
\end{figure}

\begin{figure}[H]
  \includegraphics[width=1.0\columnwidth]{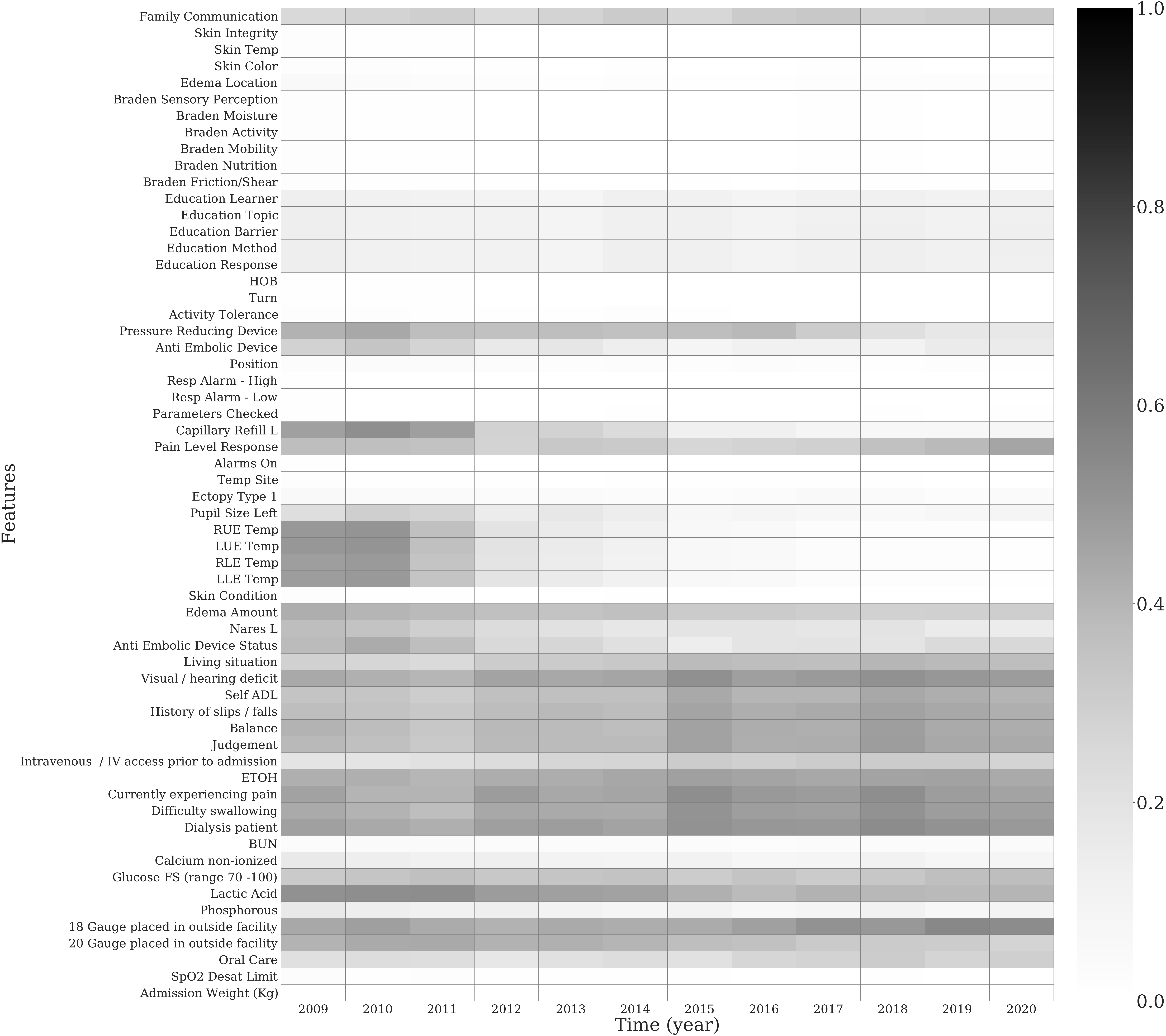}
  \caption{Missingness over time for chartevents features in MIMIC-IV dataset after cohort selection. The darker the color, the larger the proportion of missing data. (part 2)}
  \label{fig:heatmap_mimic_chart_2}
\end{figure}

\begin{figure}[H]
  \includegraphics[width=1.0\columnwidth]{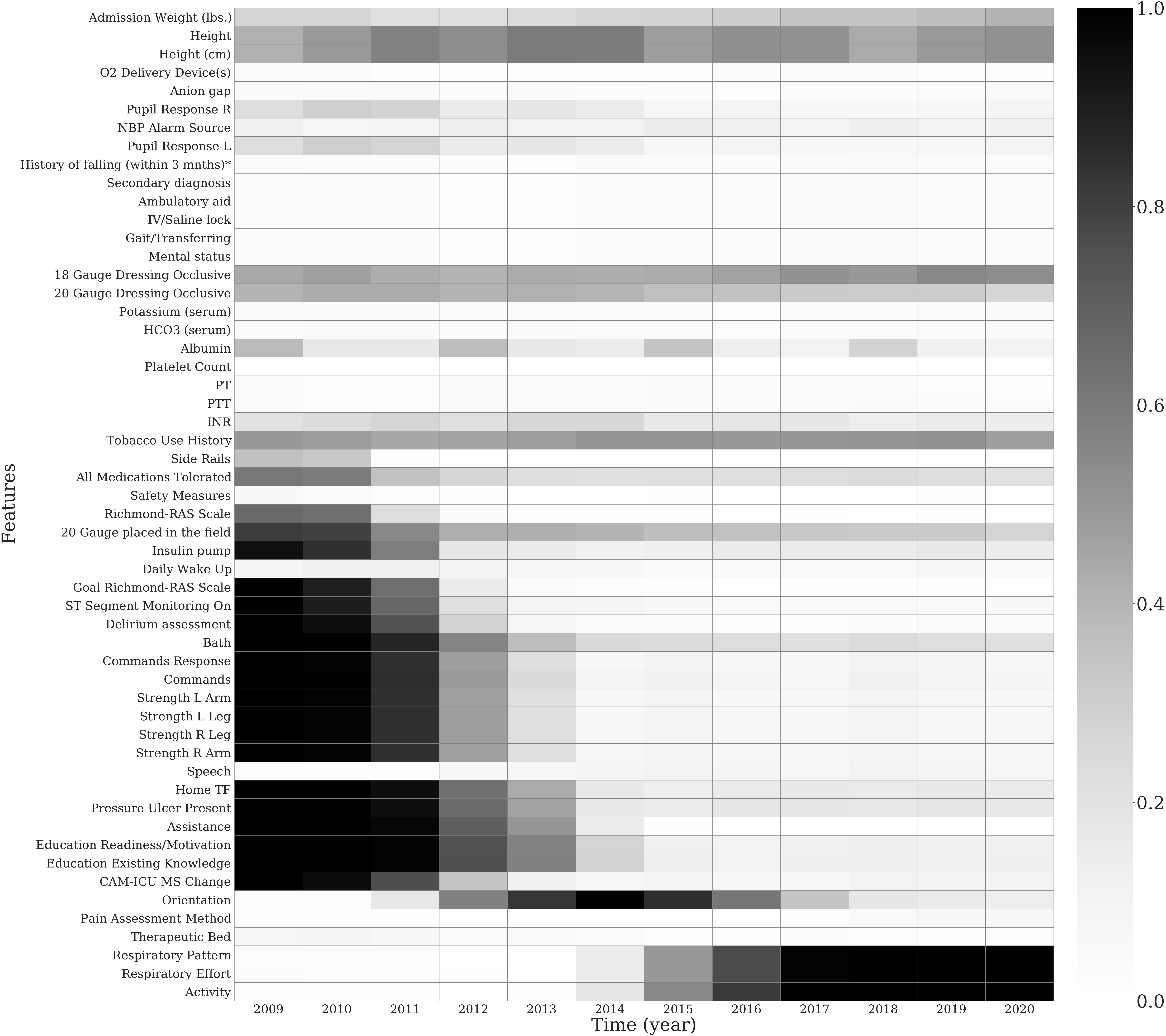}
  \caption{Missingness over time for chartevents features in MIMIC-IV dataset after cohort selection. The darker the color, the larger the proportion of missing data. (part 3)}
  \label{fig:heatmap_mimic_chart_3}
\end{figure}

\begin{itemize} 
    \item Data access: Users must create a Physionet account, become credentialed, and sign a data use agreement (DUA).
    \item Cohort selection: We select all patients in the \verb|icustays| table, filtering for their first encounter (minimum \verb|intime|), and defining a feature vector only using information available by the first 24 hrs of their first encounter. (Selection diagram in Figure \ref{fig:mimic_iv_cohort}). If there are multiple samples per patient, we filter to the first entry per patient, which corresponds to when a patient first enters the dataset. This corresponds to a particular interpretation of the prediction: when a patient first visits the ICU, given what we know about that patient, what is their estimated risk of in-ICU mortality?
    \item Outcome definition: The outcome of interest is in-ICU mortality, defined by comparing the \verb|outtime| of the patient's ICU visit with the patient's \verb|dod| (date of death, in the \verb|patients| table). As noted in the documentation, out-of-hospital mortality is not recorded. 
    \item Cohort characteristics: Cohort characteristics are given in Table \ref{tab:mimic_iv_characteristics}.
    \item Features: We list the features used in the MIMIC-IV datasets in Section \ref{app:sec_mimic_features}. We convert all categorical variables into dummy features, and apply standard scaling to numerical variables (subtract mean and divide by standard deviation). To create a fixed length feature vector, we take the most recent value of any patient history data available (e.g. most recent lab values).
    \item Missingness heat maps: are given in Figures \ref{fig:heatmap_mimic_lab}, \ref{fig:heatmap_mimic_chart_1}, \ref{fig:heatmap_mimic_chart_2},
    \ref{fig:heatmap_mimic_chart_3}.
\end{itemize}

\newpage
\subsection{Cohort Selection and Cohort Characteristics}

\begin{table*}[htb]
\caption{MIMIC-IV cohort characteristics, with count (\%) or median (Q1--Q3).}
\vspace{0.5em}
\label{tab:mimic_iv_characteristics}
    \centering
    \resizebox{0.88\columnwidth}{!}{%
    \begin{tabular}{lcccc}
    \toprule
    Characteristic &  & Missingness &       Type \\
    \midrule
    \textbf{Gender} &                   &               &              \\
    \hspace{2em}Female &    23,313 (43.9\%) &            -- &  categorical \\
    \hspace{2em}Male &    29,737 (56.1\%) &            -- &  categorical \\
    \textbf{Age at Admission} &        66 (54-78) &          0.0\% &   continuous \\
    \textbf{O2 Delivery Device(s)} &                   &               &              \\
    \hspace{2em}Use device &    33,359 (62.9\%) &            -- &  categorical \\
    \hspace{2em}None &    18,549 (35.0\%) &            -- &  categorical \\
    \hspace{2em}Missing &      1,142 (2.2\%) &            -- &  categorical \\
    \textbf{Pupil Response R} &                   &               &              \\
    \hspace{2em}Brisk  &    39,708 (74.9\%) &            -- &  categorical \\
    \hspace{2em}Sluggish &      4,603 (8.7\%) &            -- &  categorical \\
    \hspace{2em}Non-reactive &      1,812 (3.4\%) &            -- &  categorical \\
    \hspace{2em}Missing &     6,927 (13.1\%) &            -- &  categorical \\
    \textbf{first\_careunit} &                   &               &              \\
    \hspace{2em}Medical Intensive Care Unit (MICU) &    10,213 (19.3\%) &            -- &  categorical \\
    \hspace{2em}Surgical Intensive Care Unit (SICU) &     8,241 (15.5\%) &            -- &  categorical \\
    \hspace{2em}Medical/Surgical Intensive Care Unit (MICU/S... &     8,808 (16.6\%) &            -- &  categorical \\
    \hspace{2em}Cardiac Vascular Intensive Care Unit (CVICU) &     9,437 (17.8\%) &            -- &  categorical \\
    \hspace{2em}Coronary Care Unit (CCU) &     6,098 (11.5\%) &            -- &  categorical \\
    \hspace{2em}Trauma SICU (TSICU) &     6,947 (13.1\%) &            -- &  categorical \\
    \hspace{2em}Other &      3,306 (6.2\%) &            -- &  categorical \\
    \textbf{Anion Gap} &        13 (11-16) &          0.5\% &   continuous \\
    \textbf{Heart Rhythm} &                   &               &              \\
    \hspace{2em}SR (Sinus Rhythm) &    34,004 (64.1\%) &            -- &  categorical \\
    \hspace{2em}Abnormal heart rhythm &    18,657 (35.2\%) &            -- &  categorical \\
    \hspace{2em}Missing &        389 (0.7\%) &            -- &  categorical \\
    \textbf{Glucose FS (range 70 -100)} &     131 (110-164) &         32.7\% &   continuous \\
    \textbf{Eye Opening} &                   &               &              \\
    \hspace{2em}Spontaneously &    39,216 (73.9\%) &            -- &  categorical \\
    \hspace{2em}To Speech &     7,387 (13.9\%) &            -- &  categorical \\
    \hspace{2em}None &      4,538 (8.6\%) &            -- &  categorical \\
    \hspace{2em}To Pain &      1,702 (3.2\%) &            -- &  categorical \\
    \hspace{2em}Missing &        207 (0.4\%) &            -- &  categorical \\
    \textbf{Lactate} &           2 (1-2) &         22.0\% &   continuous \\
    \textbf{Motor Response} &                   &               &              \\
    \hspace{2em}Obeys Commands &    44,409 (83.7\%) &            -- &  categorical \\
    \hspace{2em}Localizes Pain &      3,419 (6.4\%) &            -- &  categorical \\
    \hspace{2em}Flex-withdraws &      1,673 (3.2\%) &            -- &  categorical \\
    \hspace{2em}No response &      2,930 (5.5\%) &            -- &  categorical \\
    \hspace{2em}Abnormal extension &        157 (0.3\%) &            -- &  categorical \\
    \hspace{2em}Abnormal Flexion &        238 (0.4\%) &            -- &  categorical \\
    \hspace{2em}Missing &        224 (0.4\%) &          --   &  categorical \\
    \textbf{Respiratory Pattern} &                   &               &              \\
    \hspace{2em}Regular &    29,373 (55.4\%) &          --   &  categorical \\
    \hspace{2em}Not regular &      1,739 (3.3\%) &        --     &  categorical \\
    \hspace{2em}Missing &    21,938 (41.4\%) &            -- &  categorical \\
    \textbf{Richmond-RAS Scale} &          0 (-1-0) &         15.4\% &  categorical \\
    \textbf{in-icu mortality} &                   &               &              \\
    \hspace{2em}0 &    49,716 (93.7\%) &         --    &  categorical \\
    \hspace{2em}1 &      3,334 (6.3\%) &          --   &  categorical \\
    \bottomrule
    \end{tabular}%
    }   
\end{table*}
\clearpage

\twocolumn
\subsection{Features}\label{app:sec_mimic_features}

{\small
\begin{verbatim}
18 Gauge Dressing Occlusive
18 Gauge placed in outside facility
20 Gauge Dressing Occlusive
20 Gauge placed in outside facility
20 Gauge placed in the field
Abdominal Assessment
Activity
Activity Tolerance
Admission Weight (Kg)
Admission Weight (lbs.)
Alanine Aminotransferase (ALT)
Alarms On
Albumin
Alkaline Phosphatase
All Medications Tolerated
Ambulatory aid
Anion Gap
Anion gap
Anti Embolic Device
Anti Embolic Device Status
Asparate Aminotransferase (AST)
Assistance
BUN
Balance
Base Excess
Basophils
Bath
Bicarbonate
Bilirubin, Total
Bowel Sounds
Braden Activity
Braden Friction/Shear
Braden Mobility
Braden Moisture
Braden Nutrition
Braden Sensory Perception
CAM-ICU MS Change
Calcium non-ionized
Calcium, Total
Calculated Total CO2
Capillary Refill L
Capillary Refill R
Chloride
Chloride (serum)
Commands
Commands Response
Cough Effort
Cough Type
Creatinine
Creatinine (serum)
Currently experiencing pain
Daily Wake Up
Delirium assessment
Dialysis patient
Diet Type
Difficulty swallowing
Dorsal PedPulse L
Dorsal PedPulse R
ETOH
Ectopy Type 1
Edema Amount
Edema Location
Education Barrier
Education Existing Knowledge
Education Learner
Education Method
Education Readiness/Motivation
Education Response
Education Topic
Eosinophils
Epithelial Cells
Eye Opening
Family Communication
Flatus
GU Catheter Size
Gait/Transferring
Glucose (serum)
Glucose FS (range 70 -100)
Goal Richmond-RAS Scale
HCO3 (serum)
HOB
HR
HR Alarm - High
HR Alarm - Low
Heart Rhythm
Height
Height (cm)
Hematocrit
Hematocrit (serum)
Hemoglobin
History of falling (within 3 mnths)*
History of slips / falls
Home TF
INR
INR(PT)
IV/Saline lock
Insulin pump
Intravenous  / IV access prior to admission
Judgement
LLE Color
LLE Temp
LLL Lung Sounds
LUE Color
LUE Temp
LUL Lung Sounds
Lactate
Lactic Acid
Living situation
Lymphocytes
MCH
MCHC
MCV
Magnesium
Mental status
Monocytes
Motor Response
NBP Alarm - High
NBP Alarm - Low
NBP Alarm Source
NBPd
NBPm
NBPs
Nares L
Nares R
Neutrophils
O2 Delivery Device(s)
Oral Care
Oral Cavity
Orientation
PT
PTT
Pain Assessment Method
Pain Cause
Pain Level
Pain Level Acceptable
Pain Level Response
Pain Location
Pain Management
Pain Present
Pain Type
Parameters Checked
Phosphate
Phosphorous
Platelet Count
Position
PostTib Pulses L
PostTib Pulses R
Potassium
Potassium (serum)
Potassium, Whole Blood
Pressure Reducing Device
Pressure Ulcer Present
Pupil Response L
Pupil Response R
Pupil Size Left
Pupil Size Right
RBC
RDW
RLE Color
RLE Temp
RLL Lung Sounds
RR
RUE Color
RUE Temp
RUL Lung Sounds
Radial Pulse L
Radial Pulse R
Red Blood Cells
Resp Alarm - High
Resp Alarm - Low
Respiratory Effort
Respiratory Pattern
Richmond-RAS Scale
ST Segment Monitoring On
Safety Measures
Secondary diagnosis
Self ADL
Side Rails
Skin Color
Skin Condition
Skin Integrity
Skin Temp
Sodium
Sodium (serum)
SpO2
SpO2 Alarm - High
SpO2 Alarm - Low
SpO2 Desat Limit
Specific Gravity
Specimen Type
Speech
Strength L Arm
Strength L Leg
Strength R Arm
Strength R Leg
Support Systems
Temp Site
Temperature F
Therapeutic Bed
Tobacco Use History
Turn
Untoward Effect
Urea Nitrogen
Urine Source
Verbal Response
Visual / hearing deficit
WBC
White Blood Cells
Yeast
admit_age
gender
pCO2
pH
pO2
\end{verbatim}}
\onecolumn

\subsection{Missingness heatmaps}

\clearpage

\clearpage

\clearpage

\section{Additional OPTN (Liver) Data Details}\label{app:optn_data}
The Organ Procurement and Transplantation Network (OPTN) database \cite{optn_data} tracks organ donation and transplant events in the U.S. Our study uses data from candidates on the liver transplant wait list. The performance over time is evaluated on a \emph{yearly} basis.

\begin{figure}[ht]
\centering
  \includegraphics[width=0.9\textwidth]{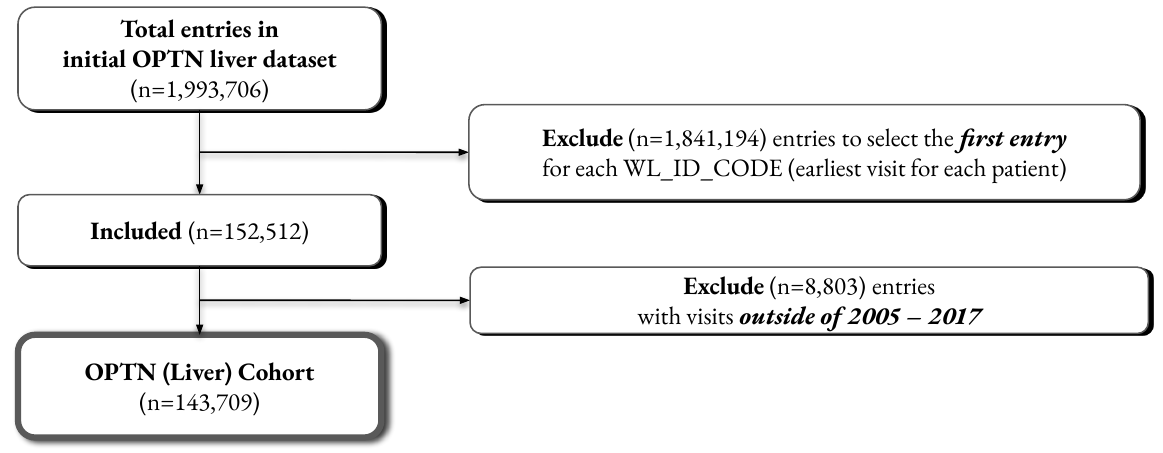}
  \vspace{-0.5em}
  \caption{Cohort selection diagram - OPTN (Liver)}
  \label{fig:optn_liver_cohort}
\end{figure}

\begin{figure}[H]
  \includegraphics[width=1.0\columnwidth]{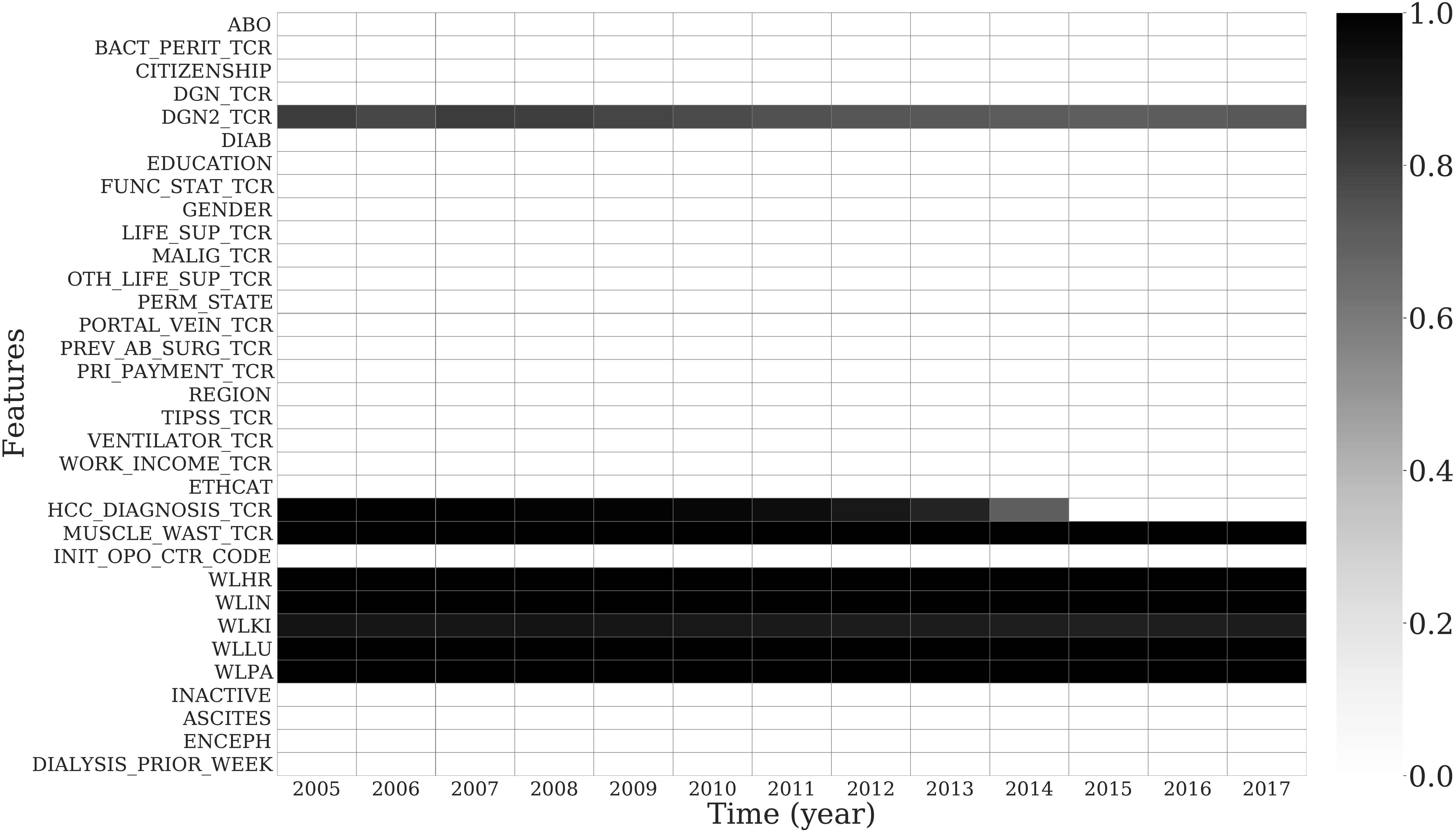}
  \caption{Missingness over time for categorical features in OPTN (Liver) dataset after cohort selection. The darker the color, the larger the proportion of missing data.}
  \label{fig:heatmap_optn_liver_cate}
\end{figure}

\begin{figure}[H]
  \includegraphics[width=1.0\columnwidth]{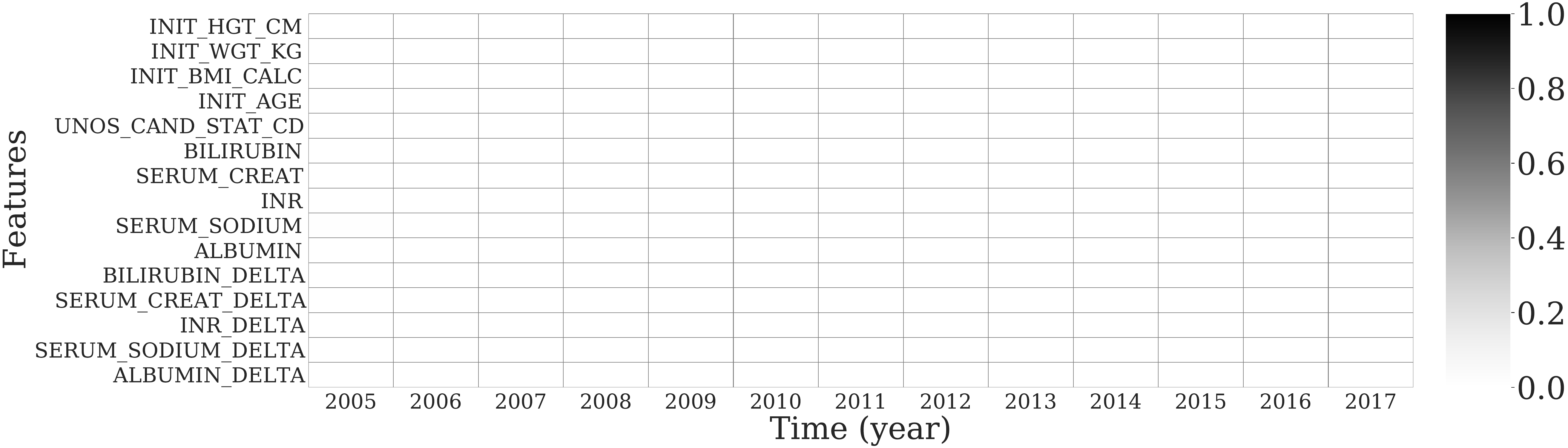}
  \caption{Missingness over time for numerical features in OPTN (Liver) dataset after cohort selection. The darker the color, the larger the proportion of missing data. (Near-zero missingness here.)}
  \label{fig:heatmap_optn_liver_num}
\end{figure}

\begin{itemize}
    \item First, we provide the disclaimer: 
``The data reported here have been supplied by the United Network for Organ Sharing as the contractor for the Organ Procurement and transplantation Network. The interpretation and reporting of these data are the responsibility of the author(s) and in no way should be seen as an official policy of or interpretation by the OPTN or the U.S. Government''.
    \item Data access: After signing the Data Use Agreement - I from Organ Procedurement And Transplantation network, users can access the OPTN (Liver) dataset.
    \item Cohort selection: The cohort consists of liver transplant candidates on the waiting list (2005-2017). We follow the same pipeline as \citet{byrd2021predicting} to extract the data, except that we select the first record for each patient. Cohort selection diagrams are given in Figures \ref{fig:optn_liver_cohort}. This corresponds to a particular interpretation of the prediction: when a patient is first added to the transplant list, given what we know about that patient, what is their estimated risk of 180-day mortality?
    \item Outcome definition: 180-day mortality from when the patient was first added to the list
    \item Cohort characteristics: Cohort characteristics are given in Table \ref{tab:optn_liver_characteristics}.
    \item Features: We list the features used in the OPTN liver dataset in Section \ref{app:sec_optn_features}. We convert all categorical variables into dummy features, and apply standard scaling to numerical variables (subtract mean and divide by standard deviation).
    \item Missingness heat maps: are given in Figures \ref{fig:heatmap_optn_liver_cate} and \ref{fig:heatmap_optn_liver_num}.
\end{itemize}

\clearpage
\subsection{Cohort Selection and Cohort Characteristics}

\begin{table*}[!htb]
\caption{OPTN (Liver) cohort characteristics, with count (\%) or median (Q1 -- Q3).}
\label{tab:optn_liver_characteristics}
    \centering
    \begin{tabular}{lccc}
    \toprule
    Feature name (value) &   & Empty (ratio) &       Type \\
    \midrule
    \textbf{Gender} &                        &               &              \\
    \hspace{2em}Male &         92,560 (64.4\%) &            -- &  categorical \\
    \hspace{2em}Female &         51,149 (35.6\%) &            -- &  categorical \\
    \textbf{INIT\_AGE} &             56 (49-62) &          0.0\% &   continuous \\
    \textbf{FUNC\_STAT\_TCR} &    2,070 (2,050-2,080) &          0.0\% &  categorical \\
    \textbf{INIT\_OPO\_CTR\_CODE} &  11,036 (3,782-19,282) &          0.0\% &  categorical \\
    \textbf{ALBUMIN} &                3 (3-4) &          0.0\% &   continuous \\
    \textbf{HCC\_DIAGNOSIS\_TCR} &                        &               &              \\
    \hspace{2em}No &         31,390 (21.8\%) &            -- &  categorical \\
    \hspace{2em}Yes &          11,312 (7.9\%) &            -- &  categorical \\
    \hspace{2em}Missing &        101,007 (70.3\%) &            -- &  categorical \\
    \textbf{PERM\_STATE} &                        &               &              \\
    \hspace{2em}CA &         19,645 (13.7\%) &            -- &  categorical \\
    \hspace{2em}TX &         14,692 (10.2\%) &            -- &  categorical \\
    \hspace{2em}NY &           9,976 (6.9\%) &            -- &  categorical \\
    \hspace{2em}GA &           4,052 (2.8\%) &            -- &  categorical \\
    \hspace{2em}MD &           4,050 (2.8\%) &            -- &  categorical \\
    \hspace{2em}FL &           7,602 (5.3\%) &            -- &  categorical \\
    \hspace{2em}PA &           8,013 (5.6\%) &            -- &  categorical \\
    \hspace{2em}MI &           3,989 (2.8\%) &            -- &  categorical \\
    \hspace{2em}Other &         71,007 (49.4\%) &            -- &  categorical \\
    \textbf{EDUCATION} &                4 (3-5) &          0.0\% &  categorical \\
    \textbf{ASCITES} &                2 (1-2) &          0.0\% &  categorical \\
    \textbf{MORTALITY\_180D} &                        &               &              \\
    \hspace{2em}1 &           4,635 (3.2\%) &            -- &  categorical \\
    \hspace{2em}0 &        139,074 (96.8\%) &            -- &  categorical \\
    \bottomrule
    \end{tabular}%
\end{table*}
\clearpage

\twocolumn
\subsection{Features}\label{app:sec_optn_features}
{\small
\begin{verbatim}
ABO
BACT_PERIT_TCR
CITIZENSHIP
DGN_TCR
DGN2_TCR
DIAB
EDUCATION
FUNC_STAT_TCR
GENDER
LIFE_SUP_TCR
MALIG_TCR
OTH_LIFE_SUP_TCR
PERM_STATE
PORTAL_VEIN_TCR
PREV_AB_SURG_TCR
PRI_PAYMENT_TCR
REGION
TIPSS_TCR
VENTILATOR_TCR
WORK_INCOME_TCR
ETHCAT
HCC_DIAGNOSIS_TCR
MUSCLE_WAST_TCR
INIT_OPO_CTR_CODE
WLHR
WLIN
WLKI
WLLU
WLPA
INACTIVE
ASCITES
ENCEPH
DIALYSIS_PRIOR_WEEK
INIT_HGT_CM
INIT_WGT_KG
INIT_BMI_CALC
INIT_AGE
UNOS_CAND_STAT_CD
BILIRUBIN
SERUM_CREAT
INR
SERUM_SODIUM
ALBUMIN
BILIRUBIN_DELTA
SERUM_CREAT_DELTA
INR_DELTA
SERUM_SODIUM_DELTA
ALBUMIN_DELTA
\end{verbatim}}
\onecolumn

\subsection{Missingness heatmaps}

\newpage
\section{Additional MIMIC-CXR Data Details}\label{app:mimic_cxr_data}
The MIMIC Chest X-ray (MIMIC-CXR-JPG) \citep{johnson_pollard_greenbaum_lungren_deng_peng_lu_mark_berkowitz_horng_etal_2019} is a publicly available dataset containing chest radiographs in JPG format from 2009--2018. Similar to MIMIC-IV, MIMIC-CXR add time annotations placing each sample into a three-year time range. We approximate the year of each sample by taking the midpoint of its time range. Each patient has an \verb|anchor_year_group|, \verb|anchor_year| and \verb|StudyDate|. For each patient, we first calculated an offset as the difference between \verb|StudyDate| and \verb|anchor_year|. Then, we approximated the admit time as the midpoint of \verb|anchor_year_group| after applying the computed offset. The performance over time is evaluated on a \emph{yearly} basis. Our study uses MIMIC-IV-JPG-2.0. A similar training setup to that in \citet{seyyed2020chexclusion} was used (learning rate, architecture, data augmentation, stopping criteria, etc.).

\begin{figure}[!h]
  \includegraphics[width=1.0\columnwidth]{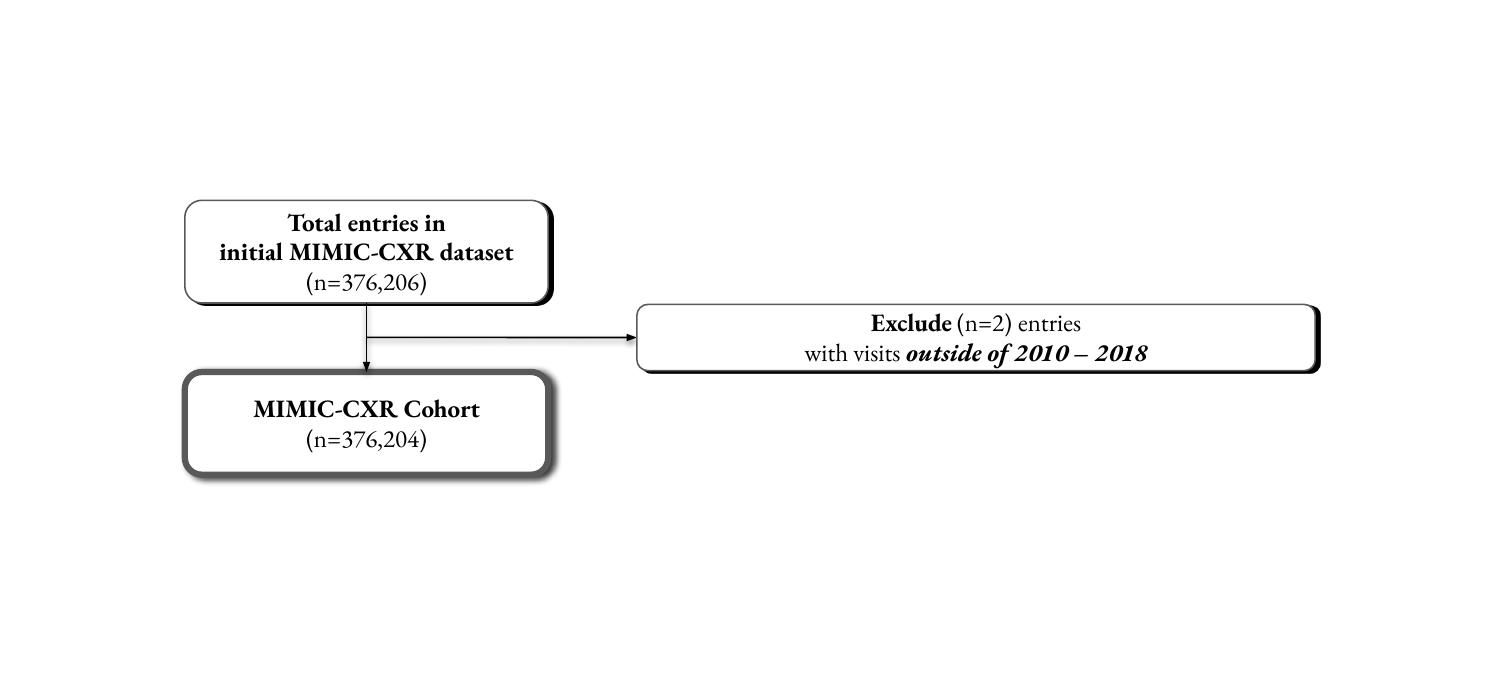}
  \caption{Cohort selection diagram - MIMIC-CXR}
  \label{fig:mimic_cxr_cohort}
\end{figure}

\begin{itemize}
    \item Data access: Users must create a Physionet account, become credentialed, and sign a data use agreement (DUA).
    \item Cohort selection: We removed the records from 2009 due to the tiny sample size. (Selection diagram in Figure \ref{fig:mimic_cxr_cohort}). We keep all records for each patients and split the data based on patient \verb|subject id|.
    \item Outcome definition: The outcome is the probabilities of all labels given the input images. The labels includes 13 abnormal outcomes and 1 normal outcome. (Atelectasis, Cardiomegaly, Consolidation, Edema, Enlarged Cardiomediastinum, Fracture, Lung Lesion, Lung Opacity, Pleural Effusion, Pneumonia, Pneumothorax, Pleural Other, Support Devices, No Finding)
    \item Cohort characteristics: Cohort characteristics are given in Table \ref{tab:mimic_cxr_characteristics}.
\end{itemize}

\subsection{Cohort Selection and Cohort Characteristics}

\begin{table*}[htb]
\caption{MIMIC-CXR cohort characteristics, with count (\%) or median (Q1--Q3).}
\vspace{0.5em}
\label{tab:mimic_cxr_characteristics}
    \centering
    \begin{tabular}{lcccc}
    \toprule
    Feature name (value) & Summary statistic & Empty (ratio) &       Status \\
    \midrule
    \textbf{Gender} &                   &               &              \\
    \hspace{2em}F &   179,765 (47.8\%) &            -- &  categorical \\
    \hspace{2em}M &   196,439 (52.2\%) &            -- &  categorical \\
    \textbf{Age} &        64 (51-76) &          0.0\% &   continuous \\
    \textbf{Diseases} &                   &               &              \\
    \hspace{2em}Atelectasis &    65,390 (17.4\%) &            -- &  categorical \\
    \hspace{2em}Cardiomegaly &    56,404 (15.0\%) &            -- &  categorical \\
    \hspace{2em}Consolidation &     14,394 (3.8\%) &            -- &  categorical \\
    \hspace{2em}Edema &     36,026 (9.6\%) &            -- &  categorical \\
    \hspace{2em}Enlarged Cardiomediastinum &      9,821 (2.6\%) &            -- &  categorical \\
    \hspace{2em}Fracture &      6,314 (1.7\%) &            -- &  categorical \\
    \hspace{2em}Lung Lesion &     10,574 (2.8\%) &            -- &  categorical \\
    \hspace{2em}Lung Opacity &    76,074 (20.2\%) &            -- &  categorical \\
    \hspace{2em}Pleural Effusion &    75,526 (20.1\%) &            -- &  categorical \\
    \hspace{2em}Pleural Other &      3,432 (0.9\%) &            -- &  categorical \\
    \hspace{2em}Pneumonia &     25,065 (6.7\%) &            -- &  categorical \\
    \hspace{2em}Pneumothorax &     12,828 (3.4\%) &            -- &  categorical \\
    \hspace{2em}Support Devices &    69,148 (18.4\%) &            -- &  categorical \\
    \hspace{2em}No Finding &   167,116 (44.4\%) &            -- &  categorical \\
    \bottomrule
    \end{tabular}%
\end{table*}

\subsection{Label level AUROC over time for MIMIC-CXR}
\begin{figure}[H]
\centering
  \includegraphics[width=0.6\columnwidth]{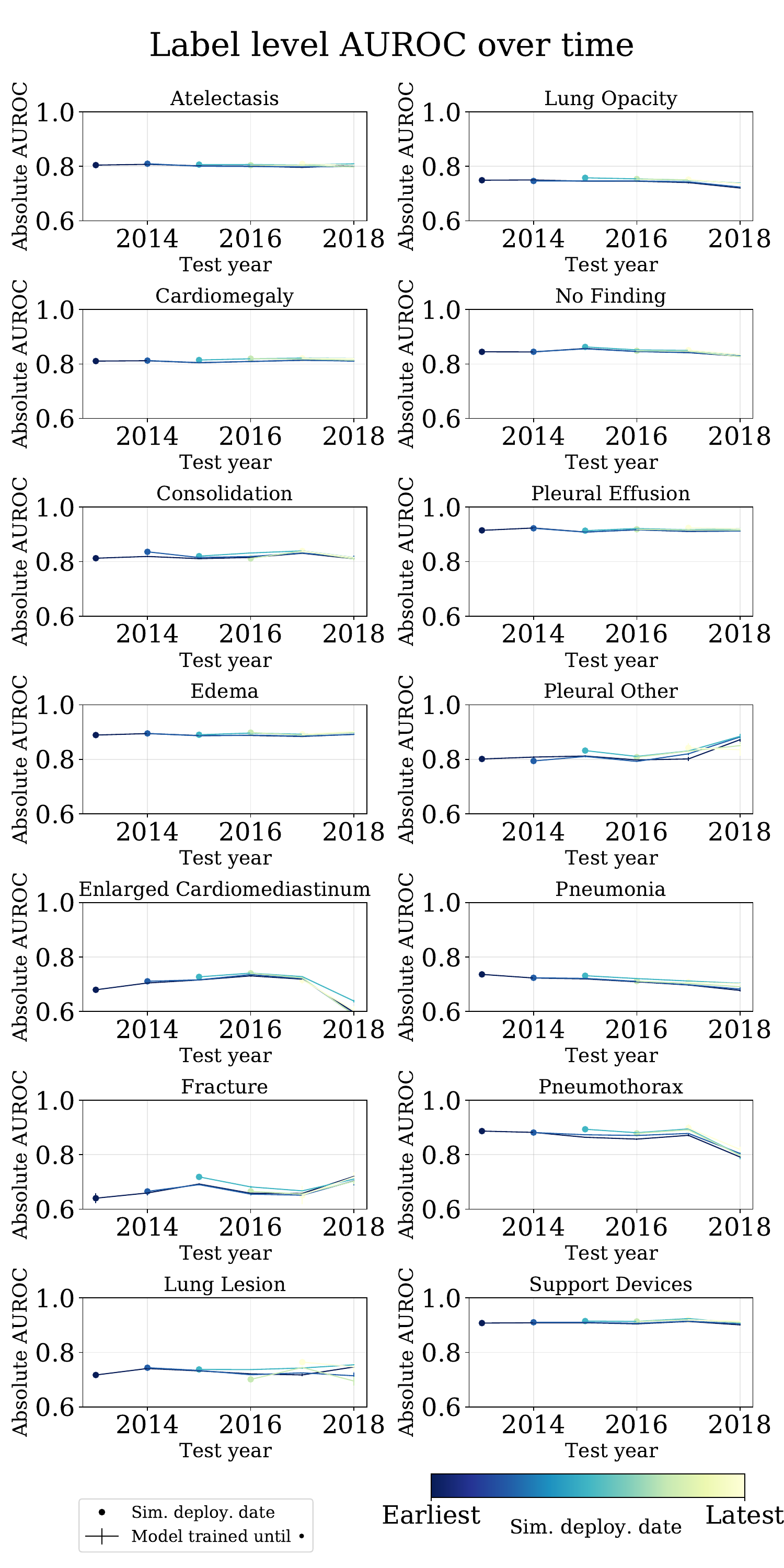}
  \caption{Absolute AUROC over time of each label in MIMIC-CXR}
  \label{fig:mimic_cxr_label_level_auc_over_time}
\end{figure}

\begin{figure}[ht]
\centering
  \includegraphics[width=0.5\columnwidth]{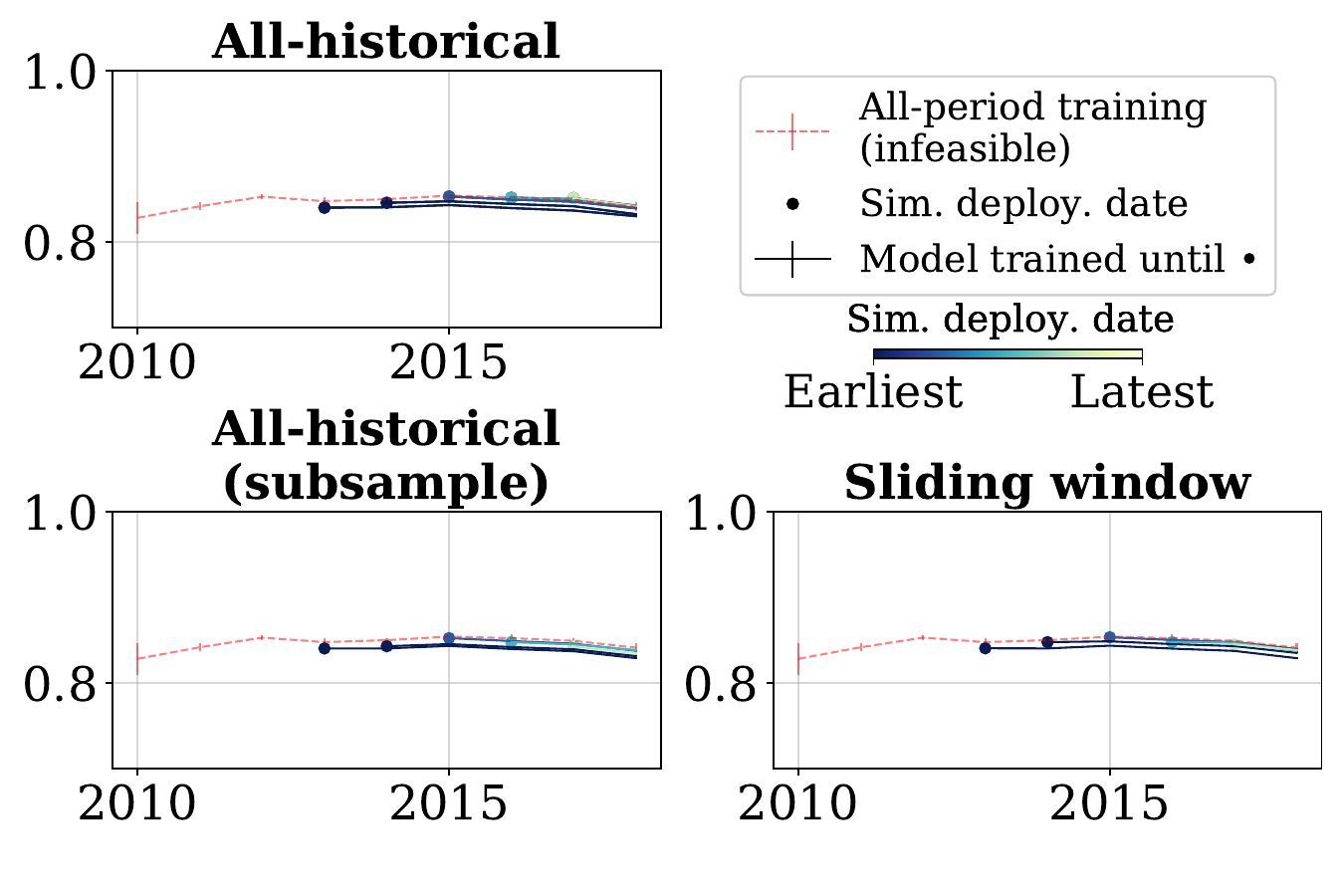}
  \vspace{-2em}
  \caption{Weighted test AUROC vs. year for the DenseNet architecture on MIMIC-CXR. 
  }
  \label{fig:mimic_cxr_absolute_auc_over_time}
\end{figure}

\begin{table*}[ht]
\caption{MIMIC-CXR label-level AUROC from time-agnostic evaluation of all-period training. The format is mean ($\pm$std. dev. across splits)}
\vspace{0.5em}
\label{tab:mimic_cxr_label_level_auc}
\centering
    \begin{tabular}{cccc}
    \toprule
    \textbf{Label} & \textbf{AUROC}  &  \textbf{Label} & 
    \textbf{AUROC} \\
    \midrule
    Atelectasis & 0.826 ($\pm$0.003) & Cardiomegaly & 0.837 ($\pm$0.002) \\
    Consolidation & 0.841 ($\pm$0.003) & Edema & 0.904 ($\pm$0.002) \\
    Enlarged Cardiomediastinum & 0.759 ($\pm$0.005) & Fracture & 0.745 ($\pm$0.006) \\
    Lung Lesion & 0.784 ($\pm$0.003) & Lung Opacity & 0.770 ($\pm$0.002) \\
    Pleural Effusion & 0.929 ($\pm$0.001) & Pleural Other & 0.844 ($\pm$0.009) \\
    Pneumonia & 0.755 ($\pm$0.004) & Pneumothorax & 0.918 ($\pm$0.006) \\
    Support Devices & 0.928 ($\pm$0.001) & No Finding & 0.876 ($\pm$0.002) \\
    \bottomrule
    \end{tabular}%
\end{table*}

\clearpage

\section{Logistic Regression Coefficients from Splitting by Patient}

To help with intuition in important features for the predictive task on each dataset, here we have the coefficients of logistic regression models trained from splitting by patient.

\begin{table*}[ht]
\caption{SEER (Breast) top 10 important features for LR models, all-period training.}
\vspace{0.5em}
\label{tab:seer_breast_coef}
\centering
    \begin{tabular}{lc}
    \toprule
    Feature &  Coefficient \\
    \midrule
    SEER historic stage A (1973-2015)\_Distant &    -2.113944 \\
    SEER historic stage A (1973-2015)\_Localized &     1.676493 \\
    Regional nodes examined (1988+)\_95.0 &    -1.167844 \\
    CS lymph nodes (2004-2015)\_750 &     1.100824 \\
    CS lymph nodes (2004-2015)\_755 &     1.023753 \\
    Histologic Type ICD-O-3\_8530 &    -0.913494 \\
    Histologic Type ICD-O-3\_8543 &     0.902798 \\
    Breast - Adjusted AJCC 6th T (1988-2015)\_T4d &     0.899491 \\
    Histologic Type ICD-O-3\_8211 &     0.877848 \\
    EOD 10 - extent (1988-2003)\_85 &    -0.791136 \\
    \bottomrule
    \end{tabular}%
\end{table*}

\begin{table*}[ht]
\caption{SEER (Colon) top 10 important features for LR models, all-period training.}
\vspace{0.5em}
\label{tab:seer_colon_coef}
\centering
    \begin{tabular}{lc}
    \toprule
    Feature &  Coefficient \\
    \midrule
    Reason no cancer-directed surgery\_Surgery performed &     2.360161 \\
    Regional nodes positive (1988+)\_00 &     1.897706 \\
    Regional nodes positive (1988+)\_01 &     1.872008 \\
    modified AJCC stage 3rd (1988-2003)\_40 &    -1.787481 \\
    EOD 10 - extent (1988-2003)\_13 &     1.766066 \\
    Reason no cancer-directed surgery\_Not recommended,  &    -1.752474 \\
    contraindicated due to other cond; autopsy only (1973-2002)&\\
    EOD 10 - extent (1988-2003)\_85 &    -1.732619 \\
    EOD 10 - extent (1988-2003)\_70 &    -1.704333 \\
    CS mets at dx (2004-2015)\_99 &     1.619905 \\
    CS mets at dx (2004-2015)\_00 &     1.609454 \\
    \bottomrule
    \end{tabular}%
\end{table*}

\begin{table*}[ht]
\caption{SEER (Lung) top 10 important features for LR models, all-period training.}
\vspace{0.5em}
\label{tab:seer_lung_coef}
\centering
    \begin{tabular}{lc}
    \toprule
    Feature &  Coefficient \\
    \midrule
    Histologic Type ICD-O-3\_8240 &     2.514539 \\
    EOD 4 - nodes (1983-1987)\_0 &     2.074730 \\
    EOD 4 - nodes (1983-1987)\_7 &    -1.777530 \\
    EOD 10 - size (1988-2003)\_140 &    -1.587893 \\
    Histologic Type ICD-O-3\_8141 &    -1.546566 \\
    CS tumor size (2004-2015)\_998.0 &    -1.515856 \\
    EOD 4 - nodes (1983-1987)\_6 &    -1.497022 \\
    Type of Reporting Source\_Nursing/convalescent home/hospice &    -1.338998 \\
    CS mets at dx (2004-2015)\_51 &    -1.326595 \\
    EOD 10 - size (1988-2003)\_150 &    -1.326196 \\
    \bottomrule
    \end{tabular}%
\end{table*}

\begin{table*}[ht]
\caption{CDC COVID-19 top 10 important features for LR models, all-period training.}
\vspace{0.5em}
\label{tab:cdc_covid_coef}
\centering
    \begin{tabular}{lc}
    \toprule
    Feature &  Coefficient \\
    \midrule
    res\_state\_DE &     2.202055 \\
    age\_group\_0 - 9 Years &    -2.114818 \\
    age\_group\_80+ Years &     1.965279 \\
    age\_group\_10 - 19 Years &    -1.681099 \\
    res\_state\_GA &     1.391469 \\
    age\_group\_70 - 79 Years &     1.379589 \\
    res\_county\_WICHITA &     1.290644 \\
    age\_group\_20 - 29 Years &    -1.189734 \\
    res\_county\_SUMNER &    -1.135073 \\
    mechvent\_yn\_Yes &     1.117372 \\
    \bottomrule
    \end{tabular}%
\end{table*}

\begin{table*}[ht]
\caption{SWPA COVID-19 top 10 important features for LR models according to experiments splitting by patient.}
\vspace{0.5em}
\label{tab:pa_covid_coef}
\centering
    \resizebox{1\columnwidth}{!}{%
    \begin{tabular}{lc}
    \toprule
    Feature &  Coefficient \\
    \midrule
    age\_bin\_(70, 200]\_0 &    -0.781337 \\
    age\_bin\_(70, 200]\_1 &     0.780673 \\
    medication\_FENTANYL (PF) 50 MCG/ML INJECTION SOLUTION\_0.0 &     0.651419 \\
    medication\_EPINEPHRINE 0.3 MG/0.3 ML INJECTION, AUTO-INJECTOR\_nan &    -0.627565 \\
    medication\_HYDROCORTISONE SOD SUCCINATE (PF) 100 MG/2 ML SOLUTION FOR INJECTION\_0.0 &     0.544222 \\
    medication\_HYDROCODONE 5 MG-ACETAMINOPHEN 325 MG TABLET\_nan &    -0.520368 \\
    medication\_DEXAMETHASONE SODIUM PHOSPHATE 4 MG/ML INJECTION SOLUTION\_0.0 &     0.502954 \\
    medication\_ASPIRIN 81 MG TABLET,DELAYED RELEASE\_nan &    -0.479100 \\
    bmi\_nan &    -0.427569 \\
    age\_bin\_(60, 70]\_0 &    -0.380688 \\
    \bottomrule
    \end{tabular}%
    }
\end{table*}

\begin{table*}[ht]
\caption{MIMIC-IV top 10 important features for LR models, all-period training.}
\vspace{0.5em}
\label{tab:mimic_iv_coef}
\centering
    \begin{tabular}{lc}
    \toprule
    Feature &  Coefficient \\
    \midrule
    O2 Delivery Device(s)\_None &    -0.307334 \\
    Eye Opening\_None &     0.301737 \\
    admit\_age &     0.299712 \\
    O2 Delivery Device(s)\_Nasal cannula &    -0.248463 \\
    Motor Response\_Obeys Commands &    -0.230931 \\
    Pupil Response L\_Non-reactive &     0.223776 \\
    Richmond-RAS Scale\_ 0  Alert and calm &    -0.205476 \\
    Temp Site\_Blood &    -0.204514 \\
    HR\_0.0 &     0.197299 \\
    Diet Type\_NPO &     0.195156 \\
    \bottomrule
    \end{tabular}%
\end{table*}

\begin{table*}[ht]
\caption{OPTN (Liver) top 10 important features for LR models, all-period training.}
\vspace{0.5em}
\label{tab:optn_liver_coef}
\centering
    \begin{tabular}{lc}
    \toprule
    Feature &  Coefficient \\
    \midrule
    SERUM\_CREAT\_DELTA &     0.660589 \\
    FUNC\_STAT\_TCR\_2020.0 &     0.241507 \\
    FUNC\_STAT\_TCR\_2080.0 &    -0.236288 \\
    DGNC\_4110.0 &    -0.234680 \\
    REGION\_5.0 &     0.223940 \\
    EDUCATION\_998.0 &     0.218549 \\
    ASCITES\_3.0 &     0.218329 \\
    ASCITES\_1.0 &    -0.214076 \\
    INIT\_OPO\_CTR\_CODE\_1054 &    -0.209265 \\
    INIT\_OPO\_CTR\_CODE\_4743 &    -0.207778 \\
    \bottomrule
    \end{tabular}%
\end{table*}
\clearpage

\section{Diagnostic plots}\label{app:diagnostic_plot}

We took the union of the top $k$ most important features from each time point to be included in the diagnostic plots, where $k$ was tuned depending on the dataset so that the resulting plots would not be overcrowded.
For categorical features, we additionally highlighted (using a thicker line) features that had consistently high prevalence ($\geq p$) or experienced a large change in prevalence across one time point ($\geq \Delta$). 
The specific parameters of each dataset are defined in each subsection.
For numerical features, we highlighted features whose average ranking across all time points was $\leq 3$ (also chosen to avoid overcrowding).

\subsection{SEER (Breast)}

For SEER (Breast) diagnostic plots, important features were selected using $k=5, p=0.4, \Delta=0.2$. 

\begin{figure}[H]
  \includegraphics[width=0.9\columnwidth]{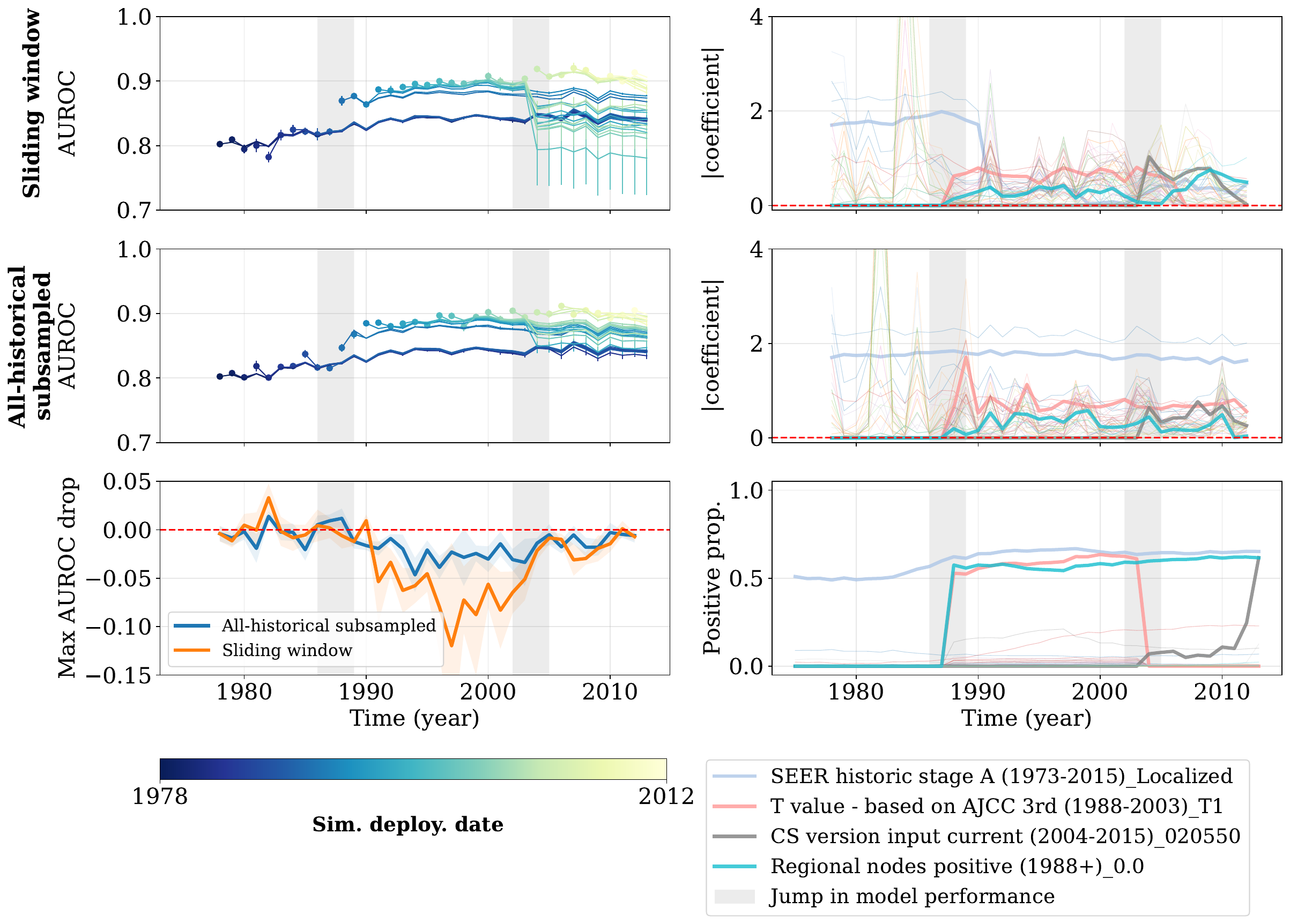}
  \caption{Diagnostic plot of SEER (Breast) dataset. The important features are selected as the union of the top 5 features that have the highest absolute value model coefficients. The left column includes AUROC versus time for both sliding window and all-historical subsampled, and the maximum AUROC drop for each trained model. The right column provides the absolute coefficients of each trained model from both regimes, and positive proportion of the significant features over time. As shown in the gray highlighted region, there are jumps in performance around 1988 and 2003, which coincides with the introducing and removal of several features (e.g. T value - based on AJCC 3rd (1988-2003)\_T1). The latency of jumps in coefficients are caused by length of sliding window.}
  \label{fig:diagnositic_plot_seer_breast}
\end{figure}

\clearpage

\subsection{SEER (Colon)}
For SEER (Colon) diagnostic plots, important features were selected using $k=3, p=0.4, \Delta=0.2$.

\begin{figure}[H]
  \includegraphics[width=0.95\columnwidth]{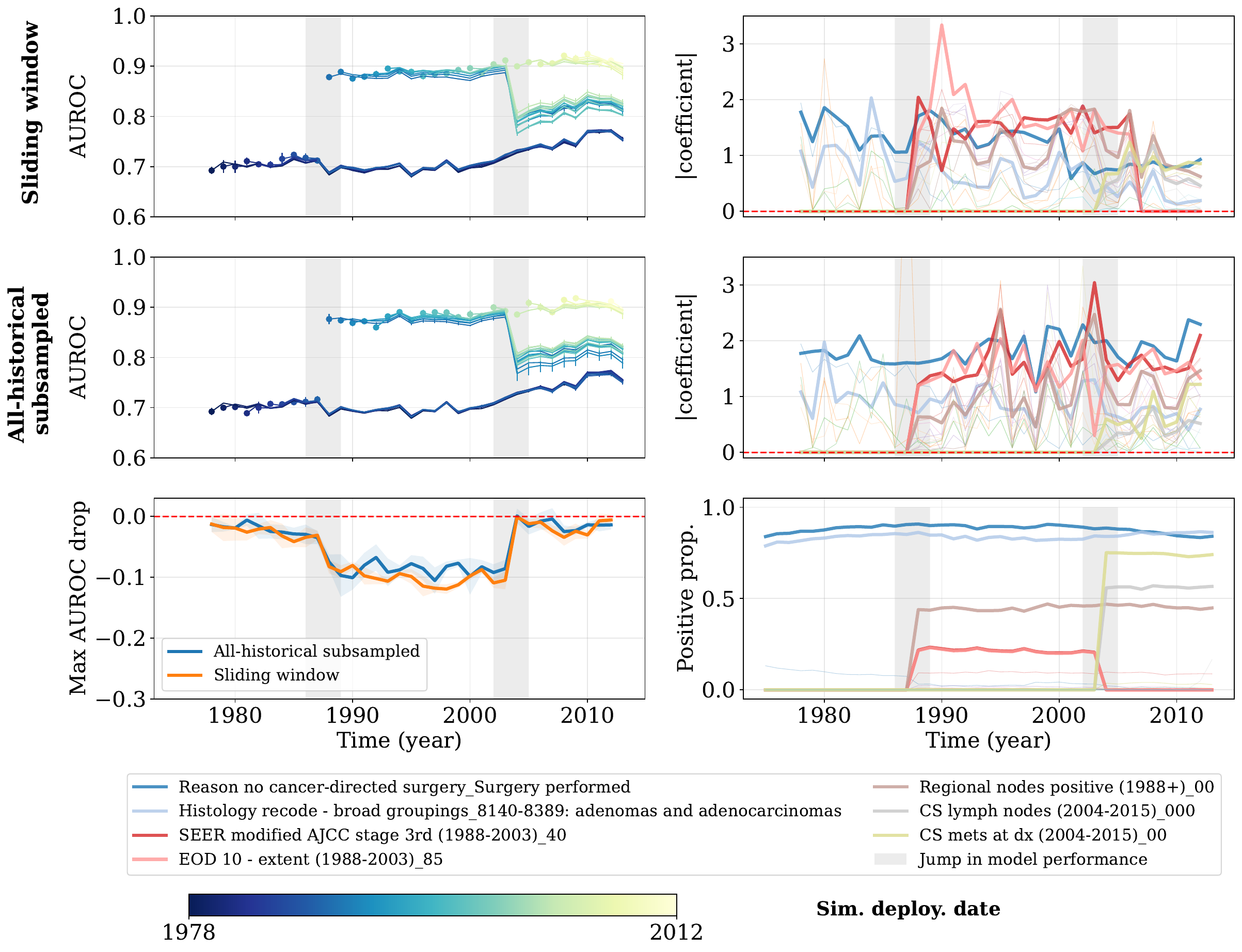}
  \caption{Diagnostic plot of SEER (Colon) dataset. The important features are selected as the union of the top 3 features that have the highest absolute model coefficients. The left column includes AUROC versus time for both sliding window and all-historical subsampled, and the maximum AUROC drop for each trained model. The right column provides the absolute coefficients of each trained model from both regimes, and positive proportion of the significant features over time. As shown in the gray highlighted region, there are jumps in performance around 1988 and 2003, which coincides with the introducing and removal of several features (e.g. SEER modified AJCC stage 3rd (1988-2003)\_40). The latency of jumps in coefficients are caused by length of sliding window.}
  \label{fig:diagnositic_plot_seer_colon}
\end{figure}

\clearpage

\subsection{SEER (Lung)}
For SEER (Lung) diagnostic plots, important features were selected using $k=5, p=0.2, \Delta = 0.2$.

\begin{figure}[H]
\centering
  \includegraphics[width=1.0\columnwidth]{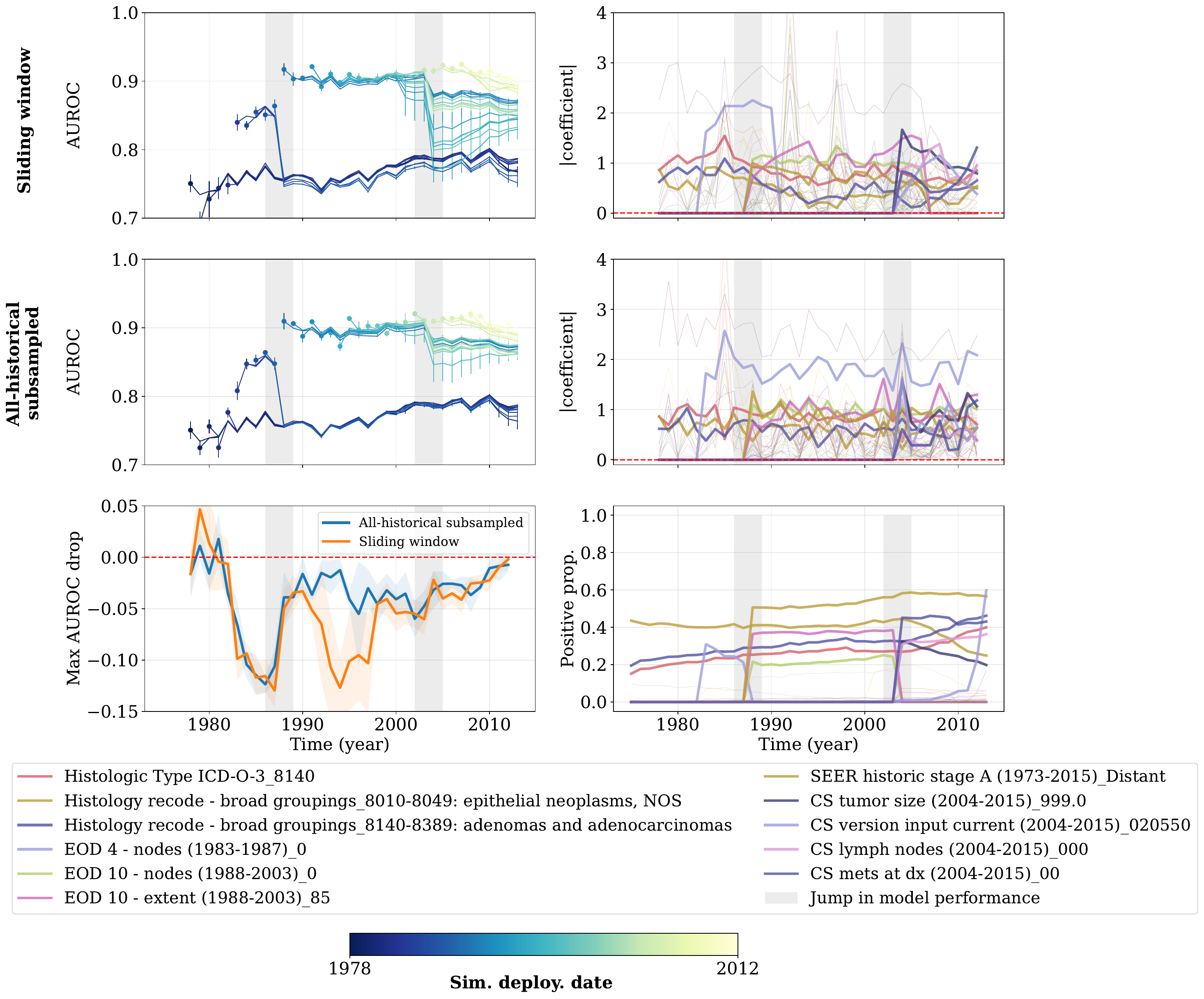}
  \caption{Diagnostic plot of SEER (Lung) dataset. The important features are selected as the union of the top 5 features that have the highest absolute model coefficients. The left column includes AUROC versus time for both sliding window and all-historical subsampled, and the maximum AUROC drop for each trained model. The right column provides the absolute coefficients of each trained model from both regimes, and positive proportion of the significant features over time. As shown in the gray highlighted region, there are jumps in performance around 1988 and 2003, which coincides with the introducing and removal of several features (e.g. EOD 10 - nodes (1988-2013)\_0 \& EOD 10 - extent (1988-2003)\_85). The latency of jumps in coefficients are caused by length of sliding window.}
  \label{fig:diagnositic_plot_seer_lung}
\end{figure}

\clearpage

\subsection{CDC COVID-19}

For CDC COVID-19 diagnostic plots, important features were selected using $k = 5, p=0.15, \Delta=0.15$.

\begin{figure}[H]
  \includegraphics[width=1.0\columnwidth]{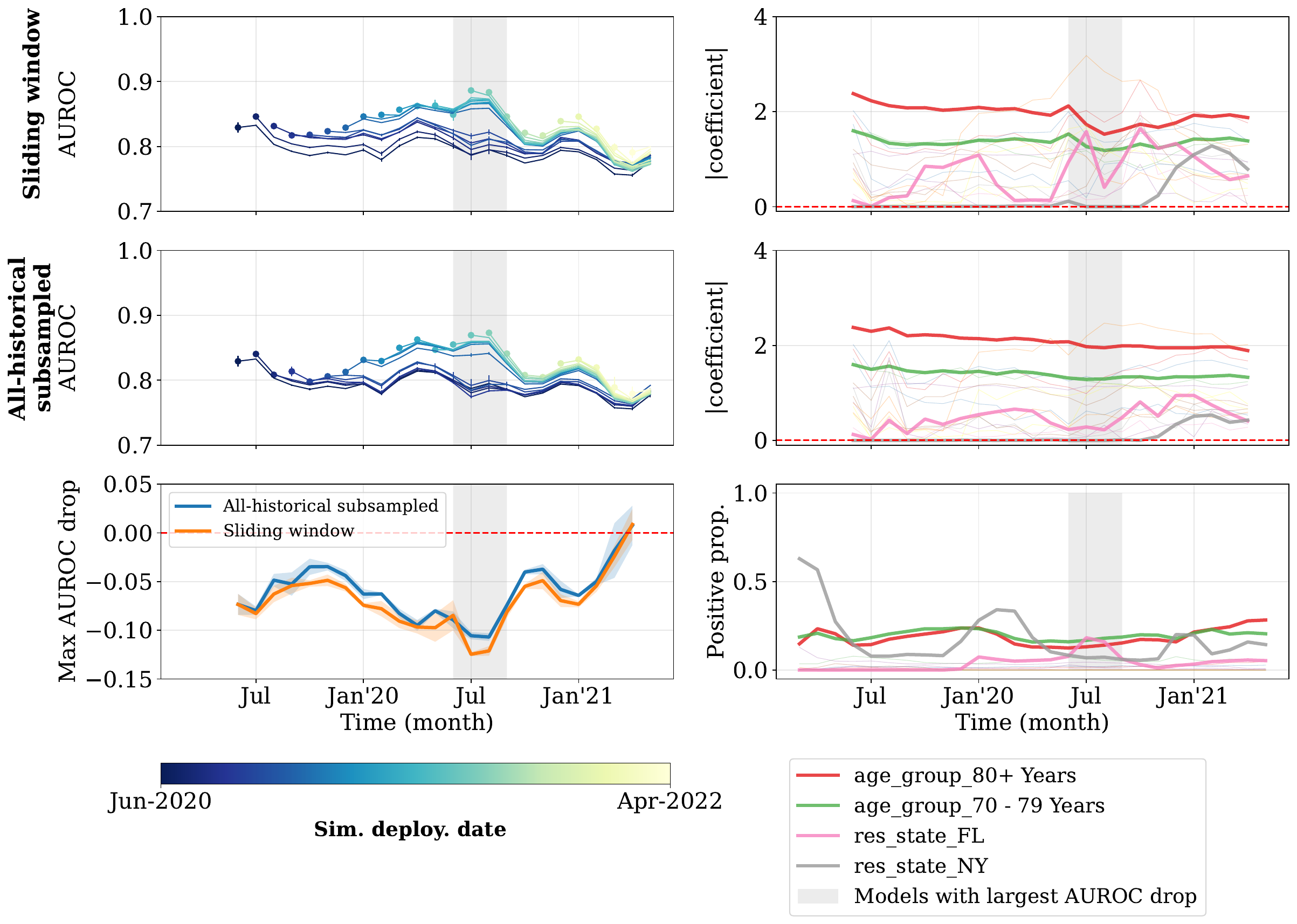}
  \caption{Diagnostic plot of CDC COVID-19. The important features are selected as the union of the top 5 features that have the highest absolute model coefficients. The left column includes AUROC versus time for both sliding window and all-historical subsampled, and the maximum AUROC drop for each trained model. The right column provides the absolute coefficients of each trained model from both regimes, and positive proportion of the significant features over time. As shown in the gray highlighted region, the models trained around June 2021 suffer the largest maximum AUROC drop, coinciding with a shift in distribution of ages (Figure \ref{fig:stack_age_group}) and states (Figure \ref{fig:stack_state_residence}). The latency of jumps in coefficients are caused by length of sliding window.}
  \label{fig:diagnositic_plot_cdc_covid}
\end{figure}

\clearpage

\subsection{SWPA COVID-19}

For SWPA COVID-19 diagnostic plots, important features were selected using $k=3, p=0.4, \Delta=0.2$.

\begin{figure}[H]
  \includegraphics[width=1.0\columnwidth]{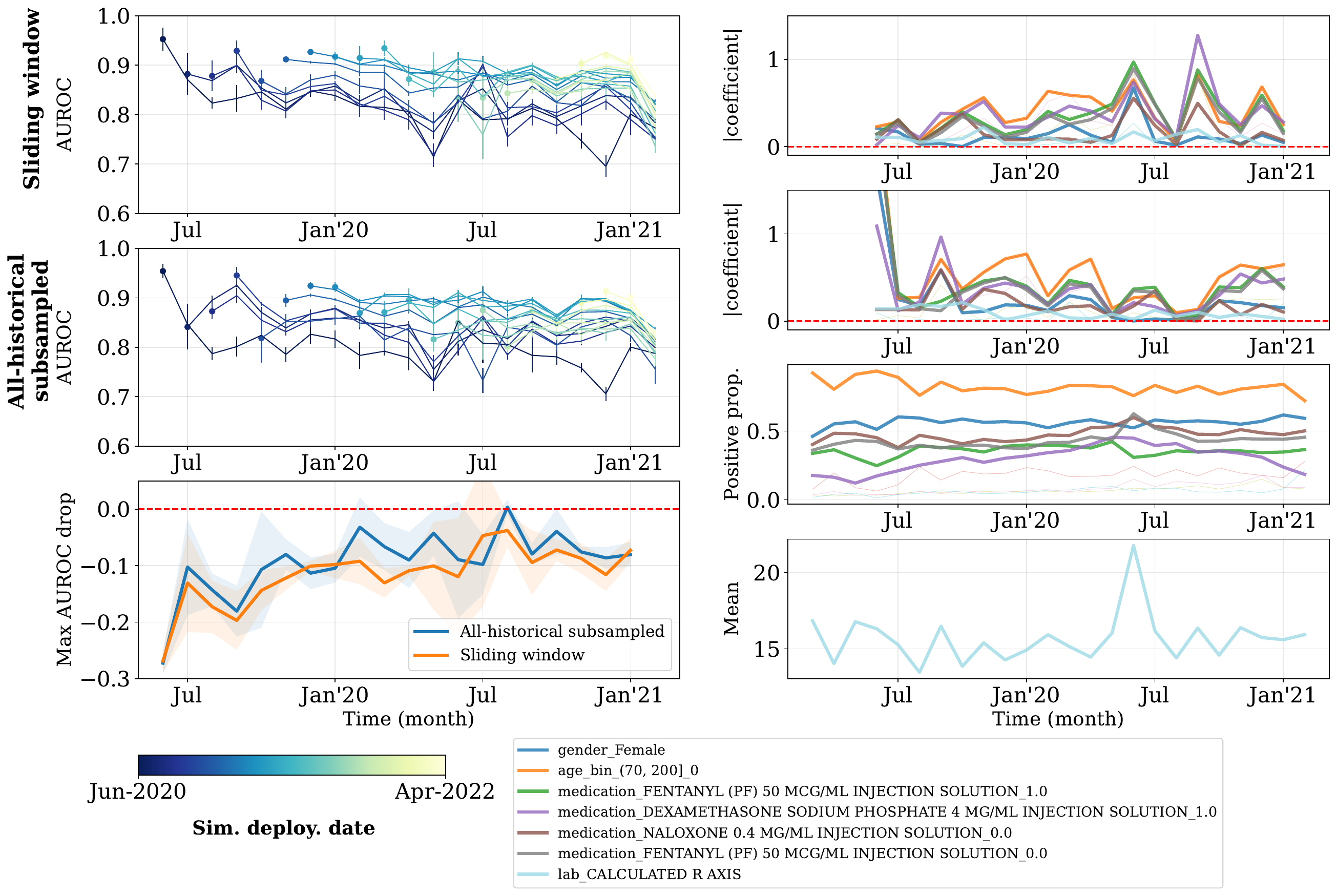}
  \caption{Diagnostic plot of SWPA COVID-19. The important features are selected as the union of the top 3 features that have the highest absolute model coefficients. The left column includes AUROC versus time for both sliding window and all-historical subsampled, and the maximum AUROC drop for each trained model. The right column provides the absolute coefficients of each trained model from both regimes, and positive proportion of the significant features over time. One of the hypotheses for relatively large uncertainty is smaller sample size.}
  \label{fig:diagnositic_plot_pa_covid}
\end{figure}

\clearpage

\subsection{MIMIC-IV}

For MIMIC-IV diagnostic plots, important features were selected using $k=3, p=0.4, \Delta=0.2$.

\begin{figure}[H]
  \includegraphics[width=1.0\columnwidth]{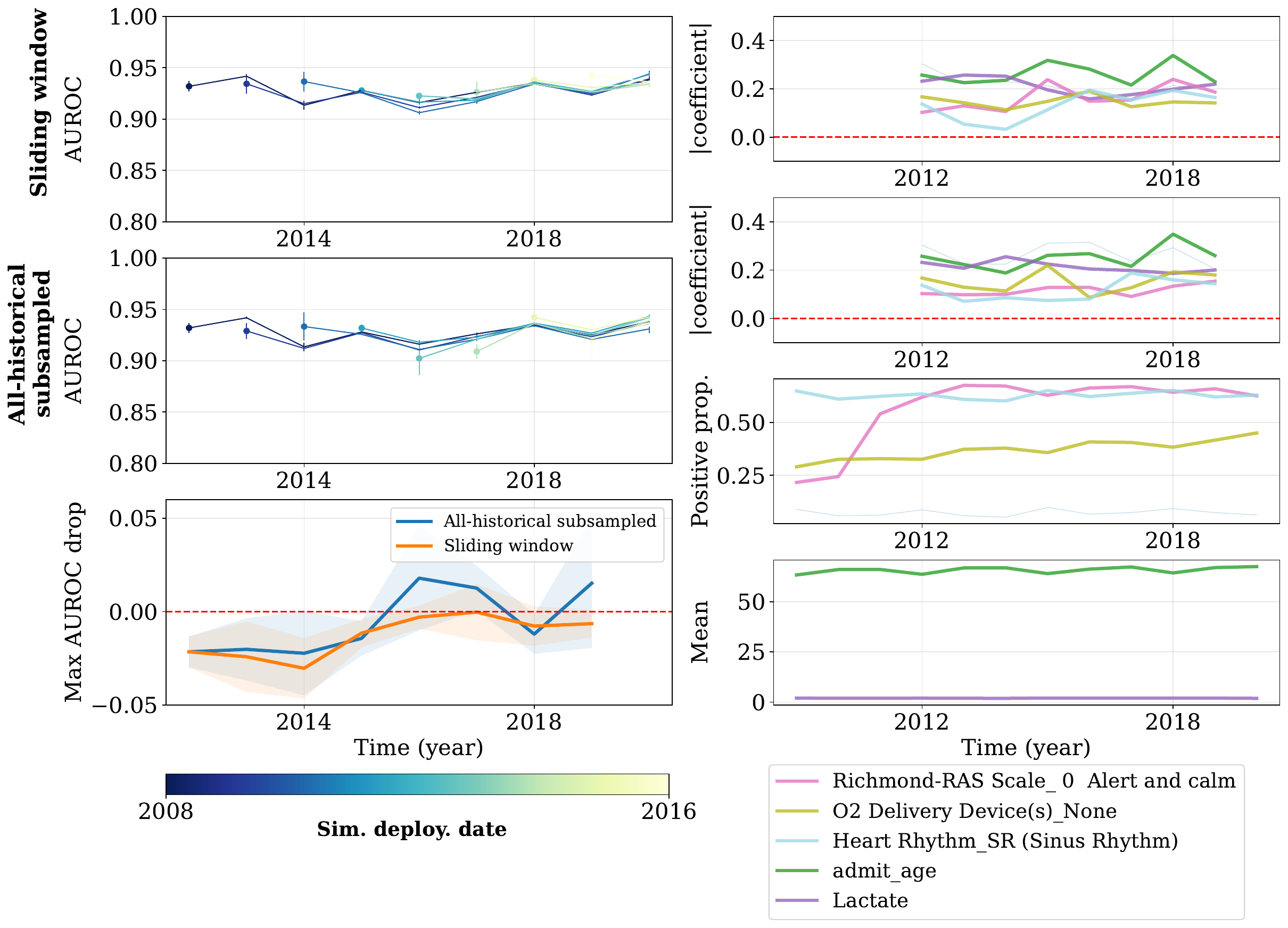}
  \caption{Diagnostic plot of MIMIC-IV. The important features are selected as the union of the top 3 features that have the highest absolute model coefficients. The left column includes AUROC versus time for both sliding window and all-historical subsampled, and the maximum AUROC drop for each trained model. The right column provides the absolute coefficients of each trained model from both regimes, and positive proportion of the significant features over time. The model performance is relatively stable, coinciding with relatively stable distributions of a majority of important features.}
  \label{fig:diagnositic_plot_mimic_iv}
\end{figure}

\clearpage

\subsection{OPTN (Liver)}

For OPTN (Liver) diagnostic plots, important features were selected using $k=3, p=0.4, \Delta=0.2$.

\begin{figure}[H]
  \includegraphics[width=0.95\columnwidth]{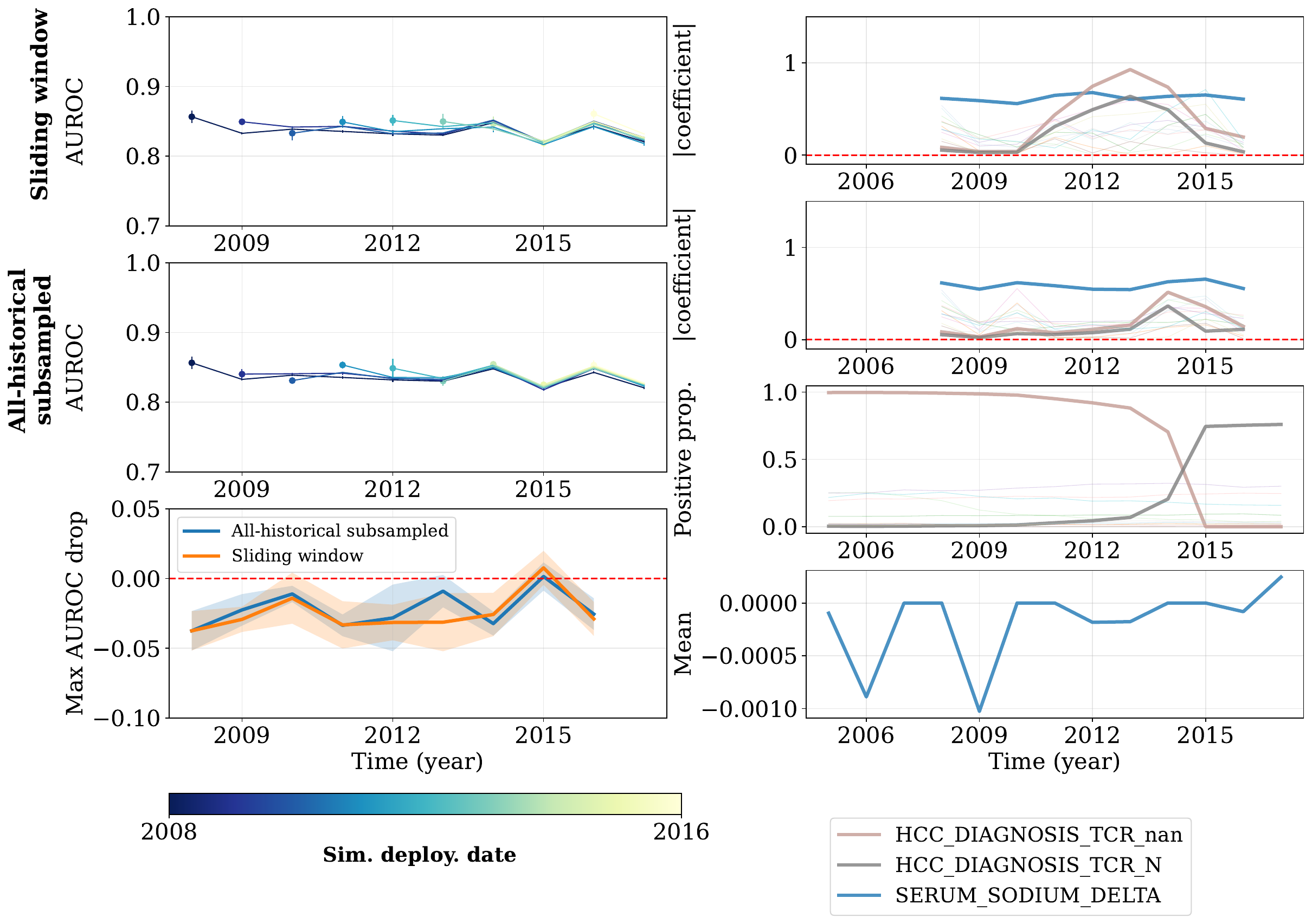}
  \caption{Diagnostic plot of OPTN (Liver). The important features are selected as the union of the top 3 features that have the highest absolute model coefficients. The left column includes AUROC versus time for both sliding window and all-historical subsampled, and the maximum AUROC drop for each trained model. The right column provides the absolute coefficients of each trained model from both regimes, and positive proportion of the significant features over time. Although the HCC DIAGNOSIS TCR binary features change in positive proportion over time, these features were not always important, and the other important features (faded) maintain relatively stable proportions across time. Overall, model performance is quite stable over time.}
  \label{fig:diagnositic_plot_optn_liver}
\end{figure}

\clearpage

\subsection{MIMIC-CXR}
\begin{figure}[H]
\centering
\includegraphics[width=0.7\columnwidth]{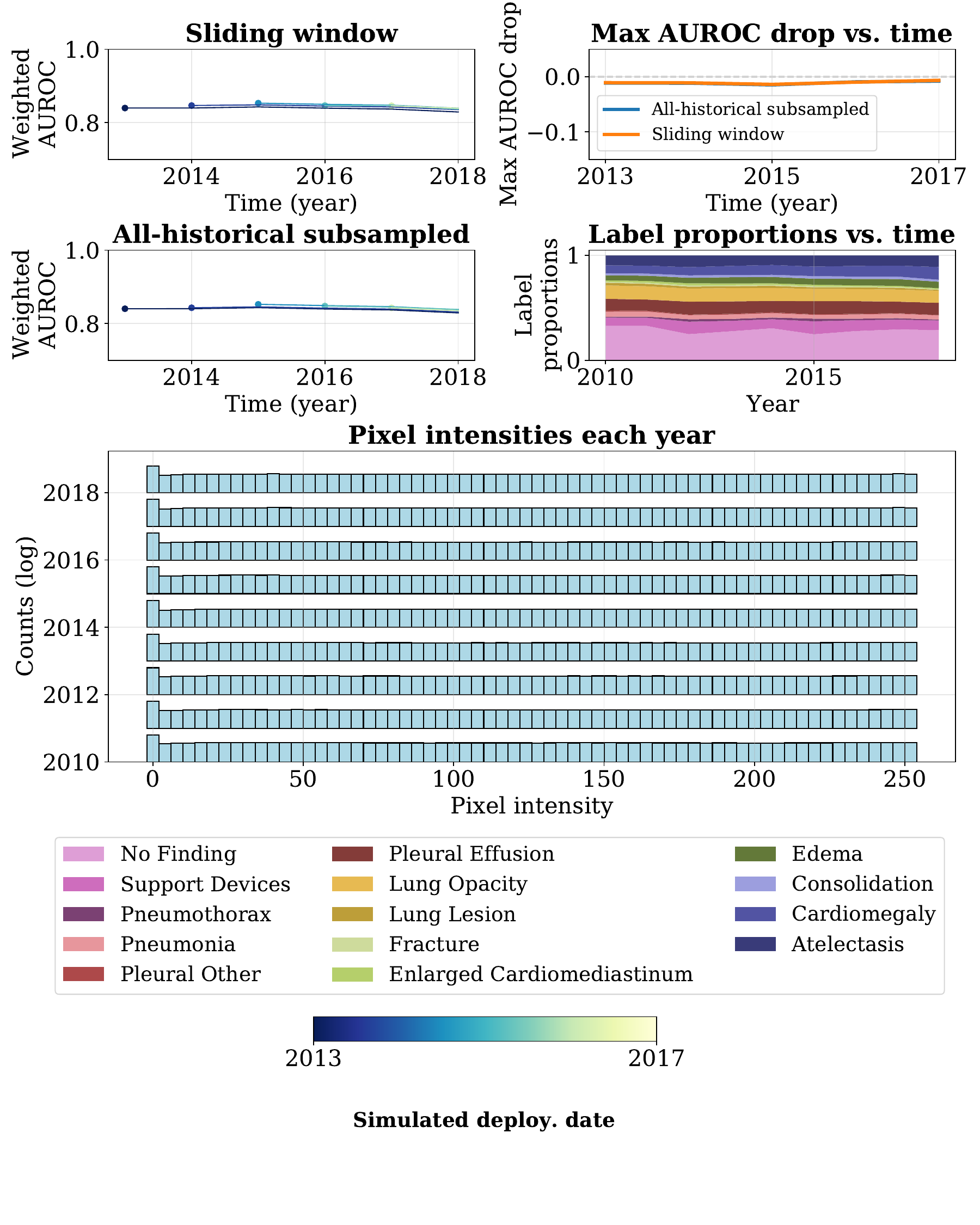}
  \caption{Diagnostic plot of MIMIC-CXR. The top and mid left includes AUROC versus time for both sliding window and all-historical subsampled. The top right is the maximum AUROC drop for each trained model. The mid-right provides the label proportions over time. The bottom shows pixel intensities for images in each year. The histogram of pixel intensity is stable over time, which is consistent with the small variation in model performance over time}
  \label{fig:diagnositic_plot_mimic_cxr}
\end{figure}

\clearpage

\section{Model performance over time from three models}
\label{app:alternative_metrics}

\subsection{AUROC}

All plots in this section are for the all-historical training regime.

\begin{figure}[H]
\centering
  \includegraphics[width=0.81\columnwidth]{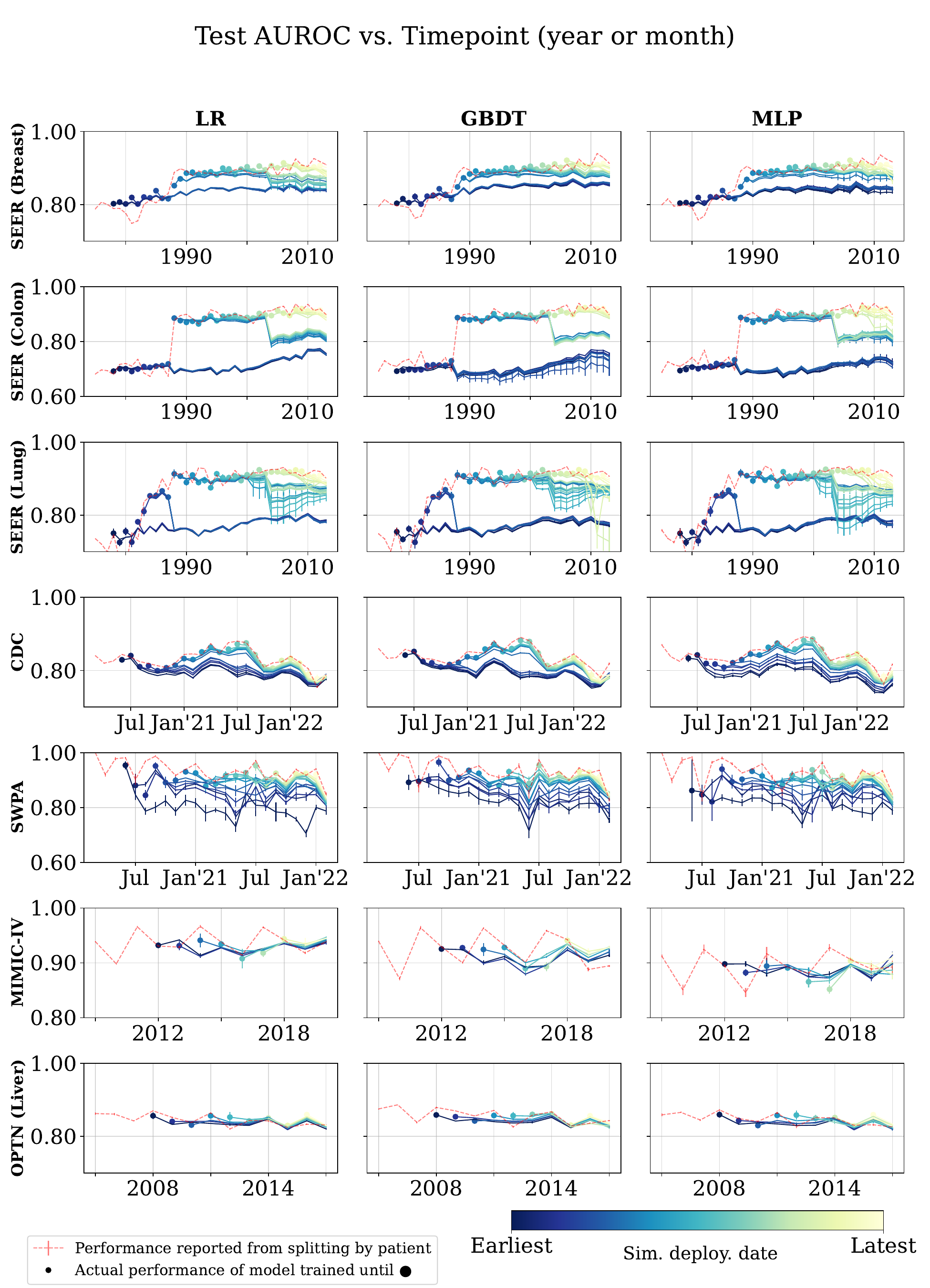}
  \caption{AUROC versus test timepoints from three model classes on all datasets.}
  \label{fig:auc_over_time}
\end{figure}

\clearpage

\subsection{AUPRC}
All plots in this section are for the all-historical training regime.

\begin{figure}[H]
\centering
  \includegraphics[width=0.81\columnwidth]{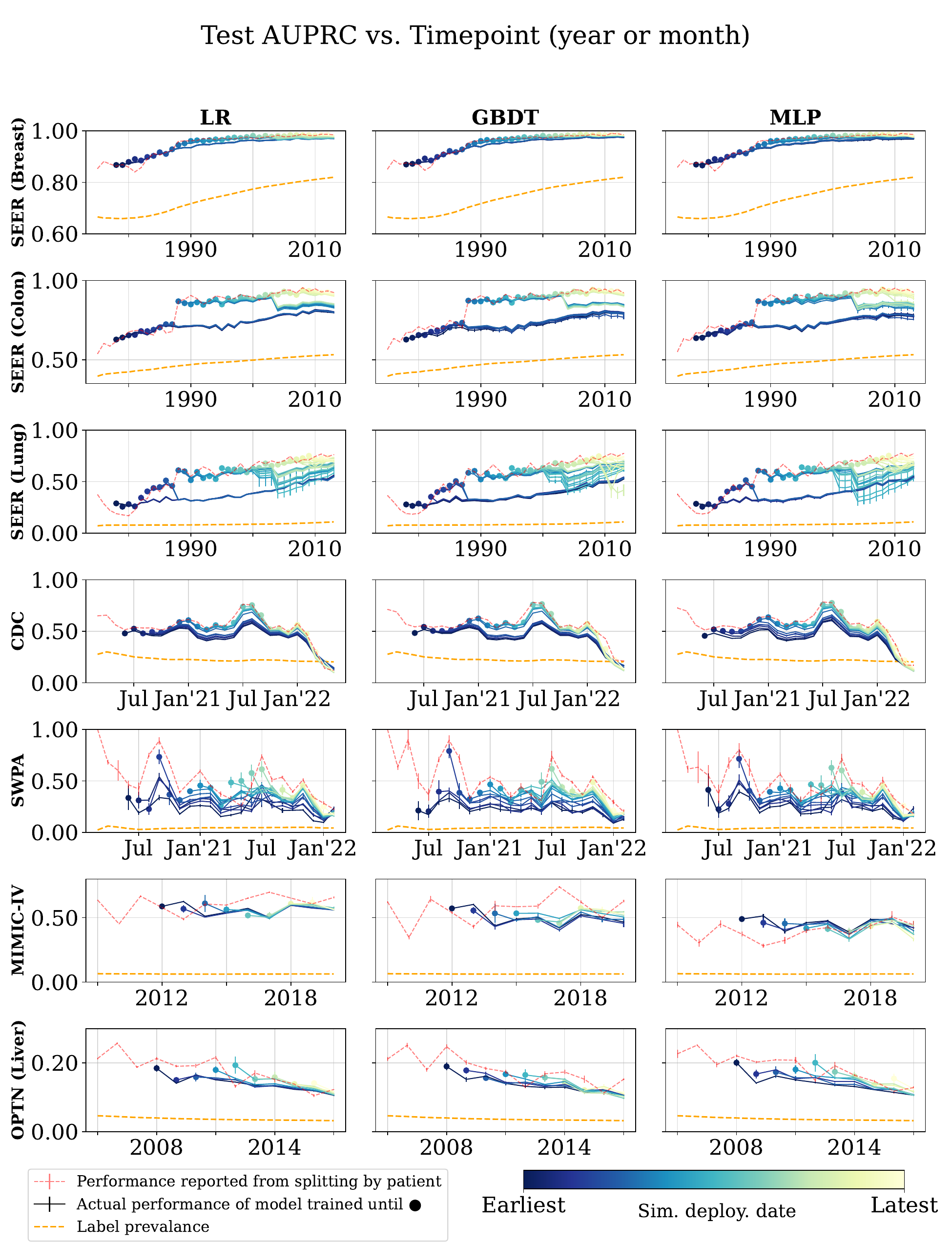}
  \caption{AUPRC versus test timepoints from three model classes on all datasets. Label prevalance refers to the ratio of accumulated positive labels over time.}
  \label{fig:auprc_over_time}
\end{figure}

\clearpage

\section{Data Split Details} \label{app:datasplit_details}
\begin{table*}[h]
\caption{Split ratio for each dataset for training, validation and testing (both for time-agnostic splits and in-period splits).}
\label{tab:datasplits}
\centering
    \begin{tabular}{lc}
    \toprule
    Dataset & Split ratio \\
    \midrule
    SEER (Breast) & 0.8-0.1-0.1 \\
    SEER (Colon) & 0.8-0.1-0.1 \\
    SEER (Lung) & 0.8-0.1-0.1 \\
    CDC COVID-19 & 0.8-0.1-0.1 \\
    SWPA COVID-19 & 0.5-0.25-0.25 \\
    MIMIC-IV & 0.5-0.25-0.25 \\
    OPTN (Liver) & 0.5-0.25-0.25 \\
    MIMIC-CXR & 0.5-0.25-0.25 \\
    \bottomrule
    \end{tabular}
\end{table*}

\section{Hyperparameter Grids}\label{app:hyperparameter_grids}
\begin{table*}[ht]
\caption{Hyperparameter grids for model training.}
\label{tab:hyper_grid}
\centering
    \begin{tabular}{lc}
    \toprule
    Parameter & Values Considered \\
    \midrule
    \textbf{LR} & \\
    \hspace{1em} C & 0.01, 0.1, 1, 10, $10^2$, $10^3$, $10^4$, $10^5$ \\
    \textbf{GBDT} & \\
    \hspace{1em} n\_estimators & 50, 100 \\
    \hspace{1em} max\_depth & 3, 5 \\
    \hspace{1em} learning\_rate & 0.01, 0.1 \\
    \textbf{MLP} & \\
    \hspace{1em} hidden\_layer\_sizes & 3, 5 \\
    \hspace{1em} learning\_rate\_init & $10^{-4}$, $10^{-3}$, 0.01 \\
    \bottomrule
    \end{tabular}
\end{table*}

\clearpage
\section{AUROC from full-period training} \label{app:standard_auc}
\begin{table*}[h]
\caption{AUROC report from full-period training, the results are in format mean ($\pm$std. dev. across splits)}
\label{tab:standard_auc}
\centering
    \begin{tabular}{ccc}
    \toprule
    Dataset & Model & Full-period AUROC \\
    \midrule
     & LR & 0.888 ($\pm$\num{0.002}) \\
     SEER (Breast) & GBDT & 0.891 ($\pm$\num{0.002}) \\
     & MLP & 0.891 ($\pm$\num{0.002}) \\
     \midrule
     & LR & 0.863 ($\pm$\num{0.003}) \\
     SEER (Colon) & GBDT & 0.868 ($\pm$\num{0.002}) \\
     & MLP & 0.869 ($\pm$\num{0.003}) \\
     \midrule
     & LR & 0.894 ($\pm$\num{0.002}) \\
     SEER (Lung) & GBDT & 0.894 ($\pm$\num{0.002}) \\
     & MLP & 0.898 ($\pm$\num{0.002}) \\
     \midrule
     & LR & 0.837 ($\pm$\num{0.001}) \\
     CDC COVID-19 & GBDT & 0.851 ($\pm$\num{0.001}) \\
     & MLP & 0.852 ($\pm$\num{0.002}) \\
     \midrule
     & LR & 0.928 ($\pm$\num{0.005}) \\
     SWPA COVID-19 & GBDT & 0.930 ($\pm$\num{0.004}) \\
     & MLP & 0.928 ($\pm$\num{0.006}) \\
     \midrule
     & LR & 0.935 ($\pm$\num{0.003}) \\
     MIMIC-IV & GBDT & 0.931 ($\pm$\num{0.002}) \\
     & MLP & 0.898 ($\pm$\num{0.008}) \\
     \midrule
     & LR & 0.846 ($\pm$\num{0.005}) \\
     OPTN (Liver) & GBDT & 0.854 ($\pm$\num{0.005}) \\
     & MLP & 0.847 ($\pm$\num{0.006}) \\
     \midrule
     MIMIC-CXR & DenseNet & 0.860 ($\pm$\num{0.001}) \\
    \bottomrule
    \end{tabular}
\end{table*}

\end{document}